\documentclass[12pt]{jhuthesis}
\usepackage{latexsym}
\usepackage{mycommands}

\begin{document}
\bibliographystyle{plain}

\title{Exploiting Syntactic Structure for Natural Language Modeling}
\author{Ciprian Chelba} \doctorphilosophy \dissertation
\copyrightnotice

\maketitle
\begin{frontmatter}
\begin{abstract}
  The thesis presents an attempt at using the syntactic structure in
  natural language for improved language models for speech
  recognition. The structured language model merges techniques in
  automatic parsing and language modeling using an original
  probabilistic parameterization of a shift-reduce parser. A maximum
  likelihood reestimation procedure belonging to the class of
  expectation-maximization algorithms is employed for training the
  model. Experiments on the Wall Street Journal, Switchboard and
  Broadcast News corpora show improvement in both perplexity and word
  error rate --- word lattice rescoring --- over the standard 3-gram
  language model.
  
  The significance of the thesis lies in presenting an original
  approach to language modeling that uses the hierarchical ---
  syntactic --- structure in natural language to improve on current
  3-gram modeling techniques for large vocabulary speech recognition.

  \vspace{3cm}
  \noindent Advisor: Prof. Frederick Jelinek\\
  Readers: Prof. Frederick Jelinek and Prof. Michael Miller\\
\end{abstract}

\begin{acknowledgements}
  The years I have spent at Hopkins taught me many valuable lessons,
  usually through people I have interacted with and to whom I am
  grateful.
  
  I am thankful to my advisor Frederick Jelinek.
  
  Bill Byrne and Sanjeev Khudanpur for their insightful comments and
  assistance on technical issues and not only.
  
  The members of the Dependency Modeling during the summer '96 DoD
  Workshop, especially: Harry Printz, Eric Ristad and Andreas Stolcke
  for their support on technical and programming matters. This thesis
  would have been on a different topic without the creative
  environment during that workshop.
  
  The people on the STIMULATE grant: Eric Brill, Fred Jelinek, Sanjeev
  Khudanpur, David Yarowski.
  
  Ponani Gopalakrishnan who patiently guided my first steps in the
  practical aspects of speech recognition.
  
  My former academic advisor in Bucharest, Vasile Buzuloiu, for
  encouraging me to further my education.
  
  My colleagues and friends at the CLSP, for bearing with me all these
  years, helping me in my work, engaging in often useless
  conversations and making this a fun time of my life: Radu Florian,
  Asela Gunawardana, Vaibhava Goel, John Henderson, Xiaoxiang Luo,
  Lidia Mangu, John McDonough, Makis Potamianos, Grace Ngai, Murat
  Saraclar, Eric Wheeler, Jun Wu, Dimitra Vergyri.
  
  Amy Berdann, Janet Lamberti and Kimberly Shiring Petropoulos at the
  CLSP for help on all sort of things, literally.
  
  Jacob Laderman for keeping the CLSP machines up and running.
  
  My friends who were there when thesis work was the last thing I
  wanted to discuss about: Lynn Anderson, Delphine Dahan, Wolfgang
  Himmelbauer, Derek Houston, Mihai Pop, Victor and Delia Velculescu.
  
  My host family, Ed and Sue Dickey, for offering advice and help in a
  new culture, making me so welcome at the beginning of my stay in
  Baltimore and thereafter.
  
  My parents, whose support and encouragement was always there when I
  needed it.
\end{acknowledgements}
\begin{dedication}
  To my parents
\end{dedication}
\tableofcontents
\listoftables \listoffigures
\end{frontmatter}

\chapter*{Introduction}
%\section{Language Modeling for Speech Recognition}

In the accepted statistical formulation of the speech recognition
problem \cite{jelinek97} the recognizer seeks to find the word string
$$
\widehat{W}\doteq \arg \max _{W}P(A|W)\,P(W)$$
where $A$ denotes
the observable speech signal, $P(A|W)$ is the probability that when
the word string $W$ is spoken, the signal $A$ results, and $P(W)$ is
the a priori probability that the speaker will utter $W$.

The language model estimates the values $P(W)$. With $W=
w_1,w_2,\ldots,w_n$ we get by Bayes' theorem,
\begin{eqnarray}
  P(W)=\prod_{i=1}^nP(w_i|w_1,w_2,\ldots,w_{i-1}) 
\end{eqnarray}
Since the parameter space of $P(w_k|w_1,w_2,\ldots,w_{k-1})$ is too
large \footnote{The words $w_j$ belong to a vocabulary ${\mathcal V}$
  whose size is in the tens of thousands.}, the language model is
forced to put the {\it history } $W_{k-1}=w_1,w_2,\ldots,w_{k-1}$ into
an equivalence class determined by a function $\Phi(W_{k-1})$. As a
result,
\begin{equation}
\label{e1}P(W)\cong\prod_{k=1}^nP(w_k|\Phi (W_{k-1})) 
\end{equation}
Research in language modeling consists of finding appropriate
equivalence classifiers $\Phi$ and methods to estimate
$P(w_k|\Phi(W_{k-1}))$.

The language model of state-of-the-art speech recognizers uses
$(n-1)$-gram
equivalence classification, that is, defines%
$$
\Phi (W_{k-1})\doteq w_{k-n+1},w_{k-n+2},\ldots,w_{k-1}
$$
Once the form $\Phi (W_{k-1})$ is specified, only the problem of
estimating $P(w_k|\Phi (W_{k-1}))$ from training data remains.

In most cases, $n=3$ which leads to a {\it trigram }language model.
The latter has been shown to be surprisingly powerful and,
essentially, all attempts to improve on it in the last 20 years have
failed. The one interesting enhancement, facilitated by maximum
entropy estimation methodology, has been the use of {\it triggers}
\cite{rosenfeld94} or of {\it singular value decomposition}
\cite{bellegarda97} (either of which dynamically identify the topic of
discourse) in combination with $n-$gram models .

\subsection*{Measures of Language Model Quality}

\paragraph{Word Error Rate}
 One possibility to measure the quality of a language
model is to evaluate it as part of a speech recognizer.
The measure of success is the word error rate; to
calculate it we need to first find the most
favorable word alignment between the hypothesis put out by the
recognizer $\widehat{W}$ and the true sequence of words uttered by the
speaker $W$ --- assumed to be
known a priori for evaluation purposes only ---  and then count the
number of incorrect words in $\widehat{W}$  per total number of words in
$W$.

\begin{verbatim}
TRANSCRIPTION: UP UPSTATE NEW YORK SOMEWHERE UH     OVER OVER HUGE AREAS 
HYPOTHESIS:       UPSTATE NEW YORK SOMEWHERE UH ALL ALL  THE  HUGE AREAS 
               1  0       0   0    0         0  1   1    1    0    0    
:4 errors per 10 words in transcription; WER = 40%
\end{verbatim}

\paragraph{Perplexity}

As an alternative to the computationally expensive word error rate
(WER), a statistical language model is evaluated by how well it
predicts a string of symbols $W_t$ --- commonly referred to as
\emph{test data} --- generated by the source to be modeled.

Assume we compare two models $M_1$ and $M_2$; they assign probability
$P_{M_1}(W_t)$ and $P_{M_2}(W_t)$, respectively, to the sample test
string $W_t$. The test string has neither been used nor seen at the
estimation step of either model and it was generated by the same
source that we are trying to model. ``Naturally'', we consider $M_1$
to be a better model than $M_2$ if $P_{M_1}(W_t) > P_{M_2}(W_t)$. 

A commonly used quality measure for a given model $M$ is related to
the entropy of the underlying source and was introduced under
the name of perplexity (PPL)~\cite{jelinek97}:
\begin{eqnarray}
PPL(M) = exp(-1/N \sum_{k=1}^{N}\ln{[P_M(w_k|W_{k-1})]}) 
\end{eqnarray}

\subsection*{Thesis Layout}

The thesis is organized as follows:

After a brief introduction to language modeling for speech
recognition, Chapter~\ref{chapter:slm} gives a basic description of
the structured language model (SLM) followed by
Chapters~\ref{chapter:EM_ML}~and~\ref{chapter:model_reest}
explaining the model parameters reestimation algorithm we used.
Chapter~\ref{chapter:slm_experiments} presents a series of experiments
we have carried out on the UPenn Treebank corpus~(\cite{Upenn}).

Chapters~\ref{chapter:a_star_lattice_decoder}~and~\ref{chapter:lattice_decoding_experiments}
describe the setup and speech recognition experiments using the
structured language model on different corpora: Wall Street Journal
(WSJ, \cite{wsj0}), Switchboard (SWB, \cite{SWB}) and Broadcast News
(BN).

We conclude with Chapter~\ref{chapter:conclusions}, outlining the
relationship between our approach to language modeling --- and parsing
--- and others in the literature and pointing out what we believe to
be worthwhile future directions of research.

A few appendices detail mathematical aspects of the reestimation
technique we have used.
%%% Local Variables: 
%%% mode: latex
%%% TeX-master: "main"
%%% TeX-master: "main"
%%% TeX-master: "main"
%%% TeX-master: "main"
%%% End: 

\chapter{Language Modeling for Speech Recognition}

The task of a speech recognizer is to automatically transcribe speech
into text. Given a string of acoustic features $A$ extracted by its signal
processing front-end from the raw acoustic
waveform, the speech recognizer tries to identify the
word sequence $W$ that produced $A$ --- typically one sentence at a time. Let
$\hat{W}$ be the word string --- hypothesis --- output by the speech
recognizer. The measure of success is the word error rate; to
calculate it we need to first find the most
favorable word alignment between $\hat{W}$ and $W$ --- assumed to be
known a priori for evaluation purposes only ---  and then count the
number of incorrect words in the hypothesized sequence $\hat{W}$  per
total number of words in $W$.
\begin{verbatim}
TRANSCRIPTION: UP UPSTATE NEW YORK SOMEWHERE UH     OVER OVER HUGE AREAS 
HYPOTHESIS:       UPSTATE NEW YORK SOMEWHERE UH ALL ALL  THE  HUGE AREAS 
               1  0       0   0    0         0  1   1    1    0    0    
:4 errors per 10 words in transcription; WER = 40%
\end{verbatim}

The most successful approach to speech recognition so far is a
statistical one pioneered by Jelinek and his
colleagues~\cite{jelinek83}; speech recognition is viewed as a
Bayes decision problem: given the observed string of acoustic
features $A$, find the most likely word string $\hat{W}$ among those that could
have generated $A$:
\begin{eqnarray}
  \hat{W} = argmax_{W} P(W|A) = argmax_{W} P(A|W)\cdot P(W) \label{intro:bayes}
\end{eqnarray}

There are three broad subproblems to be solved:
\begin{itemize}
\item  decide on a feature extraction algorithm and model the channel
  probability $P(A|W)$ --- commonly referred to as \emph{acoustic modeling};
\item  model the source probability $P(W)$ --- commonly referred to as \emph{language modeling};
\item  \emph{search} over all possible word strings $W$ that could have given rise
  to $A$ and find out the most likely one $\hat{W}$; due to the large
  vocabulary size --- tens of thousands of words --- an exhaustive
  search is intractable.
\end{itemize}

The remaining part of the chapter is organized as follows: we will first describe 
language modeling in more detail by taking a source modeling
view; then we will describe current approaches to the problem,
outlining their advantages and shortcomings.

%Section{basic_lm}
%language modeling as source modeling; 
%measure of success: perplexity;
%problems: huge vocabulary(words treated as
%          integers), need for smoothing;
\section{Basic Language Modeling}

As explained in the introductory section, the language modeling
problem is to estimate the source probability $P(W)$ where $W =
w_1,w_2,\ldots,w_n$ is a sequence of words. 

This probability is estimated from a training corpus --- thousands of
words of text --- according to a \emph{modeling assumption} on the
source that generated the text. Usually the source model is
parameterized according to a set of parameters $P_{\theta}(W), \theta
\in \Theta$ where $\Theta$ is referred to as the parameter space. 
 
One first choice faced by the modeler is the alphabet $\mathcal{V}$
--- also called vocabulary --- in which the $w_i$ symbols take
value. For practical purposes one has to limit the size of the vocabulary.
A common choice is to use a finite set of words $\mathcal{V}$ and map
any word not in this set to the distinguished type \verb+<unknown>+.

A second, and much more important choice is the source model to be
used. A desirable way of making this choice takes into account:
\begin{itemize}
\item  a priori knowledge of how the source might work, if available;
\item  possibility to reliably estimate source model parameters; reliability
of estimates limits the number and type of parameters one can
estimate given a certain amount of training data;
\item  preferably, due to the sequential nature of an efficient
search algorithm, the model should operate left-to-right, allowing
the computation of\\ $P(w_1,w_2,\ldots,w_n) =
P(w_1) \cdot \prod_{i=2}^n P(w_i|w_1 \ldots w_{i-1})$.
\end{itemize}

We thus seek to develop parametric conditional models:
\begin{eqnarray}
  P_{\theta}(w_i|w_1 \ldots w_{i-1}), \theta \in \Theta, w_i \in \mathcal{V}
\end{eqnarray}
The currently most successful model assumes a Markov source of a given 
order $n$ leading to the \emph{n-gram language model}:
\begin{eqnarray}
  P_{\theta}(w_i|w_{1} \ldots w_{i-1}) = P_{\theta}(w_i|w_{i-n+1} \ldots w_{i-1}) \label{basic_lm:n-gram}
\end{eqnarray}

\subsection{Language Model Quality}

Any parameter estimation algorithm needs an objective function with
respect to which the parameters are optimized. As stated in the
introductory section, the ultimate goal of a speech recognizer is low
word error rate (WER). However, all attempts to derive an algorithm
that would directly estimate the model parameters so as to minimize WER have
failed. As an alternative, a statistical model is evaluated by
how well it predicts a string of symbols $W_t$ --- commonly referred
to as \emph{test data} --- generated by the source to be modeled.

\subsection{Perplexity}

Assume we compare two models $M_1$ and $M_2$; they assign probability
$P_{M_1}(W_t)$ and $P_{M_2}(W_t)$, respectively, to the sample test
string $W_t$. The test string has neither been used nor seen at the
estimation step of either model and it was generated by the same
source that we are trying to model. ``Naturally'', we consider $M_1$
to be a better model than $M_2$ if $P_{M_1}(W_t) > P_{M_2}(W_t)$. It
is worth mentioning that this is different than maximum likelihood
estimation: the test data is not seen during the model estimation
process and thus we cannot directly estimate the parameters of the
model such that it assigns maximum probability to the test string.

A commonly used quality measure for a given model $M$ is related to
the entropy of the underlying source and was introduced under
the name of perplexity (PPL)~\cite{jelinek97}:
\begin{eqnarray}
  PPL(M) = exp(-1/N \sum_{i=1}^{N}\ln{[P_M(w_i|w_1 \ldots w_{i-1})]}) \label{basic_lm:ppl}
\end{eqnarray}
It is easily seen that if our model estimates the source probability
exactly:\\ $P_M(w_i|w_1 \ldots w_{i-1}) = P_{source}(w_i|w_1 \ldots w_{i-1}), i=1 \ldots N$\\
then~(\ref{basic_lm:ppl}) is a consistent estimate of the
exponentiated source entropy $exp(H_{source})$.
To get an intuitive understanding for PPL~(\ref{basic_lm:ppl}) we can state that it measures
the average surprise of model $M$ when it predicts the next word $w_i$ in
the current context $w_1 \ldots w_{i-1}$.

\subsubsection{Smoothing}

One important remark is worthwhile at this point: assume that our model $M$
is faced with the prediction $w_i|w_1 \ldots w_{i-1}$ and that $w_i$
has not been seen in the training corpus in context $w_1 \ldots
w_{i-1}$ which itself possibly has not been encountered in the training corpus. 
If $P_M(w_i|w_1 \ldots w_{i-1}) = 0$ then $P_M(w_1 \ldots w_N) = 0$ thus forcing a recognition error;
good models $M$ are smooth, in the sense that\\ $\exists
\epsilon(M) > 0$~s.t.~$ P_M(w_i|w_1 \ldots w_{i-1}) > \epsilon,
\forall w_i \in \mathcal{V}$, $(w_1 \ldots w_{i-1}) \in {\mathcal{V}}^{i-1}$.

%Section{current_approaches}
%describe n-gram models as Markov sources;
%describe deleted interpolation as a smoothing method;
\section{Current Approaches}

In the previous section we introduced the class of n-gram models.
They assume a Markov source of order $n$, thus making the following equivalence
classification of a given context:
\begin{eqnarray}
  [w_{1} \ldots w_{i-1}] = w_{i-n+1} \ldots w_{i-1} = h_n
\end{eqnarray}
An equivalence classification of some similar sort is needed because
of the impossibility to get reliable relative frequency estimates for
the full context prediction $w_i|w_{1} \ldots w_{i-1}$. Indeed, as
shown in~\cite{rosenfeld94}, for a 3-gram model the coverage for the
$(w_i|w_{i-2},w_{i-1})$ events is far from sufficient: the rate of new
(unseen) trigrams in test data relative to those observed in a training
corpus of size 38 million words is 21\% for a 5,000-words vocabulary
and 32\% for a 20,000-words vocabulary. Moreover, approx. 70\% of the
trigrams in the training data have been seen once, thus making a
relative frequency estimate unusable because of its unreliability.

One standard approach that also ensures smoothing is the deleted
interpolation method~\cite{jelinek80}. It interpolates
linearly  among contexts of different order $h_n$:
\begin{eqnarray}
  P_{\theta}(w_i|w_{i-n+1} \ldots w_{i-1}) & = & \sum_{k=0}^{k=n} \lambda_k \cdot f(w_i|h_k)
\end{eqnarray}
where:
\begin{itemize}
\item  $h_k = w_{i-k+1} \ldots w_{i-1}$ is the context of order $k$
when predicting $w_i$;
\item  $f(w_i|h_k)$ is the relative frequency estimate for the
conditional probability $P(w_i|h_k)$;
\begin{eqnarray*}
  f(w_i|h_k) & = & C(w_i,h_k)/C(h_k),\\ 
  C(h_k)     & = & \sum_{w_{i} \in \mathcal{V}} C(w_i,h_k), k = 1 \ldots n, \\
  f(w_i|h_1) & = & C(w_i)/\sum_{w_i \in \mathcal{V}}C(w_i), \\
  f(w_i|h_{0}) & = & 1/|{\mathcal V}|, \forall w_i \in \mathcal{V},\ uniform;
\end{eqnarray*}
\item  $\lambda_k, k = 0 \ldots n$ are the interpolation
  coefficients satisfying $\lambda_k > 0,  k = 0 \ldots n$ and
  $\sum_{k=0}^{k=n} \lambda_k = 1$.
\end{itemize}

The model parameters $\theta$ are:
\begin{itemize}
\item  the counts $C(h_n,w_i)$; lower order counts are inferred
recursively by: \\ $C(h_k,w_i) = \sum_{w_{i-k} \in \mathcal{V}} C(w_{i-k},h_k,w_i)$;
\item  the interpolation coefficients $\lambda_k, k = 0 \ldots n$.
\end{itemize}

A simple way to estimate the model parameters involves a two stage process:
\begin{enumerate}
  \item gather counts from \emph{development data} --- about 90\% of
    training data;
  \item estimate interpolation coefficients to minimize the perplexity
    of \emph{cross-validation data} --- the remaining 10\% of the training data --- using the
    expectation-maximization (EM) algorithm~\cite{em77}.
\end{enumerate}

Other approaches use different smoothing techniques --- maximum
entropy~\cite{berger:max_ent}, back-off~\cite{katz:back_off} --- but
they all share the same Markov assumption on the underlying source. 

An attempt to overcome this limitation is developed in~\cite{rosenfeld94}.
Words in the context outside the range of the 3-gram model are
identified as ``triggers'' and retained together with the ``target'' word 
in the predicted position. The (trigger, target) pairs are treated as
complementary sources of information and integrated with the n-gram
predictors using the maximum entropy method. 
The method has proven successful, however computationally burdensome.

Our attempt will make use of the hierarchical structuring of word strings in
natural language for expanding the memory length of the source. 

\chapter{A Structured Language Model} \label{chapter:slm}

It has been long argued in the linguistics community that the simple
minded Markov assumption is far from accurate for modeling the natural
language source. However so far very few approaches managed to outperform
the n-gram model in perplexity or word error rate, none of them
exploiting syntactic structure for better modeling of the natural
language source.

The model we present is closely related to the one
investigated in~\cite{ws96}, however different in a few important
aspects:
\begin{itemize}
\item our model operates in a left-to-right manner, thus allowing its
  use directly in the hypothesis search for $\hat{W}$ in (\ref{intro:bayes});
\item our model is a factored version of the one in~\cite{ws96}, thus enabling the
  calculation of the joint probability of words and parse structure;
  this was not possible in the previous case due to the huge
  computational complexity of that model;
\item our model assigns probability at the word level, being a proper
  language model.
\end{itemize}

\section{Syntactic Structure in Natural Language}

Although not complete, there is a certain agreement in the linguistics
community as to what constitutes syntactic structure in natural
language. In an effort to provide the computational linguistics
community with a database that reflects the current basic level of
agreement, a treebank was developed at the University of Pennsylvania,
known as the UPenn Treebank~\cite{Upenn}. The treebank contains
sentences which were manually annotated with syntactic structure.
\begin{figure}
  \begin{center} 
\begin{verbatim}
( (S 
    (NP-SBJ 
      (NP (NNP Pierre) (NNP Vinken) )
      (, ,) 
      (ADJP 
        (NP (CD 61) (NNS years) )
        (JJ old) )
      (, ,) )
    (VP (MD will) 
      (VP (VB join) 
        (NP (DT the) (NN board) )
        (PP-CLR (IN as) 
          (NP (DT a) (JJ nonexecutive) (NN director) ))
        (NP-TMP (NNP Nov.) (CD 29) )))
    (. .) ))
\end{verbatim}
  \end{center}
  \caption{UPenn Treebank Parse Tree Representation} \label{fig:UPenn_parse_tree}
\end{figure}
A sample parse tree from the treebank is shown in Figure~\ref{fig:UPenn_parse_tree}.
Each word bears a \emph{part of speech tag} (POS tag): e.g.\ Pierre is
annotated as being a proper noun (NNP). Round brackets are used to
mark constituents, each constituent being tagged with a
\emph{non-terminal label} (NT label): e.g.\ 
\verb+(NP (NNP Pierre) (NNP Vinken) )+ is marked as noun phrase
(NP). Some non-terminal labels are enriched with additional information 
which is usually discarded as a first approximation: e.g.\
\verb+NP-TMP+ becomes \verb+NP+. The task of recovering the parsing
structure with POS/NT annotation for a given word sequence (sentence)
is referred to as \emph{automatic parsing} of natural language (or
simply parsing). A sub-task whose aim is to recover the part of speech
tags for a given word sequence is referred to as \emph{POS-tagging}.

This effort fostered research in automatic part-of-speech tagging and
parsing of natural language, providing a base for developing and
testing algorithms that try to describe computationally the
constraints in natural language.

State of the art parsing and POS-tagging technology developed in the
computational linguistics community operates at the sentence level. 
Statistical approaches employ conditional probabilistic models
$P(T/W)$ where $W$ denotes the  sentence to be parsed and $T$ is the
hidden parse structure or POS tag sequence. Due to the left-to-right
constraint imposed by the speech recognizer on the language model
operation, we will  be forced to develop syntactic structure for
sentence \emph{prefixes}. This is just one of the limitations imposed
by the fact that we aim at incorporating the language model in a
speech recognizer. Information that is present in written text but
silent in speech --- such as case information (Pierre vs.\ pierre ) and
punctuation --- will not be used by our model either. 

The use of headwords has become standard in the computational
linguistics community: the \emph{headword} of a phrase is the word that best
represents the phrase, all the other words in the phrase being
modifiers of the headword. 
For example we refer to \verb+years+ as the \emph{headword} of the
phrase \verb+(NP (CD 61) (NNS years) )+. The lexicalization ---
headword percolation --- of the treebank has proven extremely useful
in increasing the accuracy of automatic parsers. 

There are ongoing arguments about the adequacy of the tree
representation for syntactic dependencies in natural language. One
argument debates the usage of binary branching --- in which one word
modifies exactly one other word in the same sentence --- versus
trees with unconstrained branching. Learnability issues favor the former,
as argued in~\cite{binary_structure_learnability}. It is not surprising that 
the binary structure also lends itself to a simpler algorithmic
description and is the choice for our modeling approach.

\begin{figure}
  \begin{center} 
    \epsfig{file=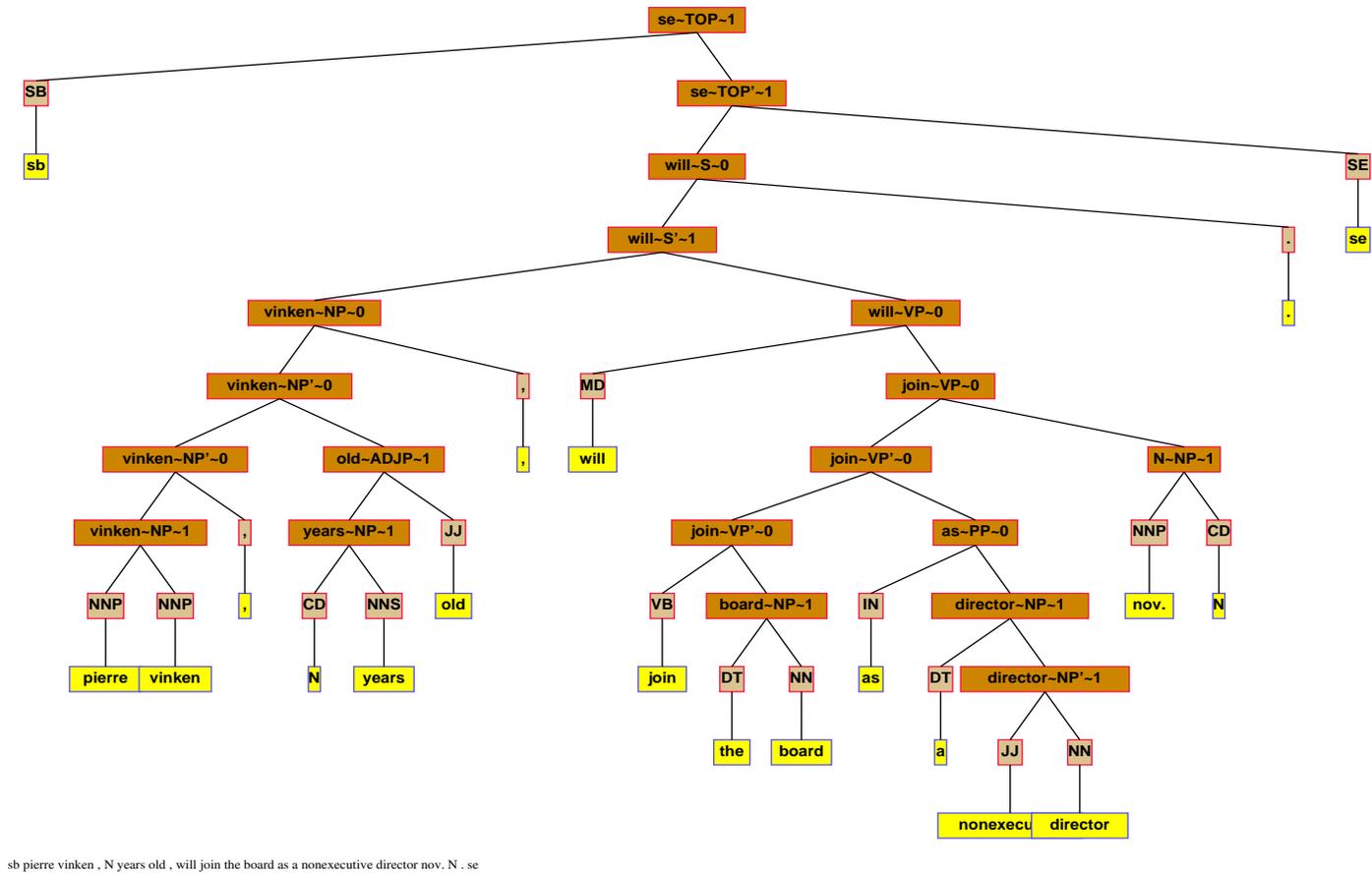, width=12cm, height=20cm}%, angle=270}
  \end{center}
  \caption{Parse Tree Representation after Headword Percolation and Binarization} \label{fig:CNF_parse_tree}
\end{figure}

As an example, the output of the headword percolation and binarization 
procedure for the parse tree in Figure~\ref{fig:UPenn_parse_tree} is
presented in Figure~\ref{fig:CNF_parse_tree}. The headwords are now
percolated at each intermediate node in the tree; the additional bit
--- value 0 or 1 --- indicates the origin of the headword in each
constituent.

\subsection{Headword Percolation and Binarization} \label{section:headword_percolation}

In order to obtain training data for our model we need
to binarize the UPenn~Treebank~\cite{Upenn} parse trees and percolate
headwords. The procedure we used was to first percolate headwords
using a context-free (CF) rule-based approach and then binarize the parses
by again using a rule-based approach.

\subsubsection{Headword Percolation}

Inherently a heuristic process, we were satisfied with the output of
an enhanced version of the  procedure described in~\cite{mike96} ---
also known under the name ``Magerman \& Black Headword Percolation Rules''.

The procedure first decomposes a parse tree from the treebank into its context-free
constituents, identified solely by the non-terminal/POS labels.
Within each constituent we then identify the headword position and
then, in a recursive third step, we fill in the headword position
with the actual word percolated up from the leaves of the tree.

The headword percolation procedure is based on rules for
identifying the headword position within each constituent. They are
presented in table~\ref{table:headw_rules}.
\begin{table}
  \begin{verbatim}
TOP     right   _SE _SB 
ADJP    right   <~QP|_JJ|_VBN|~ADJP|_$|_JJR> 
                <^~PP|~S|~SBAR|_.|_,|_''|_``|_`|_'|_:|_LRB|_RRB> 
ADVP    right   <_RBR|_RB|_TO|~ADVP> 
                <^~PP|~S|~SBAR|_.|_,|_''|_``|_`|_'|_:|_LRB|_RRB>        
CONJP   left    _RB <^_.|_,|_''|_``|_`|_'|_:|_LRB|_RRB>                                   
FRAG    left    <^_.|_,|_''|_``|_`|_'|_:|_LRB|_RRB>
INTJ    left     <^_.|_,|_''|_``|_`|_'|_:|_LRB|_RRB>
LST     left    _LS <^_.|_,|_''|_``|_`|_'|_:|_LRB|_RRB>
NAC     right   <_NNP|_NNPS|~NP|_NN|_NNS|~NX|_CD|~QP|_VBG> 
                <^_.|_,|_''|_``|_`|_'|_:|_LRB|_RRB>
NP      right   <_NNP|_NNPS|~NP|_NN|_NNS|~NX|_CD|~QP|_PRP|_VBG> 
                <^_.|_,|_''|_``|_`|_'|_:|_LRB|_RRB>
NX      right   <_NNP|_NNPS|~NP|_NN|_NNS|~NX|_CD|~QP|_VBG> 
                <^_.|_,|_''|_``|_`|_'|_:|_LRB|_RRB>
PP      left    _IN _TO _VBG _VBN ~PP  
                <^_.|_,|_''|_``|_`|_'|_:|_LRB|_RRB>                        
PRN     left    ~NP ~PP ~SBAR ~ADVP ~SINV ~S ~VP   
                <^_.|_,|_''|_``|_`|_'|_:|_LRB|_RRB>              
PRT     left    _RP <^_.|_,|_''|_``|_`|_'|_:|_LRB|_RRB>  
QP      left    <_CD|~QP> <_NNP|_NNPS|~NP|_NN|_NNS|~NX> <_DT|_PDT> 
                <_JJR|_JJ> <^_CC|_.|_,|_''|_``|_`|_'|_:|_LRB|_RRB>
RRC     left    ~ADJP ~PP ~VP <^_.|_,|_''|_``|_`|_'|_:|_LRB|_RRB>
S       right   ~VP <~SBAR|~SBARQ|~S|~SQ|~SINV> 
                <^_.|_,|_''|_``|_`|_'|_:|_LRB|_RRB> 
SBAR    right   <~S|~SBAR|~SBARQ|~SQ|~SINV> 
                <^_.|_,|_''|_``|_`|_'|_:|_LRB|_RRB>                         
SBARQ   right   ~SQ ~S ~SINV ~SBAR <^_.|_,|_''|_``|_`|_'|_:|_LRB|_RRB>
SINV    right   <~VP|_VBD|_VBN|_MD|_VBZ|_VB|_VBG|_VBP> ~S ~SINV 
                <^_.|_,|_''|_``|_`|_'|_:|_LRB|_RRB>
SQ      left    <_VBD|_VBN|_MD|_VBZ|_VB|~VP|_VBG|_VBP> 
                <^_.|_,|_''|_``|_`|_'|_:|_LRB|_RRB>
UCP     left    <^_.|_,|_''|_``|_`|_'|_:|_LRB|_RRB>
VP      left    <_VBD|_VBN|_MD|_VBZ|_VB|~VP|_VBG|_VBP> 
                <^_.|_,|_''|_``|_`|_'|_:|_LRB|_RRB>
WHADJP  right   <^_.|_,|_''|_``|_`|_'|_:|_LRB|_RRB>
WHADVP  right   _WRB   <^_.|_,|_''|_``|_`|_'|_:|_LRB|_RRB>                                    
WHNP    right   _WP _WDT _JJ _WP$ ~WHNP 
                <^_.|_,|_''|_``|_`|_'|_:|_LRB|_RRB>  
WHPP    left    _IN <^_.|_,|_''|_``|_`|_'|_:|_LRB|_RRB>                                    
X       right   <^_.|_,|_''|_``|_`|_'|_:|_LRB|_RRB>
\end{verbatim}
  \caption{Headword Percolation Rules} \label{table:headw_rules}
\end{table}

Let $Z \rightarrow Y_1 \ldots Y_n$ be one of the context-free (CF)
rules that make up a given parse. We identify the headword position as
follows: 
\begin{itemize}
\item identify in the first column of the table the entry that
  corresponds to the $Z$ non-terminal label;
  
\item search $Y_1 \ldots Y_n$ from either left or right, as indicated in the second column of
  the entry, for the $Y_i$ label that matches the regular
  expressions listed in the entry; the first matching $Y_i$ is going
  to be the headword of the $(Z$ $(Y_1 \ldots) \ldots (Y_n \ldots) )$
  constituent; the regular expressions listed in one entry are ranked
  in left to right order: first we try to match the first one, if
  unsuccessful we try the second one and so on.
\end{itemize}

A regular expression of the type \verb+<_CD|~QP>+ matches any of the
constituents listed between angular parentheses. For example, the
\verb+<^_.|_,|_''|_``|_`|_'|_:|_LRB|_RRB>+ %''
regular expression will match any constituent that is \emph{not} ---
list begins with  \verb+<^+ --- among any of the elements in the list
between \verb+<^+ and \verb+>+, in this case any constituent which is
not a punctuation mark. The terminal labels have \verb+_+ prepended to
them --- as in \verb+_CD+ --- the non-terminal labels have the
\verb+~+ prefix --- as in \verb+~QP+; \verb+|+ is merely a
separator in the list.

\subsubsection{Binarization}

\begin{figure}
  \begin{center} 
    \epsfig{file=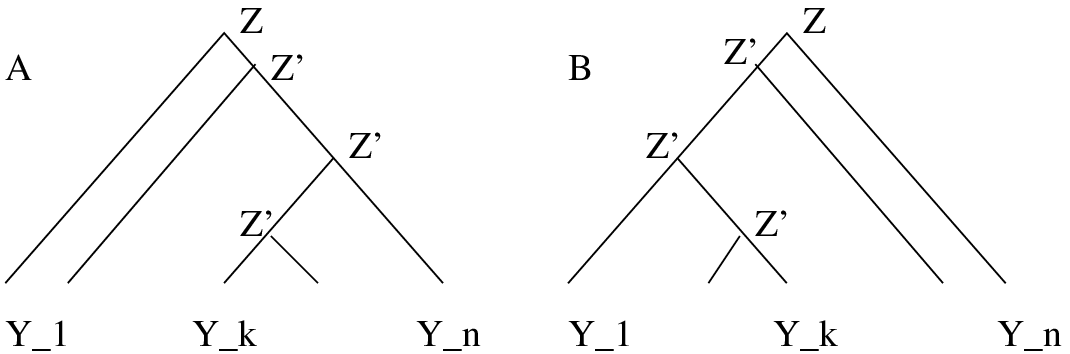,height=2cm,width=6cm}
  \end{center}
  \caption{Binarization schemes} \label{fig:bin_schemes}
\end{figure}

Once the position of the headword within a constituent
--- equivalent with a CF production of the type 
$Z \rightarrow Y_1 \ldots Y_n$ , where $Z, Y_1, \ldots Y_n$ are
non-terminal labels or POStags (only for $Y_i$) ---  is identified to be $k$, we binarize
the constituent as follows: depending on the $Z$ identity, a fixed rule is used to decide
which of the two binarization schemes in Figure~\ref{fig:bin_schemes} to apply.
The intermediate nodes created by the above binarization schemes
receive the non-terminal label $Z'$. 

The choice among the two schemes is made according to the list of
rules presented in table~\ref{table:binarization_rules}, based on the
identity of the label on the left-hand-side of a CF rewrite rule.

\begin{table}
  \begin{verbatim}
## first column : constituent label
## second column: binarization type : A or B
## A means right modifiers go first, left branching, then left
##         modifiers are attached via right branching
## B means left modifiers go first, right branching, then right
##         modifiers are attached via left branching
TOP     A
ADJP    B
ADVP    B
CONJP   A
FRAG    A
INTJ    A
LST     A
NAC     B
NP      B
NX      B
PP      A
PRN     A
PRT     A
QP      A
RRC     A
S       B
SBAR    B
SBARQ   B
SINV    B
SQ      A
UCP     A
VP      A
WHADJP  B
WHADVP  B
WHNP    B
WHPP    A
X       B
\end{verbatim}

  \caption{Binarization Rules} \label{table:binarization_rules}
\end{table}

Notice that whenever $k = 1$ or $k = n$ --- a case which is very
frequent --- the two schemes presented above yield the same binary structure.

Another problem when binarizing the parse trees is the presence of unary
productions. Our model allows only unary productions of the type $Z \rightarrow Y$
where $Z$ is a non-terminal label and $Y$ is a POS tag. The unary
productions $Z \rightarrow Y$ where both $Z$ and $Y$ are non-terminal
labels were deleted from the treebank, only the $Z$ constituent being retained:
\verb+(Z (Y (.) (.)))+ becomes \verb+(Z (.) (.))+.

\section{Exploiting Syntactic Structure for Language Modeling} \label{section:basic_idea}

Consider predicting the word \verb+after+ in the sentence:\\
\verb+the contract ended with a loss of 7 cents+ \\ \verb+after trading as low as 89 cents+.\\
A 3-gram approach would predict \verb+after+ from \verb+(7, cents)+
whereas it is intuitively clear that the strongest predictor would be
\verb+contract ended+ which is outside the reach of even 7-grams. What
would enable us to identify the predictors in the sentence prefix?

\begin{figure}[ha]
  \begin{center} 
    \epsfig{file=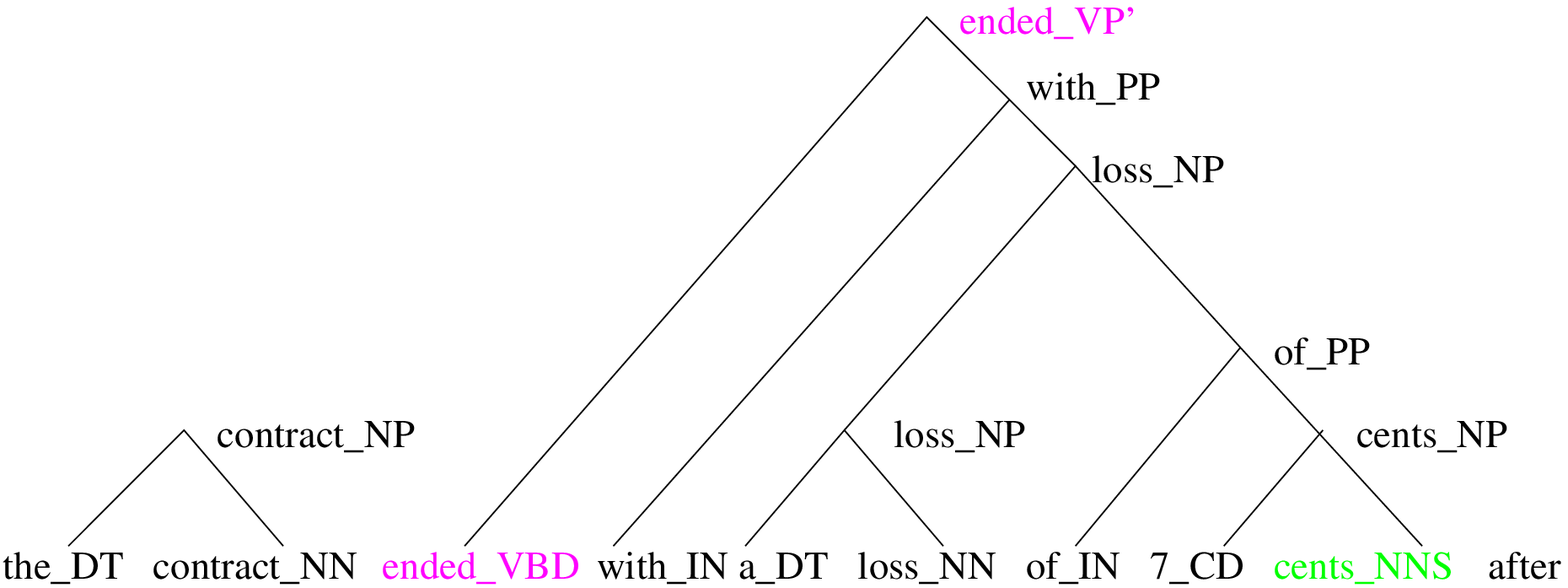,width=10cm}
  \end{center}
  \caption{Partial parse} \label{fig:p_parse}
\end{figure}

The linguistically correct \emph{partial parse} of
the sentence prefix when predicting \verb+after+ is shown in
Figure~\ref{fig:p_parse}. 
The word \verb+ended+ is called the \emph{headword} of the
\emph{constituent} \verb+(ended (with (...)))+
and \verb+ended+ is an \emph{exposed headword} when predicting
\verb+after+ --- topmost headword in the largest constituent that
contains it. 
Our working hypothesis is that the syntactic structure filters out
irrelevant words and points to the important ones, thus enabling the
use of information in the more distant past when predicting the next word. We
will attempt to model this using the concept of 
\emph{exposed headwords} introduced before.

We will give two heuristic arguments that justify the use of exposed
headwords:
\begin{itemize}
\item the 3-gram context for predicting \verb+after+ --- \verb+(7, cents)+ ---
  is intuitively less satisfying than using the two most recent
  \emph{exposed headwords} \verb+(contract, ended)+ --- identified by
  the parse tree;
\item the headword context does not change if we  remove the 
  \verb+(of (7 cents))+ constituent --- the resulting sentence is still a valid
  one --- whereas the 3-gram context becomes  \verb+(a, loss)+.
\end{itemize}

The preliminary experiments reported in~\cite{chelba97} --- although the perplexity
results are \emph{conditioned} on parse structure developed by human annotators
by having the entire sentence at their disposal --- showed the usefulness of
headwords accompanied by non-terminal labels for making a better
prediction of the word following a given sentence prefix.

Our model will attempt to build the syntactic structure incrementally
while traversing the sentence left-to-right.
The word string $W$ can be observed whereas the parse structure with headword and POS/NT label 
annotation --- denoted by $T$ --- remains \emph{hidden}. The model will
assign a probability $P(W,T)$ to every sentence $W$ with every
possible POStag assignment, binary branching parse, non-terminal label 
and headword annotation for every constituent of $T$. 

\begin{figure}[ha]
  \begin{center}
    \epsfig{file=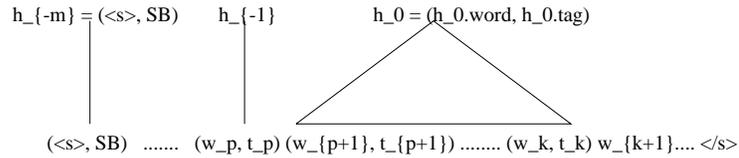,width=10cm}
  \end{center}
  \caption{A word-parse k-prefix} \label{fig:w_parse}
\end{figure}

\begin{figure}[ha]
  \begin{center} 
    \epsfig{file=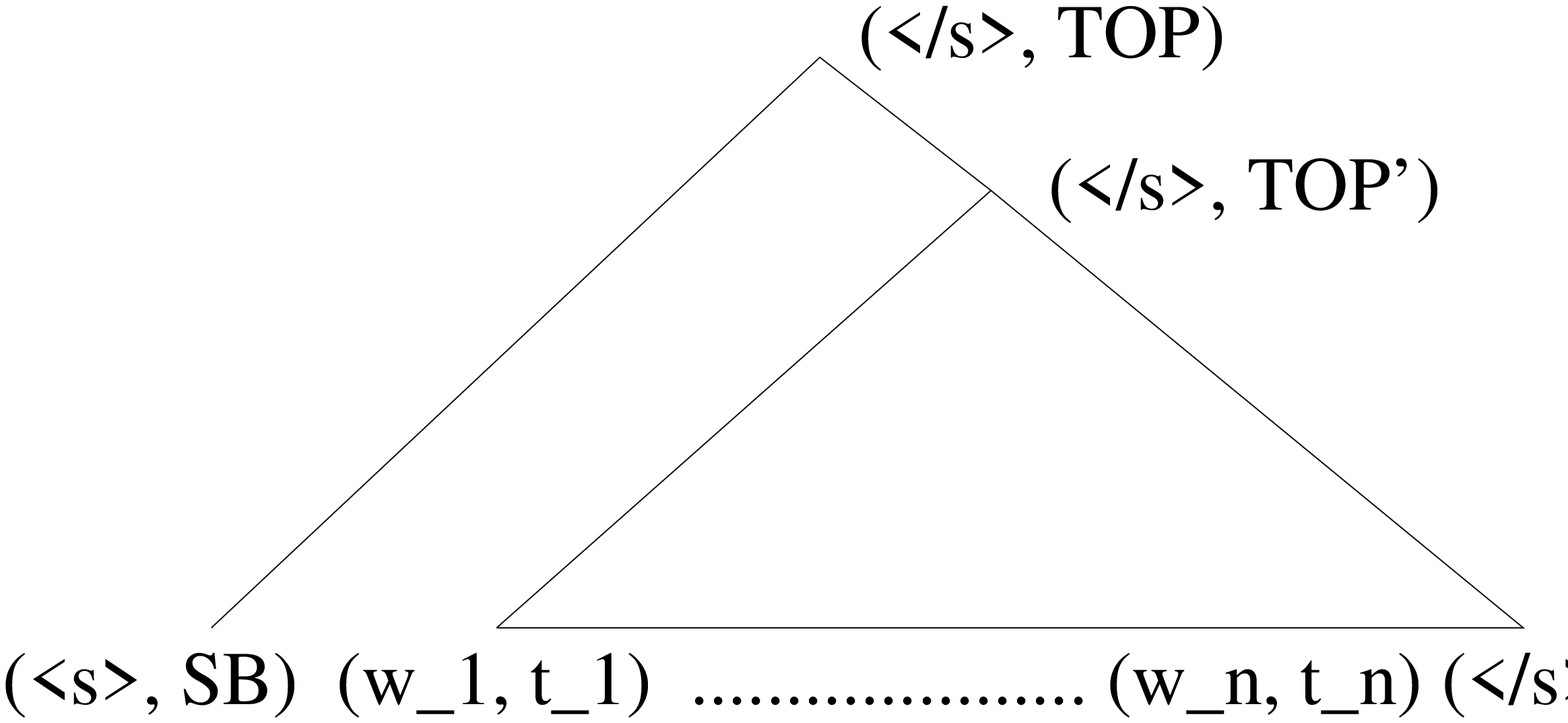,width=6cm}
  \end{center}
  \caption{Complete parse} \label{fig:c_parse}
\end{figure}

Let $W$ be a sentence of length $n$ words to which we have prepended
\verb+<s>+ and appended \verb+</s>+ so that $w_0 = $\verb+<s>+ and
$w_{n+1} = $\verb+</s>+.
Let $W_k$ be the word k-prefix $w_0 \ldots w_k$ of the sentence and 
\mbox{$W_k T_k$} the \emph{word-parse k-prefix}. To stress this point, a
\mbox{word-parse k-prefix} contains --- for a given parse --- those
and only those binary subtrees whose span is completely included in the word k-prefix, excluding 
$w_0 = $\verb+<s>+. Single words along with their POStag can be
regarded as root-only trees. Figure~\ref{fig:w_parse} shows a
word-parse k-prefix; \verb|h_0 .. h_{-m}| are the \emph{exposed
 heads}, each head being a pair(headword, non-terminal label), or
(word,  POStag) in the case of a root-only tree. 
A \emph{complete parse} --- Figure~\ref{fig:c_parse} --- is defined as 
a binary parse of the $(w_1,t_1) \ldots (w_n,t_n)$\verb+ (</s>,SE)+
\footnote{SB is a distinguished POStag for the sentence
  beginning symbol <s>; SE is a distinguished POStag for the sentence
  end symbol </s>;}
sequence with the restriction that \verb+(</s>, TOP')+ is the only
allowed head. 
Note that \mbox{$((w_1,t_1) \ldots (w_n,t_n))$} \emph{needn't} be a constituent,
but for the parses where it is, there is no a priori restriction on which of
its words is the headword or what is the non-terminal label that
accompanies the headword. This is one other notable difference
between our model and the traditional ones developed in the
computational linguistics community imposed by the bottom-up
operation of the model. The manually annotated trees in the treebank
(see Figure~\ref{fig:CNF_parse_tree}) have all the words in a sentence
as one single constituent bearing a restricted set of non-terminal
labels: the sentence \mbox{$(S (w_1,t_1) \ldots (w_n,t_n))$} is a
constituent labeled with \verb+S+.

As it can be observed the UPenn treebank -style trees are a subset
of the family of trees allowed by our parameterization, making a direct 
comparison between our model and state of the art parsing techniques
--- which insist on generating UPenn treebank -style parses  --- less
meaningful.

The model will operate by means of three modules:
\begin{itemize}
\item WORD-PREDICTOR predicts the next word $w_{k+1}$ given the
  word-parse k-prefix \mbox{$W_k T_k$} and then passes control to the TAGGER;
\item TAGGER predicts the POStag $t_{k+1}$ of the next word given the
  word-parse k-prefix and the newly predicted word $w_{k+1}$ and then passes
  control to the PARSER;
\item PARSER grows the already existing binary branching structure by
  repeatedly generating transitions from the following set:\\ \verb+(unary, NTlabel)+,
  \verb+(adjoin-left, NTlabel)+ or  \verb+(adjoin-right, NTlabel)+
  until it passes control to the PREDICTOR
  by  taking a \verb+null+ transition. \verb+NTlabel+ is the non-terminal
  label assigned to the newly built constituent and
  \verb+{left,right}+ specifies where the new headword is percolated from.
\end{itemize}

\begin{figure}[h]
  \begin{center} 
    \epsfig{file=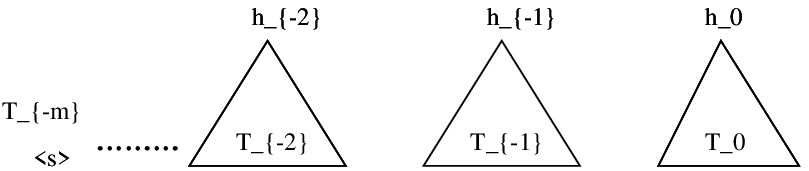,width=8cm}
    \caption{Before an adjoin operation} \label{fig:before}  
    \vspace{0.1cm}
    \epsfig{file=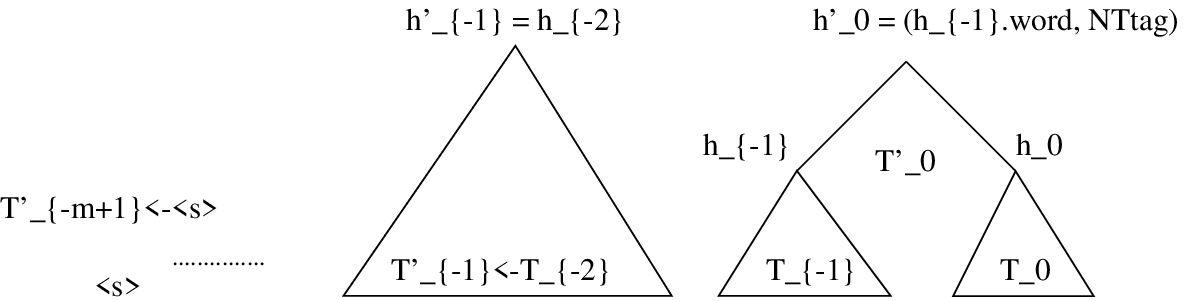,width=8cm}
    \caption{Result of adjoin-left under NTtag} \label{fig:after_a_l}
    \vspace{0.1cm}
    \epsfig{file=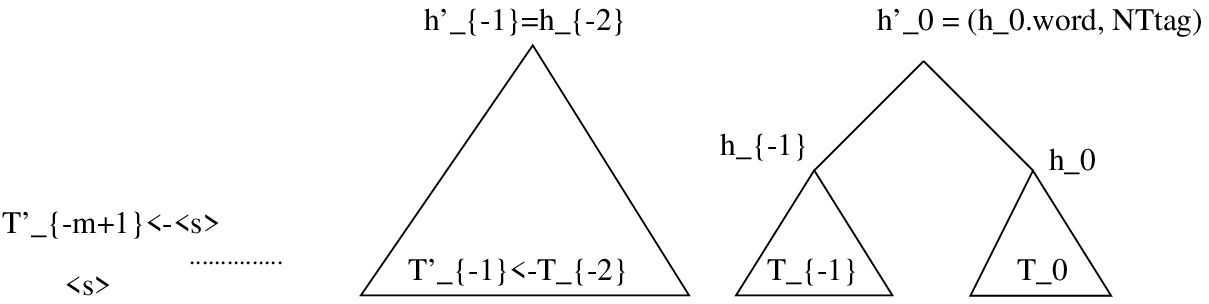,width=8cm}
    \caption{Result of adjoin-right under NTtag} \label{fig:after_a_r}
  \end{center}
\end{figure}

The operations performed by the PARSER are illustrated in
Figures~\ref{fig:before}-\ref{fig:after_a_r} and they ensure that all possible binary
branching parses with all possible headword and non-terminal label
assignments for the $w_1 \ldots w_k$ word sequence can be generated.
Algorithm~\ref{algorithm:slm_operation} at the end of this chapter
formalizes the above description of the sequential generation of a
sentence with a complete parse.
\begin{algorithm}[h]
\begin{verbatim}
Transition t;         // a PARSER transition
predict (<s>, SB);
do{
  //WORD-PREDICTOR and TAGGER
  predict (next_word, POStag);   
  //PARSER
  do{                            
    if(h_{-1}.word != <s>){
      if(h_0.word == </s>)
        t = (adjoin-right, TOP');
      else{
        if(h_0.tag == NTlabel)
          t = [(adjoin-{left,right}, NTlabel), 
               null];
        else
          t = [(unary, NTlabel), 
               (adjoin-{left,right}, NTlabel), 
               null];
      }
    }
    else{
      if(h_0.tag == NTlabel)
        t = null;
      else
        t = [(unary, NTlabel), null];
    }
  }while(t != null) //done PARSER
}while(!(h_0.word==</s> && h_{-1}.word==<s>))
t = (adjoin-right, TOP); //adjoin <s>_SB; DONE;
\end{verbatim}
\caption{Language Model Operation} \label{algorithm:slm_operation}
\end{algorithm}
The unary transition is allowed only when the most recent exposed
head is a leaf of the tree --- a regular word along with its POStag
--- hence it can be taken at most once at a given position in the
input word string. The second subtree in Figure~\ref{fig:w_parse} provides an example
of a unary transition followed by a null transition.

It is easy to see that any given word sequence with a possible parse
and headword annotation is generated by a unique sequence of model
actions. This will prove very useful in initializing our model
parameters from a treebank.

\section{Probabilistic Model} \label{section:prob_model}

\begin{figure}[ha]
  \begin{center}
    \epsfig{file=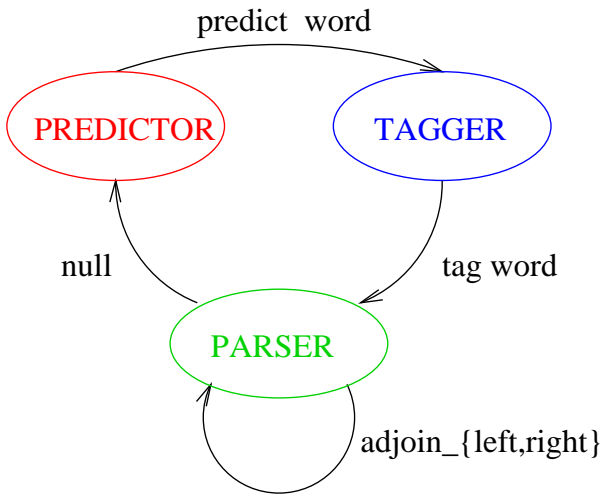, width=8cm}
  \end{center}
  \caption{Language Model Operation as a Finite State Machine} \label{fig:fsm}
\end{figure}

The language model operation provides an encoding of a given word
sequence along with a parse tree $W,T$ into a sequence of elementary
model actions and it can be formalized as a finite state machine (FSM)
--- see Figure~\ref{fig:fsm}. In order to obtain a correct probability
assignment $P(W,T)$ one has to simply assign proper conditional
probabilities on each transition in the FSM that describes the model.

The probability $P(W,T)$ of a word sequence $W$ and a complete parse
$T$ can be broken into:
\begin{eqnarray}
  \label{eq:probability_factorization}
  \lefteqn{P(W,T) =}\nonumber \\
  & & \prod_{k=1}^{n+1}[P(w_k|W_{k-1}T_{k-1}) \cdot
  P(t_k|W_{k-1}T_{k-1},w_k) \cdot P(T_{k-1}^k|W_{k-1}T_{k-1},w_k,t_k)]\vspace{0.3cm}\\
%\end{eqnarray}
%\begin{eqnarray}
  \lefteqn{P(T_{k-1}^k|W_{k-1}T_{k-1},w_k,t_k) =
  \prod_{i=1}^{N_k}P(p_i^k|W_{k-1}T_{k-1},w_k,t_k,p_1^k\ldots p_{i-1}^k)}
\end{eqnarray}
where:
\begin{itemize}
\item $W_{k-1} T_{k-1}$ is the word-parse $(k-1)$-prefix
\item $w_k$ is the word predicted by WORD-PREDICTOR
\item $t_k$ is the tag assigned to $w_k$ by the TAGGER
\item $T_{k-1}^k$ is the parse structure attached to $T_{k-1}$ in
  order to generate\\ $T_k = T_{k-1} \parallel T_{k-1}^k$
\item $N_k - 1$ is the number of operations the PARSER executes at
  position $k$ of the input string before passing control to the
  WORD-PREDICTOR (the $N_k$-th operation at position k is the
  \verb+null+ transition); $N_k$ is a function of $T$
\item $p_i^k$ denotes the i-th PARSER operation carried out at
  position k in the word string:\\
  $p_{i}^k \in\{$ \verb+(adjoin-left, NTtag)+$,$
  \verb+(adjoin-right, NTtag)+$\}, 1 \leq i < N_k$ ,\\
  $p_{i}^k = $\verb+null+$, i = N_k$
\end{itemize}

Each \mbox{$(W_{k-1} T_{k-1},w_k,t_k,p_1^k \ldots p_{i-1}^k)$} is a
valid word-parse k-prefix $W_k T_k$ at position $k$ in the sentence,
$i=\overline{1,N_k}$.

To ensure a proper probabilistic model over the set of \emph{complete
  parses} for any sentence $W$, certain PARSER and WORD-PREDICTOR
probabilities must be given specific values\footnote{Not all the paths
  through the FSM that describes the language model will result in a
  correct binary tree as defined by the complete parse,
  Figure~\ref{fig:c_parse}. In order to prohibit such paths, we impose
  a set of constraints on the probability values of different model
  components, consistent with
  Algorithm~\ref{algorithm:slm_operation}}:
\begin{itemize} \label{prob_model_requirements}
\item $P($\verb+null+$|W_kT_k) = 1$, if \verb+h_{-1}.word = <s>+ and
  \verb+h_{0}+ $\neq$ \verb+(</s>, TOP')+ --- that is, before
  predicting \verb+</s>+ --- ensures that \verb+(<s>, SB)+ is adjoined
  in the last step of the parsing process;
\item
  \begin{itemize}
  \item $P($\verb+(adjoin-right, TOP)+$|W_k T_k) =1$,\\ if 
    \verb+h_0 = (</s>, TOP')+ and \verb+h_{-1}.word = <s>+
%    \\and\\
  \item $P($\verb+(adjoin-right, TOP')+$|W_k T_k) =1$,\\
    if \verb+h_0 = (</s>, TOP')+ and \verb+h_{-1}.word+ $\neq$
    \verb+<s>+
  \end{itemize}
  ensure that the parse generated by our model is consistent with the
  definition of a complete parse;
\item $\exists\epsilon > 0, \forall W_{k-1}T_{k-1},
  P(w_k$=\verb+</s>+$|W_{k-1}T_{k-1})\geq\epsilon$ ensures that the
  model halts with probability one.
\end{itemize}

A few comments on Eq.~(\ref{eq:probability_factorization}) are in
order at this point. Eq.~(\ref{eq:probability_factorization}) assigns
probability to a directed acyclic graph $(W,T)$. Many other possible
probability assignments are possible, and probably the most obvious
choice would have been the factorization used in context free
grammars. Our choice is dictated by its simplicity and left-to-right
bottom-up operation. This also leads to a proper and very simple word
level probability estimate --- see
Section~\ref{section:word_level_ppl} --- even when pruning the set of
parses $T$.

Our factorization Eq.~(\ref{eq:probability_factorization}) assumes
certain dependencies between the nodes in the graph $(W,T)$.  Also, in
order to be able to reliably estimate the model components we need to
make appropriate equivalence classifications of the conditioning part
for each component, respectively. This is equivalent to making certain
conditional independence assumptions which may not be --- and probably
are not --- correct and thus have a damaging effect on the modeling
power of our model.

The equivalence classification should identify the strong predictors
in the context and allow reliable estimates from a treebank. Our
choice is inspired by~\cite{mike96} and intuitively explained in
Section~\ref{section:basic_idea}:
\begin{eqnarray}
  P(w_k|W_{k-1} T_{k-1}) =& P(w_k|[W_{k-1} T_{k-1}]) & = P(w_k|h_0, h_{-1}) \label{eq:word_predictor_prob}\\
  P(t_k|w_k,W_{k-1} T_{k-1}) =& P(t_k|w_k,[W_{k-1} T_{k-1}]) & = P(t_k|w_k, h_0.tag, h_{-1}.tag)\label{eq:tag_predictor_prob}\\
  P(p_i^k|W_{k}T_{k}) =& P(p_i^k|[W_{k}T_{k}]) & = P(p_i^k|h_0, h_{-1})\label{eq:parser_prob}
\end{eqnarray}
The above equivalence classifications are limited by the severe data
sparseness problem faced by the 3-gram model and by no means do we
believe that they are adequate, especially that used in PARSER model
(\ref{eq:parser_prob}). Richer equivalence classifications should use
a probability estimation method that deals better with sparse data
than the one presented in section~\ref{subsection:modeling_tools}.
The limit in complexity on the WORD-PREDICTOR
(Eq.\ref{eq:word_predictor_prob}) also makes our model directly
comparable with a 3-gram model. A few different equivalence
classifications have been tried as described in section~\ref{section:choosing_params}.

It is worth noting that if the binary branching structure developed by
the parser were always right-branching and we mapped the POStag and
non-terminal tag vocabularies to a single type, then our model would
be equivalent to a trigram language model.

\section{Modeling Tool} \label{subsection:modeling_tools}

All model components --- WORD-PREDICTOR, TAGGER, PARSER --- are
conditional probabilistic models of the type $P(u|z_1, z_2, \ldots ,z_n)$ where $u,
z_1, z_2, \ldots ,z_n$ belong to a mixed set of words, POStags,
non-terminal tags and parser operations ($u$ only). 
Let $\mathcal{U}$ be the vocabulary in which the predicted random variable $u$ takes values.

\begin{figure}[ha]
  \begin{center}
    \epsfig{file=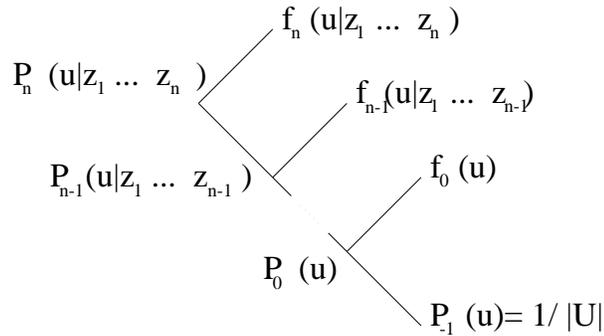, width=8cm}
  \end{center}
  \caption{Recursive Linear Interpolation} \label{fig:mixing_scheme}
\end{figure}

For simplicity, the probability estimation method we chose was recursive linear
interpolation among relative frequency estimates of different orders
$f_k(\cdot), k=0 \ldots n$ using a recursive mixing scheme (see Figure~\ref{fig:mixing_scheme}):
\begin{eqnarray}
\lefteqn{P_n(u|z_1,\ldots,z_n) =} \nonumber\\
&\lambda(z_1,\ldots,z_n) \cdot P_{n-1}(u|z_1,\ldots,z_{n-1}) + (1 - \lambda(z_1, \ldots ,z_n)) \cdot f_n(u|z_1, \ldots ,z_n),\\
\lefteqn{P_{-1}(u) = uniform(\mathcal{U})}
\end{eqnarray}
where:
\begin{itemize}
\item  $z_1,\ldots,z_n$ is the context of order $n$ when predicting $u$;
\item  $f_k(u|z_1,\ldots,z_k)$ is the order-k relative frequency estimate for the
conditional probability $P(u|z_1,\ldots,z_k)$:
\begin{eqnarray*}
  f_k(u|z_1,\ldots,z_k) & = & C(u,z_1,\ldots,z_k)/C(z_1,\ldots,z_k),\ k = 0 \ldots n,\\
  C(u,z_1,\ldots,z_k)   & = & \sum_{z_{k+1} \in \mathcal{Z}_{k+1}} \ldots
  \sum_{z_{n} \in \mathcal{Z}_{n}} C(u,z_1,\ldots,z_k, z_{k+1} \ldots z_n),\\
  C(z_1,\ldots,z_k)     & = & \sum_{u \in \mathcal{U}} C(u,z_1,\ldots,z_k),\\
\end{eqnarray*}
\item  $\lambda(z_1,\ldots,z_k)$ are the interpolation
  coefficients satisfying\\ $\lambda(z_1,\ldots,z_k) \in [0,1], k = 0 \ldots n$.
\end{itemize}

The $\lambda(z_1,\ldots,z_k)$ coefficients are grouped into
equivalence classes --- ``tied'' --- based on the range into which the count
$C(z_1, \ldots , z_k)$ falls; the count ranges for each equivalence
class --- also called ``buckets'' --- are set such that a
statistically sufficient number of events $(u|z_1,\ldots,z_k)$ fall in
that range. The approach is a standard one~\cite{jelinek80}.
In order to determine the interpolation weights, we apply the deleted
interpolation technique: 
\begin{itemize}
\item we split the training data in two sets --- ``development'' and\\
  ``cross-validation'', respectively;
\item we get the relative frequency --- maximum likelihood --- estimates\\
  $f_k(u|z_1,\ldots,z_k),\ k = 0 \ldots n$ from ``development'' data
\item we employ the expectation-maximization (EM) algorithm~\cite{em77} for
  determining the maximum likelihood estimate from ``cross-validation'' data of the
  ``tied'' interpolation weights $\lambda(C(z_1, \ldots ,
  z_k))$\footnote{The ``cross-validation'' data cannot be the same as the development 
    data; if this were the case, the maximum likelihood estimate for
    the interpolation weights would be 
    $\lambda(C(z_1, \ldots , z_k)) = 0$, 
    disallowing the mixing of different order relative frequency
    estimates and thus performing no smoothing at all};
\end{itemize}

We have written a general deleted interpolation tool which takes as
input:
\begin{itemize}
\item joint counts $z_1, z_2, \ldots , z_n, u$ gathered from the
``development'' and ''cross-validation data'', respectively
\item initial interpolation values and bucket descriptors for all levels in the
deleted interpolation scheme
\end{itemize}
The program runs a pre-specified number
of EM iterations at each level in the deleted interpolation scheme ---
from bottom up, $k=0 \ldots n$ --- and returns a descriptor file
containing the estimated coefficients.

The descriptor file can then be used for initializing the module and
thus rendering it usable for the calculation of conditional
probabilities $P(u/z_1, z_2, \ldots , z_n)$.
A sample descriptor file for the deleted interpolation statistics
module is shown in Table~\ref{table:del_int_descriptor}.
\begin{table}
  \begin{verbatim}
## Stats_Del_Int descriptor file
## $Id: del_int_descriptor.tex,v 1.3 1999/03/16 17:54:16 chelba Exp $
Stats_Del_Int::_main_counts_file = counts.devel.HH_w.E0.gz ;
Stats_Del_Int::_held_counts_file = counts.check.HH_w.E0.gz ;
Stats_Del_Int::_max_order = 4 ;
Stats_Del_Int::_no_iterations = 0 ;
Stats_Del_Int::_no_iterations_at_read_in = 100 ;
Stats_Del_Int::_predicted_vocabulary_chunk = 0 ;
Stats_Del_Int::_prob_Epsilon = 1e-07 ;

Stats_Del_Int::lambdas_level.0 =  2:__1__0.019 ;
Stats_Del_Int::buckets_level.0 =  2:__0__10000000 ;

Stats_Del_Int::lambdas_level.1 = 13:__1__0.5__0.5__0.5__0.5__0.5__1__1
                                    __0.449__1__0.260__0.138__0.073 ;
Stats_Del_Int::buckets_level.1 = 13:__0__1__2__4__8__16__32__64
                                    __128__256__512__1024__10000000 ;

Stats_Del_Int::lambdas_level.2 = 13:__1__0.853__0.787__0.745__0.692
                                    __0.637__0.579__0.489__0.427__0.358
                                    __0.296__0.258__0.213 ;
Stats_Del_Int::buckets_level.2 = 13:__0__1__2__4__8
                                    __16__32__64__128__256
                                    __512__1024__10000000 ;

Stats_Del_Int::lambdas_level.3 = 13:__1__0.935__0.905__0.878__0.855
                                    __0.812__0.743__0.686__0.633__0.595
                                    __0.548__0.515__0.517 ;
Stats_Del_Int::buckets_level.3 = 13:__0__1__2__4__8
                                    __16__32__64__128__256
                                    __512__1024__10000000 ;

Stats_Del_Int::lambdas_level.4 = 13:__1__0.887__0.859__0.838__0.801
                                    __0.761__0.710__0.627__0.586__0.532
                                    __0.523__0.485__0.532 ;
Stats_Del_Int::buckets_level.4 = 13:__0__1__2__4__8
                                    __16__32__64__128__256
                                    __512__1024__10000000 ;
\end{verbatim}

%%% Local Variables: 
%%% mode: latex
%%% TeX-master: "extended"
%%% End: 

  \caption{Sample descriptor file for the deleted interpolation
    module} 
  \label{table:del_int_descriptor}
\end{table}

The deleted interpolation method is not optimal for our
problem. Our models would require a method able to optimally combine
the predictors of different nature in the conditioning part of the
model and this is far from being met by the fixed hierarchical scheme
used for context mixing in deleted interpolation estimation. The best
method would be maximum entropy~\cite{berger:max_ent} but due to its
computational burden we have not used it.

\section{Pruning Strategy}\label{section:pruning}

Since the number of parses for a given word prefix $W_k$ grows faster
than exponential\footnote{Thanks to Bob Carpenter, Lucent Technologies
  Bell Labs, for pointing out this inaccuracy in our~\cite{chelba98}
  paper} with $k$, ${\Omega}(2^k)$, the state space of our model is
huge even for relatively short sentences. We thus have to prune most
parses without discarding the most likely ones for a given prefix
$W_k$.  Our pruning strategy is a synchronous multi-stack search
algorithm.

Each stack contains hypotheses --- partial parses --- that have
been constructed by \emph{the same number of predictor and the same number of parser
operations}. The hypotheses in each stack are ranked according to the
$\ln(P(W_k,T_k))$ score, highest on top. 
The amount of search is controlled by two parameters:
\begin{itemize}
\item the maximum stack depth --- the maximum number of hypotheses
  the stack can contain at any given time; 
\item log-probability threshold --- the difference between the log-probability score of the top-most
  hypothesis and the bottom-most hypothesis at any given state of the
  stack cannot be larger than a given threshold.
\end{itemize}

\begin{figure}
  \begin{center} 
    \epsfig{file=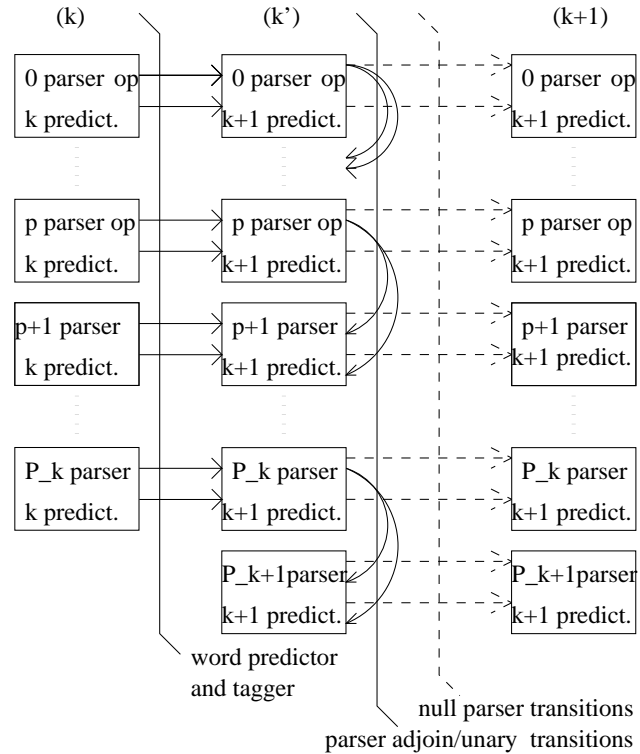,height=10cm}
  \end{center}
  \caption{One search extension cycle} \label{fig:search}
\end{figure}

Figure~\ref{fig:search} shows schematically the operations associated
with the scanning of a new word $w_{k+1}$\footnote{$P_k$ is the
  maximum number of adjoin operations for a k-length word prefix;
  since the tree is binary we have $P_k = k-1$}.

First, all hypotheses in a given stack-vector are expanded with the
following word. Then, for each possible POS tag the following word can
take, we expand the hypotheses further. Due to the finite stack size,
some are discarded. We then proceed with the PARSER expansion
cycle, which takes place in two steps:
\begin{enumerate}
\item first all hypotheses in a given stack are expanded with all
  possible PARSER actions excepting the \verb+null+ transition.
  The resulting hypotheses are sent to the immediately lower stack of
  the same stack-vector --- same number of WORD-PREDICTOR operations
  and exactly one more PARSER move. Some are discarded due to
  finite stack size.
\item after completing the previous step, all resulting hypotheses are
  expanded with the \verb+null+ transition and sent into the next
  stack-vector. Pruning can still occur due to the log-probability
  threshold on each stack.
\end{enumerate}

The pseudo-code for parsing a given input sentence is given in
Algorithms~\ref{algorithm:search_alg}-~\ref{algorithm:parser_alg}.
\begin{algorithm}[h]
\begin{verbatim}
current_stack_vector // set of stacks at current input position
future_stack_vector  // set of stacks at future  input position
hypothesis           // initial hypothesis
stack                // initial empty stack

// initialize algorithm
insert hypothesis in stack;
push stack at end of current_stack_vector;

// traverse input sentence
for each position in input sentence{
  PREDICTOR and TAGGER extension cycle;
  current_stack_vector = future_stack_vector;
  erase future_stack_vector;

  PARSER extension cycle;
  current_stack_vector = future_stack_vector;
  erase future_stack_vector;
}
// output the hypothesis with the highest score;
output max scoring hypothesis in current_stack_vector;
\end{verbatim}
\caption{Pruning Algorithm} \label{algorithm:search_alg}
\end{algorithm}

\begin{algorithm}[h]
\begin{verbatim}
current_stack_vector // set of stacks at current input position
future_stack_vector  // set of stacks at future  input position
word                 // word at current input position
for each stack in current_stack_vector{
  // based on number of predictor and parser operations
  identify corresponding future_stack in future_stack_vector;

  for each hypothesis in stack{
    for all possible POStag assignments for word{ //CACHE-ING
      expand hypothesis with word, POStag;
      insert hypothesis in future_stack;
    }
  }
}
\end{verbatim}
\caption{PREDICTOR and TAGGER Extension Algorithm} \label{algorithm:predictor_alg}
\end{algorithm}

\begin{algorithm}[h]
\begin{verbatim}
current_stack_vector // set of stacks at current input position
future_stack_vector  // set of stacks at future  input position

// all possible parser transitions but the null-transition
for each stack in current_stack_vector, from bottom up{
  // based on number of parser operations
  identify corresponding future_stack in current_stack_vector;
  for each hypothesis in current_stack{         // HARD PRUNING
    for each parser_transition except the null-transition{//CACHE-ING
      expand hypothesis with parser_transition;
      insert hypothesis in future_stack;
    }
  }
}
// null-transition moves us to the next position in the input
for each stack in current_stack_vector{
  // based on number of predictor and parser operations
  identify corresponding future_stack in future_stack_vector;
  for each hypothesis in current_stack{
    expand hypothesis with null-transition;
    insert hypothesis in future_stack;
  }
}
prune future_stack_vector                 //SECOND PRUNING STEP
\end{verbatim}
\caption{Parser Extension Algorithm} \label{algorithm:parser_alg}
\end{algorithm}

\subsubsection{Second Pruning Step}
The pruning strategy described so far proved to be
insufficient\footnote{Assuming that all stacks contain the maximum
  number of entries --- equal to the stack-depth --- the search effort
  grows squared with the sentence length} so in order to approximately
linearize the search effort with respect to sentence length, we chose
to discard also the hypotheses whose score is more than a fixed
log-probability relative threshold below the score of the topmost
hypothesis in the current stack vector. This additional pruning step
is performed after all hypotheses in stage $k'$ have been extended
with the \verb+null+ parser transition.

\subsubsection{Cashed TAGGER and PARSER Lists}
Another opportunity for speeding up the search is to have a cached
list of possible POStags/parser-operations in a given TAGGER/PARSER
context. A good cache-ing scheme should use an equivalence
classification of the context that is specific enough to actually
reduce the list of possible options and general enough to apply in
almost all the situations. For the TAGGER model we cache the list of
POStags for a given word seen in the training data and scan only those
in the TAGGER extension cycle --- see
Algorithm~\ref{algorithm:predictor_alg}. For the PARSER model we cache
the list of parser operations seen in a given $(h_0.tag, h_{-1}.tag)$
context in the training data; parses that expose heads whose pair of
NTtags has not been seen in the training data are discarded--- see
Algorithm~\ref{algorithm:parser_alg}.

\section{Word Level Perplexity}\label{section:word_level_ppl}

Attempting to calculate the conditional perplexity by assigning to a whole sentence
the probability:
\begin{eqnarray}
  P(W|T^*) & = & \prod_{k=0}^n P(w_{k+1}|W_{k}T_{k}^*), \label{eq:ppl0}
\end{eqnarray}
where $T^* = argmax_{T}P(W,T)$ --- the search for $T^*$ being carried
out according to our pruning strategy --- is not valid
because it is not causal: when predicting $w_{k+1}$ we would be using $T^*$ which was
determined by looking at the entire sentence. To be able to compare 
the perplexity of our model with that resulting from the standard
trigram approach, we would need to factor in the entropy of
guessing the prefix of the final best parse $T_{k}^*$ \emph{before predicting} $w_{k+1}$,
based solely on the word prefix $W_{k}$.

To maintain a left-to-right operation of the language model, the
probability assignment for the word at position $k+1$ in the input
sentence was made using:
\begin{eqnarray}
  P(w_{k+1}|W_{k}) & = & \sum_{T_{k}\in S_{k}}P(w_{k+1}|W_{k}T_{k})\cdot\rho(W_{k},T_{k}),\label{eq:ppl1}\\
  \rho(W_{k},T_{k}) & = & P(W_{k}T_{k})/\sum_{T_{k} \in
    S_{k}}P(W_{k}T_{k}) \nonumber
\end{eqnarray}
where $S_{k}$ is the set of all parses present in our stacks at the
current stage $k$. This leads to the following formula for evaluating
the perplexity:
\begin{eqnarray}
  PPL(SLM) = exp(-1/N \sum_{i=1}^{N}\ln{[P(w_i||W_{i-1})]})
\end{eqnarray}

Note that if we set $\rho(W_{k},T_{k}) = \delta(T_k, T_k^*|W_k)$ ---
0-entropy guess for the prefix of the parse $T_k$ to equal that of the
final best parse $T_k^*$--- the two probability assignments
(\ref{eq:ppl0}) and  (\ref{eq:ppl1}) would be the same, yielding a lower
bound on the perplexity achievable by our model when using a given
pruning strategy. 

Another possibility for evaluating the word level perplexity of our
model is to approximate the probability of a whole sentence:
\begin{eqnarray}
  P(W) = \sum_{k=1}^N P(W,T^{(k)}) \label{eq:ppl2}
\end{eqnarray}
where $T^{(k)}$ is one of the ``N-best'' --- in the sense defined by
our search --- parses for $W$. This is a deficient probability
assignment, however useful for justifying the model parameter
re-estimation to be presented in Chapter~\ref{chapter:model_reest}.

The two estimates (\ref{eq:ppl1}) and (\ref{eq:ppl2}) are both
consistent in the sense that if the sums are carried over all possible
parses we get the correct value for the word level perplexity of our model. 

Another important observation is that the next-word predictor
probability\\ $P(w_{k+1}|W_{k}T_{k})$ in~(\ref{eq:ppl1}) 
{\it need not be the same} as the WORD-PREDICTOR
probability~(\ref{eq:word_predictor_prob}) used to extract the structure
$T_k$, thus leaving open the possibility of estimating it
separately. To be more specific, we can in principle have a
WORD-PREDICTOR model component that operates within the parser model
whose role strictly to extract syntactic structure and a second model
that is used only for the left to right probability assignment:
\begin{eqnarray}
  P_2(w_{k+1}|W_{k}) & = & \sum_{T_{k}\in
    S_{k}}P_{WP}(w_{k+1}|W_{k}T_{k})\cdot\rho(W_{k},T_{k}),\label{eq:ppl1_l2r}\\
  \rho(W_{k},T_{k}) & = & P(W_{k}T_{k})/\sum_{T_{k} \in
    S_{k}}P(W_{k}T_{k}) \label{eq:ppl1_rho}
\end{eqnarray}
In this case the interpolation coefficient given by~\ref{eq:ppl1_rho}
uses the regular WORD-PREDICTOR model whereas the prediction of the
next word for the purpose of word level probability assignment is made 
using a separate model $P_{WP}(w_{k+1}|W_{k}T_{k})$.

\chapter{Structured Language Model Parameter Estimation}\label{chapter:model_reest}

As outlined in section~\ref{section:word_level_ppl}, the word level
probability assigned to a training/test set by our model is calculated
using the proper word-level probability assignment
in equation~(\ref{eq:ppl1}). An alternative which leads to a deficient
probability model is to sum over all the complete parses that survived
the pruning strategy, formalized in equation~(\ref{eq:ppl2}). Let the likelihood assigned to a
corpus $\mathcal{C}$ by our model $P_{\theta}$ be denoted by:
\begin{itemize}
\item $\mathcal{L}^{L2R}(\mathcal{C},P_{\theta})$, where $P_{\theta}$
  is calculated using~(\ref{eq:ppl1}), repeated here for clarity:
  \begin{eqnarray}
    P(w_{k+1}|W_{k}) & = & \sum_{T_{k}\in 
      S_{k}}P(w_{k+1}|W_{k}T_{k})\cdot\rho(W_{k},T_{k}),\nonumber \\
    \rho(W_{k},T_{k}) & = & P(W_{k}T_{k})/\sum_{T_{k} \in
      S_{k}}P(W_{k}T_{k}) \nonumber
  \end{eqnarray}
  Note that this is a proper probability model.
\item $\mathcal{L}^{N}(\mathcal{C},P_{\theta})$, where $P_{\theta}$ is 
  calculated using~(\ref{eq:ppl2}):
  \begin{eqnarray}
    P(W) = \sum_{k=1}^N P(W,T^{(k)}) \nonumber
  \end{eqnarray}
  This is a deficient probability model: due to the fact that we are
  not summing over all possible parses for a given word sequence $W$
  --- we discard most of them through our pruning strategy ---
  we underestimate the probability $P(W)$ and thus $\sum_{W} P(W) < 1$.
\end{itemize}

One seeks to devise an algorithm that finds the model
parameter values which maximize the likelihood of a test corpus. This
is an unsolved problem; the standard approach is to resort to maximum
likelihood estimation techniques on a training corpus and make
provisions that will ensure that the increase in likelihood on
training data carries over to unseen test data.

In our case we would like to estimate the model component
probabilities~(\ref{eq:word_predictor_prob}~--~\ref{eq:parser_prob}). 
The smoothing scheme outlined in Section~\ref{subsection:modeling_tools}
is intended to prevent overtraining and tries to ensure that maximum likelihood
estimates on the training corpus will carry over to test data. Since
our problem is one of maximum likelihood estimation from incomplete data --- the
parse structure along with POS/NT tags and headword annotation for a
given observed sentence is hidden --- our approach will make heavy use 
of the EM algorithm variant presented in chapter~\ref{chapter:EM_ML}.

The estimation procedure proceeds in two stages: first the ``N-best
training'' algorithm~(see
Section~\ref{section:model_reest_first_pass}) is employed to increase
the training data ``likelihood''
$\mathcal{L}^{N}(\mathcal{C},P_{\theta})$; 
we rely on the consistency
property outlined at the end of Section~\ref{section:word_level_ppl}
to correlate the increase in $\mathcal{L}^{N}(\mathcal{C},P_{\theta})$
with the desired increase of
$\mathcal{L}^{L2R}(\mathcal{C},P_{\theta})$. The initial parameters
for this first estimation stage are gathered from a treebank as
described in Section~\ref{section:initial_parameters}. 

The second stage estimates the model parameters such that
$\mathcal{L}^{L2R}(\mathcal{C},P_{\theta})$ is increased. The basic
idea is to realize that the WORD-PREDICTOR in the structured
language model~(as described in chapter \ref{chapter:slm}) and that
used for word prediction in the $\mathcal{L}^{L2R}(\mathcal{C},P_{\theta})$
calculation can be estimated as two separate components: one that is
used for structure generation and a second one which is
used strictly for predicting the next word as described
in equation~(\ref{eq:ppl1}). The initial parameters for the second
component are obtained by copying the WORD-PREDICTOR estimated at
stage one.

As a final step in refining the model we have linearly interpolated the structured
language model~(\ref{eq:ppl1}) with a trigram model. Results and
comments on them are presented in the last section of the chapter.

\section{Maximum Likelihood Estimation from Incomplete Data} \label{chapter:EM_ML}
% state the EM algorithm; 
% show that it is equivalent to alternating minimization between sets using minimum-divergence;
% prove the convergence of our reestimation procedure;
In many practical situations we are confronted with the following
situation: we are given a collection of data points $\mathcal{T} =
\{y_1, \ldots, y_n\}, y_i \in \mathcal{Y}$ --- training data --- which
we model as independent samples drawn from the $Y$ marginal of the
parametric distribution:
$$q_{\theta}(x,y), \theta \in \Theta, x \in \mathcal{X}, y \in
\mathcal{Y}$$
where $X$ is referred to as the hidden variable and
$\mathcal{X}$ as the hidden event space, respectively.  The set
$$Q(\Theta) \doteq \{q_{\theta}(X,Y):\theta \in \Theta \}$$
is
referred to as the \emph{model set}. Let $f_{\mathcal{T}}(Y)$ be the
relative frequency probability distribution induced on $\mathcal{Y}$
by the collection $\mathcal{T}$.
  
We wish to find the maximum-likelihood estimate of $\theta$:
\begin{eqnarray}
  \mathcal{L}(\mathcal{T}; q_{\theta}) & \doteq &
  \sum_{y \in \mathcal{Y}} f_{\mathcal{T}}(y) \log (\sum_{ x \in
    \mathcal{X}} q_{\theta}(x,y)) \label{eq:T_q_likelihood}\\
  \theta^{*} & = & \arg \max_{\theta \in \Theta}  \mathcal{L}(\mathcal{T}; q_{\theta}) \label{eq:ML_theta}
\end{eqnarray}

%Although not guaranteed to find the global maximum $\theta^{*} \in
%\Theta$, the EM(Expectation-Maximization) iterative algorithm~\cite{em77} will
%find a stationary point\footnote{A stationary point is a point where
%  the gradient of the likelihood is null: 
%  $\nabla_{\theta} \mathcal{L}(\mathcal{T}; q_{\theta}) = 0$ is null.} of the
%training data likelihood $\mathcal{L}(\mathcal{T}; q_{\theta})$~(see
%\ref{eq:T_q_likelihood}).

Starting with an initial parameter value $\theta_i$, it is shown that
a sufficient condition for increasing the likelihood of the training
data $\mathcal{T}$ (see Eq.~\ref{eq:T_q_likelihood}) is to find a new
parameter value $\theta_{i+1}$ that maximizes the so called EM
auxiliary function defined as:
\begin{eqnarray}
  EM_{\mathcal{T},\theta_i}(\theta) & \doteq & \sum_{y \in \mathcal{Y}}
    f_{\mathcal{T}}(y) E_{q_{\theta_i}(X|Y)}[\log(q_{\theta}(X,Y)|y)], 
    \theta \in \Theta \label{eq:EM_aux}
\end{eqnarray}

The EM theorem proves that choosing:
\begin{eqnarray}
  \theta_{i+1} & = & \arg \max_{\theta \in \Theta}
  EM_{\mathcal{T},\theta_i}(\theta)   \label{eq:EM_update}
\end{eqnarray}
ensures that the likelihood of the training data under the new
parameter value is not lower than that under the old one, formally:
\begin{eqnarray}
  \mathcal{L}(\mathcal{T}; q_{\theta_{i+1}}) & \geq & 
  \mathcal{L}(\mathcal{T}; q_{\theta_i})   \label{eq:EM_increase}
\end{eqnarray}

%A fixed point of the EM algorithm --- $\theta_i = \theta_{i+1}$ ---
%is also a stationary point of the likelihood function $\mathcal{L}(\mathcal{T};
%q_{\theta}), \theta \in \Theta$. 
Under more restrictive conditions on the model family $Q(\Theta)$ it
can be shown that the fixed points of the EM procedure --- $\theta_i =
\theta_{i+1}$ --- are in fact local maxima of the likelihood
function $\mathcal{L}(\mathcal{T}; q_{\theta}), \theta \in \Theta$.
The study of convergence properties under different assumptions on the
model class as well as different flavors of the EM algorithm is an
open area of research.

The fact that the algorithm is naturally formulated to operate with
probability distributions --- although this constraint can be relaxed
--- makes it attractive from a computational point of view: an
alternative to maximizing the training data likelihood would be to
apply gradient maximization techniques; this may be particularly
difficult if not impossible when the analytic description of the
likelihood as a function of the parameter $\theta$ is complicated.

To further the understanding of the computational aspects of using the
EM algorithm we notice that the EM update~(\ref{eq:EM_update})
involves two steps:
\begin{itemize}
\item E-step: for each sample $y$ in the training data $\mathcal{T}$,
  accumulate the expectation of $\log(q_{\theta}(X,Y)|y)$ under the
  distribution $q_{\theta_i}(x|y)$; no matter what the actual analytic
  form of $\log(q_{\theta}(X,Y))$ is, this requires to traverse all
  possible derivations $(x,y)$ of the seen event $y$ that have
  non-zero conditional probability $q_{\theta_i}(X=x|Y=y)>0$;
\item M-step: find maximizer of the auxiliary
  function~(\ref{eq:EM_aux}).
\end{itemize}

Typically the M-step is simple and the computational bottleneck is the
E-step.  The latter becomes intractable with large training data set
size and rich hidden event space, as usually required by practical
problems.

In order to overcome this limitation, the model space $Q(\Theta)$ is
usually structured such that dynamic programming techniques can be
used for carrying out the E-step --- see for example the hidden Markov
model(HMM) parameter reestimation procedure~\cite{hmm72}. However this
advantage does not come for free: in order to be able to structure the
model space we need to make independence assumptions that weaken the
modeling power of our parameterization. Fortunately we are not in a
hopeless situation: a simple modification of the EM algorithm allows
the traversal of only a subset of all possible $(x,y), x \in
\mathcal{X}|y$ for each training sample $y$ --- the procedure is
dubbed ``N-best training`` --- thus rendering it applicable to a much
broader and more powerful class of models.

\subsection{N-best Training Procedure} \label{section:nbest_training}

Before proceeding with the presentation of the N-best training
procedure, we would like to introduce a view of the 
EM algorithm based on information geometry. Having gained this
insight we can then easily justify the N-best
training procedure. This is an interesting area of research to which we 
were introduced by the presentation in~\cite{byrne98}.

\subsubsection{Information Geometry and EM}

The problem of maximum likelihood estimation from incomplete data can
be viewed in an interesting geometric framework. Before proceeding,
let us introduce some concepts and the associated notation. 

\paragraph{Alternating Minimization} \label{sec:AM}

Consider the problem of finding the minimum Euclidean distance between
two convex sets A and B:
\begin{eqnarray}
  \label{eq:AM}
  d^* \doteq d(a^*, b^*) & = & \min_{a \in A, b \in B} d(a,b)
\end{eqnarray}
\begin{figure}[htbp]
  \begin{center}
    \epsfig{file=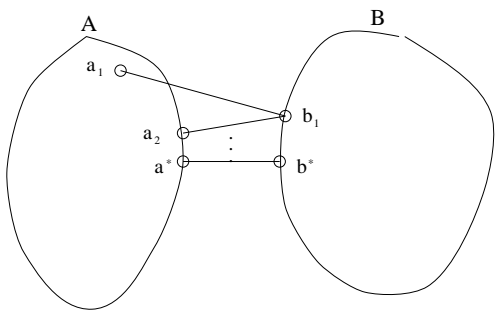, height=4cm}
    \caption{Alternating minimization between convex sets}
    \label{fig:AM}
  \end{center}
\end{figure}
The following iterative procedure(see
figure~\ref{fig:AM}) should lead to the solution: start with a random
point $a_1 \in A$; find the point $b_1 \in B$ closest to $a_1$; then
fix $b_1$ and find the point $a_2 \in A$ closest to $b_1$ and so
on. It is intuitively clear that the distance between the two points
considered at each iteration cannot increase and that the fixed point
of the above procedure --- the choice for the $(a,b)$ points does not
change from one iteration to the next --- is the minimum distance
$d^*$ between the sets $A$ and $B$.

Formalizing this intuition proves to be less simple for a more general 
setup --- the specification of sets $A$ and $B$ and the distance used.
Csiszar and Tusnady have derived
sufficient conditions under which the above alternating minimization
procedure converges to the minimum distance between the two sets~\cite{info_geom_and_AM}. As
outlined in~\cite{info_theory:AM}, this algorithm is applicable to
problems in information theory --- channel capacity and rate
distortion calculation --- as well as in statistics --- the EM
algorithm.

\paragraph{EM as alternating minimization}

Let $Q(\Theta)$ be the family of probability
distributions from which we want to choose the one maximizing the
likelihood of the training data~(\ref{eq:T_q_likelihood}). Let us also
define a family of desired distributions on $\mathcal{X} \times
\mathcal{Y}$ whose $Y$ marginal induced by the training data is the
same as the relative frequency estimate $f_{\mathcal{T}}(Y)$: 
$$P_{\mathcal{T}} = \{p(X,Y): p(Y) = f_{\mathcal{T}}(Y)\}$$ 

For any pair $(p, q) \in P_{\mathcal{T}} \times Q(\Theta)$, the
Kullback-Leibler distance (KL-distance) between $p$ and
$q$ is defined as:
\begin{eqnarray}
  D(p \parallel q) & \doteq & \sum_{x \in \mathcal{X}, y \in \mathcal{Y}} p(x,y) 
  \log \frac{p(x,y)}{q(x,y)}   \label{eq:KL_dist}
\end{eqnarray}

As shown in~\cite{info_geom_and_AM}, under certain conditions on the
families $P_{\mathcal{T}}$ and $Q(\Theta)$ and using the KL-distance, the alternating
minimization procedure described in the previous section converges to
the minimum distance between the two sets:
\begin{eqnarray}
  D(p^* \parallel q^*) & = & \min_{p \in P_{\mathcal{T}}, q \in Q(\Theta)} D(p \parallel q) \label{eq:KL_min_dist}
\end{eqnarray}

It can be easily shown (see appendix~\ref{app:min_D_max_L}) that the
model distribution $q*$ that satisfies (\ref{eq:KL_min_dist}) is also the one maximizing
the likelihood of the training data, 
$$q^* = \arg\max_{g_{\theta} \in Q(\Theta)} \mathcal{L}(\mathcal{T}; q_{\theta})$$
Moreover, the alternating minimization procedure leads exactly to the
EM update equation(\ref{eq:EM_aux}, \ref{eq:EM_update}), as shown
in~\cite{info_geom_and_AM} and sketched in
appendix~\ref{app:EM_eq_minD}.

The $P_{\mathcal{T}}$ and $Q(\Theta)$ families one encounters in
practical situations may not satisfy the conditions specified
in~\cite{info_geom_and_AM}. However, one can easily note that decrease in $D(p \parallel
q)$ at each step and correct I-projection from $q \in Q(\Theta)$ to
$P_{\mathcal{T}}$ --- finding $p \in P_{\mathcal{T}}$ such that we
minimize $D(p \parallel q)$ --- are sufficient conditions for ensuring
that the likelihood of the  training data does not decrease with each
iteration. Since in practice we are bound by computational limitations
and we typically run just a few iterations, the  guaranteed
non-decrease in training data likelihood is sufficient.

\subsection{N-best Training} \label{section:detailed_nbest_training}

In the ``N-best'' training paradigm we use only a subset of the
conditional hidden event space $\mathcal{X}|y$, for any given seen
$y$. Associated with the model space $Q(\Theta)$ we now have a family
of strategies to sample from $\mathcal{X}|y$ a set of ``N-best'' hidden events $x$, for 
any $y \in \mathcal{Y}$. The family is parameterized by $\theta \in \Theta$:
\begin{eqnarray}
  & & \mathcal{S}_{\theta} \doteq \{ s_{\theta}:\mathcal{Y}
  \rightarrow 2^{\mathcal{X}}, \forall \theta \in \Theta \}\label{eq:N-best-sampling}
\end{eqnarray}

With the following definitions:
\begin{eqnarray}
  q_{\theta}^s(X,Y) & \doteq & q_{\theta}(X,Y) \cdot 1_{s_{\theta}(Y)}(X)\label{eq:N-best-model-space1}\\
  q_{\theta}^s(X|Y) & \doteq & \frac{q_{\theta}^s(X,Y)}{\sum_{X \in
      \mathcal{s}_{\theta}(Y)} q_{\theta}(X,Y)} \cdot 1_{s_{\theta}(Y)}(X)\label{eq:N-best-model-space2}\\
  Q(\mathcal{S},\Theta) & \doteq & \{q_{\theta}^s(X,Y):\theta \in \Theta \}
\end{eqnarray}
the alternating minimization procedure between $P_{\mathcal{T}}$ and
$Q(\mathcal{S},\Theta)$ using the KL-distance will find a sequence of parameter
values $\theta_1, \ldots, \theta_n$ for which the ``likelihood'':
\begin{eqnarray}
  \mathcal{L}^s(\mathcal{T}; q_{\theta}^s) & = & \sum_{y \in
    \mathcal{Y}} f_{\mathcal{T}}(y) \log (\sum_{ x \in \mathcal{X}} q_{\theta}^s(x,y)) \label{eq:T_q_likelihood_Nbest}
\end{eqnarray}
is monotonically increasing:
$\mathcal{L}^s(\mathcal{T}; q_{\theta_1}^s) \leq
\mathcal{L}^s(\mathcal{T}; q_{\theta_2}^s) \leq \ldots \leq
\mathcal{L}^s(\mathcal{T}; q_{\theta_n}^s)$. 
Note that due to the truncation of $q_{\theta}(X,Y)$ we are dealing
with a deficient probability model. 

The parameter update at each iteration is very similar to that
specified by the EM algorithm under some sufficient conditions, as
specified in Proposition~\ref{prop:AM_Nbest_convergence} and proved in
Appendix~\ref{app:Nbest_convergence}:

\begin{prop}
  \label{prop:AM_Nbest_convergence}
  Assuming that $\forall \theta \in \Theta, 
  \supp{q_{\theta}(x,y)} = \mathcal{X} \times \mathcal{Y}$ (``smooth'' $q_{\theta}(x,y)$) holds, 
  one alternating minimization step between $P_{\mathcal{T}}$ and
  $Q(\mathcal{S},\Theta)$ ---$\theta_{i} \rightarrow \theta_{i+1}$ --- is equivalent to:
  \begin{eqnarray}
    \theta_{i+1} & = & \arg \max_{\theta \in \Theta} \sum_{y \in \mathcal{Y}}
    f_{\mathcal{T}}(y)
    E_{q_{\theta_i}^s(X|Y)}[\log(q_{\theta}(X,Y)|y)] \label{eq:EM_update_Nbest}
  \end{eqnarray}
  if $\theta_{i+1}$ satisfies:
  \begin{eqnarray}
    s_{\theta_{i}}(y) \subseteq  s_{\theta_{i+1}}(y), \forall y \in \mathcal{T}
  \end{eqnarray}
  Only $\theta \in \Theta\ s.t.\ s_{\theta_{i}}(y) \subseteq
  s_{\theta}(y), \forall y \in \mathcal{T}$ are candidates in the M-step.
\end{prop}

The fact that we are working with a deficient probability model for
which the support of the distributions $q_{\theta_i}^s(X|Y=y), \forall
y \in \mathcal{T}$ 
cannot decrease from one iteration to the next makes the above
statement less interesting: even if we didn't substantially change the model
parameters from one iteration to the next--- $\theta_{i+1} \approx
\theta_{i}$ --- but we chose the sampling function such that
$s_{\theta_{i}}(y) \subset  s_{\theta_{i+1}}(y), \forall y \in
\mathcal{T}$ the ``likelihood'' $\mathcal{L}^s(\mathcal{T};
q_{\theta}^s)$ would still be increasing due to the support expansion, 
although the quality of the model has not actually increased.

In practice the family of sampling functions~$\mathcal{S}_{\theta}$~(\ref{eq:N-best-sampling})
is chosen such that support of  $q_{\theta_i}^s(X|Y=y), \forall
y \in \mathcal{T}$ has constant size --- cardinality, for discrete
hidden spaces. Typically one retains the ``N-best'' after ranking the
hidden sequences $x \in \mathcal{X}|y$ in decreasing order according
to $q_{\theta_i}(X|Y=y), \forall y \in \mathcal{T}$. 
Proposition~\ref{prop:AM_Nbest_convergence} implies that the set of
``N-best'' should not change from one iteration to the next, being an
invariant during model parameter reestimation. In practice however we
recalculate the ``N-best'' after each iteration, allowing the
possibility that new hidden sequences $x$ are being included in the
``N-best'' list at each iteration and others discarded. We do not
have a formal proof that this procedure will ensure monotonic increase
of the ``likelihood'' $\mathcal{L}^s(\mathcal{T}; q_{\theta}^s)$.

% - what a derivation is; equivalence between sentence w/ parse
%   structure and its derivation; 
% - how each of the three model components is parameterized; make
%   clear what is fixed and what is being estimated in the first stage; 
% - explain estimation technique; calculations in appendix;
% - initialization 
\section{First Stage of Model Estimation}\label{section:model_reest_first_pass}

Let $(W,T)$ denote the joint sequence of $W$ with parse 
structure $T$ --- headword and POS/NT tag annotation included. As
described in section~\ref{section:basic_idea}, $W,T$ was produced by a unique
sequence of model actions: word-predictor, tagger, and parser moves. The
ordered collection of these moves will be called a {\it derivation}:
\begin{eqnarray*}
  d(W,T) \doteq  (e_1, \ldots, e_l)
\end{eqnarray*}
where each {\it elementary event} $$e_i \doteq
(u^{(m)}|\underline{z}^{(m)})$$ identifies a model component action:
\begin{itemize}
\item $m$ denotes the model component that took the action,\\ 
  $m \in \{$WORD-PREDICTOR, TAGGER, PARSER $\}$;
\item $u$ is the action taken:
  \begin{itemize}
  \item $u$ is a word for $m=$ WORD-PREDICTOR;
  \item $u$ is a POS tag for $m=$ TAGGER;
  \item $u \in\{$\verb+(adjoin-left, NTtag)+, \verb+(adjoin-right, NTtag)+, \verb+null+$\}$\\ for $m=$ PARSER;
  \end{itemize}
\item $\underline{h}$ is the context in which the action is taken
  (see~equations~(\ref{eq:word_predictor_prob} -- \ref{eq:parser_prob})):
  \begin{itemize}
  \item $\underline{z}=h_0.tag, h_0.word, h_{-1}.tag, h_{-1}.word$ for $m=$ WORD-PREDICTOR;
  \item $\underline{z}=w, h_0.tag, h_{-1}.tag$ for $m=$ TAGGER;
  \item $\underline{z}=h_{-1}.word, h_{-1}.tag, h_0.word, h_0.tag$ for $m=$ PARSER;
  \end{itemize}
\end{itemize}
For each given $(W,T)$ which satisfies the requirements in
section~\ref{section:basic_idea} there is a unique derivation
$d(W,T)$. The converse is not true, namely not every derivation
corresponds to a correct $(W,T)$; however, the
constraints in section~\ref{prob_model_requirements} ensure that these
derivations receive 0 probability.

The probability of a $(W,T)$ sequence is obtained by chaining the
probabilities of the elementary events in its derivation, as described in
section~\ref{section:prob_model}:
\begin{eqnarray}
  P(W,T) = P(d(W,T)) = \prod_{i=1}^{length(d(W,T))} p(e_i)
\end{eqnarray}

The probability of an elementary event is calculated using the
smoothing technique presented in
section~\ref{subsection:modeling_tools} and repeated here for clarity
of explanation:
\begin{eqnarray} \label{eqn:smoothing_reestimation}
\lefteqn{P_n(u|z_1,\ldots,z_n) =} \nonumber\\
&\lambda(z_1,\ldots,z_n) \cdot P_{n-1}(u|z_1,\ldots,z_{n-1}) + (1 - \lambda(z_1, \ldots ,z_n)) \cdot f_n(u|z_1, \ldots ,z_n),\\
\lefteqn{P_{-1}(u) = uniform(\mathcal{U})}
\end{eqnarray}

\begin{itemize}
\item  $z_1,\ldots,z_n$ is the context of order $n$ when predicting
  $u$; $\mathcal{U}$ is the vocabulary in which $u$ takes values;
\item  $f_k(u|z_1,\ldots,z_k)$ is the order-k relative frequency estimate for the
conditional probability $P(u|z_1,\ldots,z_k)$:
\begin{eqnarray*}
  f_k(u|z_1,\ldots,z_k) & = & C(u,z_1,\ldots,z_k)/C(z_1,\ldots,z_k),k = 0 \ldots n,\\
  C(u,z_1,\ldots,z_k)   & = & \sum_{z_{k+1} \in \mathcal{Z}} \ldots
  \sum_{z_{n} \in \mathcal{Z}} C(u,z_1,\ldots,z_k, z_{k+1} \ldots z_n),\\
  C(z_1,\ldots,z_k)     & = & \sum_{u \in \mathcal{U}} C(u,z_1,\ldots,z_k),\\
\end{eqnarray*}
\item  $\lambda_k$ are the interpolation
  coefficients satisfying $0 < \lambda_k < 1,  k = 0 \ldots n$.
\end{itemize}

The $\lambda(z_1,\ldots,z_k)$ coefficients are grouped into
equivalence classes --- ``tied'' --- based on the range into which the count
$C(z_1, \ldots , z_k)$ falls; the count ranges for each equivalence
class are set such that a statistically sufficient number of events
$(u|z_1,\ldots,z_k)$ fall in that range.

The parameters of a given model component $m$ are:
\begin{itemize}
\item the maximal order counts $C^{(m)}(u,z_1,\ldots,z_n)$;
\item the count ranges for grouping the interpolation values into
  equivalence classes --- ``tying'';
\item the interpolation value for each equivalence class;
\end{itemize}
Assuming that the count ranges and the corresponding interpolation
values for each order are kept fixed to their initial values --- see
section~\ref{section:initial_parameters} --- the only parameters to be
reestimated using the EM algorithm are the maximal order counts
$C^{(m)}(u,z_1,\ldots,z_n)$ for each model component.

In order to avoid traversing the entire hidden space for a given
observed word sequence\footnote{normally required in the E-step} we use the ``N-best''
training approach presented in section~\ref{section:nbest_training}
for which the sampling strategy is the same as the pruning strategy
presented in section~\ref{section:pruning}.

The derivation of the reestimation formulas is presented in
appendix~\ref{app:slm_reest}. The E-step is the one presented in
section~\ref{section:detailed_nbest_training}; the M-step takes into
account the smoothing technique presented above
(equation~(\ref{eqn:smoothing_reestimation})).

Note that due to both the smoothing involved in the
M-step and the fact that the set of sampled ``N-best'' hidden events --- parses --- 
are reevaluated at each iteration we allow new maximal order events to 
appear in each model component while discarding others. 
\emph{Not only are we estimating the counts of maximal order n-gram
  events in each model component}
 --- WORD-PREDICTOR, TAGGER, PARSER
--- \emph{but we also allow the distribution on types to change from
  one iteration to the other}. This is because the set of hidden events
allowed for a given observed word sequence is not invariant --- as it
is the case in regular EM. For example, the count set that describes
the WORD-PREDICTOR component of the model to be used at the next
iteration is going to have a different n-gram composition than that
used at the current iteration. This change is presented in the
experiments section, see Table~\ref{table:type}.

\subsection{First Stage Initial Parameters}\label{section:initial_parameters}

Each model component --- WORD-PREDICTOR, TAGGER, PARSER ---
is initialized from a set of hand-parsed sentences --- in this case
are going to use the UPenn Treebank manually annotated sentences --- after undergoing
headword percolation and binarization, as explained in
section~\ref{section:headword_percolation}. This is a subset ---
approx.\ 90\% --- of the training data. 
Each parse tree $(W,T)$ is then decomposed into its derivation $d(W,T)$.
Separately for each $m$ model component, we:
\begin{itemize}
\item gather joint counts $C^{(m)}(u^{(m)}, {\underline{z}}^{(m)})$ from the derivations that
make up the ``development data'' using $\rho(W,T) = 1$ (see appendix ~\ref{app:slm_reest});
\item estimate the interpolation coefficients on joint
counts gathered from ``check data'' --- the remaining 10\% of the
training data ---  using the EM algorithm~\cite{em77}.

These are the initial parameters used with the reestimation procedure
described in the previous section.
\end{itemize}

\section{Second Stage Parameter Reestimation}\label{section:model_reest_second_pass}

In order to improve performance, we develop a model to be used
strictly for word prediction in~(\ref{eq:ppl1}), different from the WORD-PREDICTOR
model~(\ref{eq:word_predictor_prob}). We will call this new component the
L2R-WORD-PREDICTOR.

The key step is to recognize in~(\ref{eq:ppl1}) a hidden Markov model
(HMM) with fixed transition probabilities --- although dependent on the
position in the input sentence $k$ --- specified by the
$\rho(W_{k},T_{k})$ values. 

The E-step of the  EM algorithm~\cite{em77} for gathering
joint counts  $C^{(m)}(y^{(m)}, {\underline{x}}^{(m)})$, $m = $
L2R-WORD-PREDICTOR-MODEL, is the standard one whereas the
M-step uses the same count smoothing technique as that
described in section~\ref{section:model_reest_first_pass}. 

The second reestimation pass is seeded with the $m = $ WORD-PREDICTOR model
joint counts  $C^{(m)}(y^{(m)}, {\underline{x}}^{(m)})$ resulting from
the first parameter reestimation pass (see section~\ref{section:model_reest_first_pass}).

\chapter{Experiments using the Structured Language Model}
\label{chapter:slm_experiments}

For convenience, we chose to work on the UPenn
Treebank corpus~\cite{Upenn} --- a subset of the WSJ (Wall Street
Journal) corpus. 
The vocabulary sizes were:\\
word vocabulary: 10k, open --- all words outside the vocabulary are
mapped to the \verb+<unk>+ token; POS tag vocabulary: 40, closed;
non-terminal tag vocabulary: 52, closed; parser operation vocabulary:
107, closed. The training data was split into development set
(929,564wds (sections 00-20)), check set (73,760wds (sections 21-22))
and the test data consisted of 82,430wds (sections 23-24). The ``check''
set was used strictly for initializing the model parameters as
described in section~\ref{section:initial_parameters}; the
``development'' set was used with the reestimation techniques described in
chapter~\ref{chapter:model_reest}.

\section{Perplexity Results} \label{section:experiments}

Table~\ref{table:reest_ppls} shows the results of the reestimation techniques; \verb+E0-3+ and
\verb+L2R0-5+ denote iterations of the reestimation procedure
described in sections~\ref{section:model_reest_first_pass} and
\ref{section:model_reest_second_pass}, respectively. A deleted
interpolation  trigram model derived from the same training data had
perplexity 167.14 on the same test data.
\begin{center}
  \begin{table}[h]
     \begin{center}
       \begin{tabular}{||l|c|c||} \hline
         iteration & DEV set & TEST set      \\ 
         number    & L2R-PPL & L2R-PPL       \\ \hline \hline
         E0        & 24.70   &  167.47       \\ \hline 
         E1        & 22.34   &  160.76       \\ \hline 
         E2        & 21.69   &  158.97       \\ \hline 
         E3 = L2R0 & 21.26   &  158.28       \\ \hline 
         L2R5      & 17.44   &  153.76       \\ \hline
       \end{tabular}
     \end{center}
     \caption{Parameter reestimation results} \label{table:reest_ppls}
   \end{table}
\end{center}
Simple linear interpolation between our model and the trigram model:
\begin{eqnarray*}
Q(w_{k+1}/W_k) = \lambda \cdot P(w_{k+1}/w_{k-1}, w_k) + (1 - \lambda) \cdot P(w_{k+1}/W_k) 
\end{eqnarray*} 
yielded a further improvement in PPL, as shown in
Table~\ref{table:interpolated_PPL}. The interpolation weight was
estimated on check data to be $\lambda = 0.36$. An overall relative
reduction of 11\% over the trigram model has been achieved.
\begin{center}
  \begin{table}[h]
     \begin{center}
       \begin{tabular}{||l|c|c||} \hline
         iteration & TEST set & TEST set\\ 
         number    & L2R-PPL  & 3-gram interpolated PPL   \\ \hline \hline
         E0        &  167.47  & 152.25  \\ \hline 
         E3        &  158.28  & 148.90  \\ \hline 
         L2R5      &  153.76  & 147.70  \\ \hline
       \end{tabular}
     \end{center}
     \caption{Interpolation with trigram results} \label{table:interpolated_PPL}
   \end{table}
\end{center}

As outlined in section~\ref{section:word_level_ppl}, the perplexity
value calculated using~(\ref{eq:ppl0}):
\begin{eqnarray}
  P(W|T^*) & = & \prod_{k=0}^n P(w_{k+1}|W_{k}T_{k}^*), \nonumber
  T^* = argmax_{T}P(W,T)
\end{eqnarray}
is a lower bound for the achievable perplexity of our model; for the above search parameters and E3
model statistics this bound was 99.60, corresponding to a relative reduction of
41\% over the trigram model. This suggests that a better
parameterization in the PARSER model --- one that reduces the entropy
$H(\rho(T_k|W_k))$ of guessing the ``good'' parse given the word
prefix --- can lead to a better model. 
Indeed, as  we already pointed out, the trigram model is a particular case of our model
for which the parse is always right branching and we have no POS/NT
tag information, leading to $H(\rho(T_k|W_k)) = 0$ and a standard 3-gram
WORD-PREDICTOR. The 3-gram model is thus an extreme case of the
structured language model: one for which the ``hidden'' structure is
a \emph{function of the word prefix}. Our result shows that better models can
be obtained by allowing richer ``hidden'' structure --- parses --- and 
that a promising direction of research may be to find the best compromise
between the predictive power of the WORD-PREDICTOR ---
measured by $H(w_{k+1}|T_k,W_k))$--- and the ease of
guessing the hidden structure $T_k|W_k$ --- measured by
$H(\rho(T_k|W_k))$ --- on which the WORD-PREDICTOR operation is
based. 
A better solution would be a maximum entropy PARSER model which incorporates a
richer set of predictors in a better way than the deleted
interpolation scheme we are using. Due to the computational problems
faced by such a model we have not pursued this path
although we consider it a very promising one.

\subsection{Comments and Experiments on Model Parameters Reestimation} \label{section:loose_reest}

The word level probability assigned to a training/test set by our model is calculated
using the proper word-level probability assignment in
equation~(\ref{eq:ppl1}). An alternative which leads to a deficient
probability model is to sum over all the complete parses that survived
the pruning strategy, formalized in equation~(\ref{eq:ppl2}). Let the
likelihood assigned to a corpus $\mathcal{C}$ by our model
$P_{\theta}$ be denoted by:
\begin{itemize}
\item ${\mathcal{L}}^{L2R}({\mathcal{C}},P_{\theta})$, where
  $P_{\theta}$ is calculated using~(\ref{eq:ppl1}), repeated here for
  clarity:
  \begin{eqnarray}
    P(w_{k+1}|W_{k}) & = & \sum_{T_{k}\in 
      S_{k}}P(w_{k+1}|W_{k}T_{k})\cdot\rho(W_{k},T_{k}),\nonumber \\
    \rho(W_{k},T_{k}) & = & P(W_{k}T_{k})/\sum_{T_{k} \in
      S_{k}}P(W_{k}T_{k}) \nonumber
  \end{eqnarray}
  Note that this is a proper probability model.
\item ${\mathcal{L}}^{N}({\mathcal{C}},P_{\theta})$, where
  $P_{\theta}$ is calculated using~(\ref{eq:ppl2}):
  \begin{eqnarray}
    P(W) = \sum_{k=1}^N P(W,T^{(k)}) \nonumber
  \end{eqnarray}
  This is a deficient probability model.
\end{itemize}

One seeks to devise an algorithm that finds the model parameter values
which maximize the likelihood of a test corpus. This is an unsolved
problem; the standard approach is to resort to maximum likelihood
estimation techniques on the training corpus and make provisions that
will ensure that the increase in likelihood on training data carries
over to unseen test data.

As outlined previously, the estimation procedure of the SLM parameters
takes place in two stages:
\begin{enumerate}
\item the ``N-best training'' algorithm~(see
  Section~\ref{section:model_reest_first_pass}) is employed to
  increase the training data ``likelihood''
  ${\mathcal{L}}^{N}({\mathcal{C}},P_{\theta})$.  The initial
  parameters for this first estimation stage are gathered from a
  treebank. The perplexity is still evaluated using the formula in
  Eq.~(\ref{eq:ppl1}).
\item estimate a separate L2R-WORD-PREDICTOR model such that
  ${\mathcal{L}}^{L2R}({\mathcal{C}},P_{\theta})$ is increased --- see
  Eq.~(\ref{eq:ppl1_l2r}). The initial parameters for the
  L2R-WORD-PREDICTOR component are obtained by copying the
  WORD-PREDICTOR estimated at stage one.
\end{enumerate}

As explained in Section~\ref{section:loose_reest}, the ``N-best
training'' algorithm is employed to increase the training data
``likelihood'' ${\mathcal{L}}^{N}({\mathcal{C}},P_{\theta})$; we rely
on the consistency of the probability estimates underlying the
calculation of the two different likelihoods to correlate the increase
in ${\mathcal{L}}^{N}({\mathcal{C}},P_{\theta})$ with the desired
increase of ${\mathcal{L}}^{L2R}({\mathcal{C}},P_{\theta})$.

To be more specific, ${\mathcal{L}}^{N}({\mathcal{C}},P_{\theta})$ and
${\mathcal{L}}^{L2R}({\mathcal{C}},P_{\theta})$ are calculated using
the probability assignments in Eq.~(\ref{eq:ppl2}) ---
deficient --- and Eq.~(\ref{eq:ppl1}), respectively. Both
probability estimates are consistent in the sense that if we summed
over all the parses $T$ for a given word sequence $W$ they would yield
the correct probability $P(W)$ according to our model. Although there
is no formal proof, there are reasons to believe that the N-best
reestimation procedure should not decrease the
${\mathcal{L}}^{N}({\mathcal{C}},P_{\theta})$ likelihood \footnote{It
  is very similar to a rigorous EM approach} but no claim can be made
about the increase in the
${\mathcal{L}}^{L2R}({\mathcal{C}},P_{\theta})$ likelihood --- which
is the one we are interested in. Our experiments show that the
increase in ${\mathcal{L}}^{N}({\mathcal{C}},P_{\theta})$ is
correlated with an increase in
${\mathcal{L}}^{L2R}({\mathcal{C}},P_{\theta})$, a key factor in this
being a good heuristic search strategy --- see Section~\ref{section:pruning}.
Table~\ref{table:ppl_evolution} shows the evolution of different
``perplexity'' values during N-best reestimation. L2R-PPL is
calculated using the proper probability assignment in
Eq.(\ref{eq:ppl1}). TOP-PPL and BOT-PPL are calculated using the
probability assignment in Eq.(\ref{eq:ppl0}), where $T^* =
argmax_{T}P(W,T)$ and $T^* = argmin_{T}P(W,T)$, respectively --- the
search for $T^*$ being carried out according to our pruning strategy;
we condition the word predictions on the topmost and
bottom-most parses present in the stacks after parsing the entire
sentence. SUM-PPL is calculated using the deficient probability
assignment in Eq.(\ref{eq:ppl2}). It can be noticed that TOP-PPL
and BOT-PPL stay almost constant during the reestimation process;
The value of TOP-PPL is slightly increasing and that of BOT-PPL is
slightly decreasing. As expected, the value of the SUM-PPL decreases and
its decrease is correlated with that of the L2R-PPL. 
\begin{table}[ht]
  \begin{center}
    \begin{tabular}{|l|r|r|r|}\hline
      ``Perplexity'' & \multicolumn{2}{c|}{Iteration} & Relative Change\\
                     & E0         & E3                & \\\hline
      TOP-PPL        & 97.5       & 99.3              & +1.85\%  \\\hline
      BOT-PPL        & 107.9      & 106.2             & -1.58\%  \\\hline
      SUM-PPL        & 195.1      & 175.5             & -10.05\% \\\hline\hline
      L2R-PPL        & 167.5      & 158.3             & -5.49\%  \\ \hline
    \end{tabular}
    \caption{Evolution of different "perplexity" values during training}
    \label{table:ppl_evolution}
  \end{center}
\end{table}

It is very important to note that due to both the smoothing involved
in the M-step --- imposed by the smooth parameterization of the
model\footnote{Unlike standard parameterizations, we do not reestimate
  the relative frequencies from which each component probabilistic
  model is derived; that would lead to a shrinking or, at best, fixed
  set of events} --- and the fact that the set of sampled ``N-best''
hidden events --- parses --- are reevaluated at each iteration, we
allow new maximal order events to appear in each model component while
discarding others.  \emph{Not only are we estimating the counts of
  maximal order n-gram events in each model component} ---
WORD-PREDICTOR, TAGGER, PARSER --- \emph{but we also allow the
  distribution on types to change from one iteration to the other}.
This is because the set of hidden events allowed for a given observed
word sequence is not invariant. For example, the count set that
describes the WORD-PREDICTOR component of the model to be used at the
next iteration may have a different n-gram composition than that used
at the current iteration.

We evaluated the change in the distribution on types\footnote{A type
  is a particular value, regarded as one entry in the alphabet spanned
  by a given random variable} of the maximal order events $(y^{(m)},
{\underline{x}}^{(m)})$ from one iteration to the next.
Table~\ref{table:type} shows the dynamics of the set of types of the
different order events during the reestimation process for the
WORD-PREDICTOR model component. Similar dynamics were observed for the
other two components of the model.  The equivalence classifications
corresponding to each order is:
\begin{itemize}
\item $\underline{z}=h_0.tag, h_0.word, h_{-1}.tag, h_{-1}.word$ for
  order 4;
\item $\underline{z}=h_0.tag, h_0.word, h_{-1}.tag$ for order 3;
\item $\underline{z}=h_0.tag, h_0.word$ for order 2;
\item $\underline{z}=h_0.tag$ for order 1;
\end{itemize}
An event of order 0 consists of the predicted word only.
\begin{center}
  \begin{table}[h]
     \begin{center}
       \begin{tabular}{|l|r|r|r|r|r|r|} \hline
         iteration & no.\ tokens & \multicolumn{5}{c|}{no.\ types for order}\\ \hline
                   &         &     0 &      1 &       2   &       3 &       4 \\ \hline
         E0        & 929,564 & 9,976 & 77,225 & 286,329   & 418,843 & 591,505 \\
         E1        & 929,564 & 9,976 & 77,115 & 305,266   & 479,107 & 708,135 \\
         E2        & 929,564 & 9,976 & 76,911 & 305,305   & 482,503 & 717,033 \\
         E3        & 929,564 & 9,976 & 76,797 & 307,100   & 490,687 & 731,527 \\ \hline 
         L2R0 (=E3)& 929,564 & 9,976 & 76,797 & 307,100   & 490,687 & 731,527 \\
         L2R1-5    & 929,564 & 9,976 & 257,137 &  2,075,103   & 3,772,058  &  5,577,709 \\ \hline 
       \end{tabular}
   \end{center}
   \caption{Dynamics of WORD-PREDICTOR distribution on types during reestimation} \label{table:type}
 \end{table}
\end{center}
The higher order events --- closer to the root of the linear
interpolation scheme in Figure~\ref{fig:mixing_scheme} --- become more
and more diverse during the first estimation stage, as opposed to the
lower order events. This shows that the ``N-best'' parses for a given
sentence change from one iteration to the next. Although the E0 counts
were collected from ``1-best'' parses --- binarized treebank parses
--- the increase in number of maximal order types from E0 to E1 ---
collected from ``N-best'', N = 10 --- is far from dramatic, yet higher
than that from E1 to E2 --- both collected from ``N-best'' parses.

The big increase in number of types from E3 (=L2R0) to L2R1 is due to
the fact that at each position in the input sentence, WORD-PREDICTOR
counts are now collected for all the parses in the stacks, many of
which do not belong to the set of N-best parses for the {\em complete}
sentence used for gathering counts during E0-3.

Although the perplexity on test data still decreases during the second
reestimation stage --- we are not over-training --- this decrease is
very small and not worth the computational effort if the model is
linearly interpolated with a 3-gram model, as shown in
Table~\ref{table:interpolated_PPL}. Better integration of the 3-gram
and the head predictors is desirable.

\section{Miscellaneous Other Experiments}

\subsection{Choosing the Model Components Parameterization} 
\label{section:choosing_params}
 
The experiments presented in \cite{chelba97} show the usefulness of
the two most recent exposed heads for word prediction.  The same
criterion --- conditional perplexity --- can be used as a guide in
selecting the parameterization of each model component:
WORD-PREDICTOR, TAGGER, PARSER.  For each model component we
gather the counts from the UPenn Treebank as explained in
Section~\ref{section:initial_parameters}.  The relative frequencies
are determined from the ``development'' data, the interpolation
weights estimated on ``check'' data --- as described in
Section~\ref{section:initial_parameters}. We then test each model
component on counts gathered from the ``test'' data.
Note that the smoothing scheme described in
Section~\ref{subsection:modeling_tools} discards elements of the context
$\underline{z}$ from right to left.

\subsubsection{Selecting the WORD-PREDICTOR Equivalence Classification}
 
The experiments in \cite{chelba97} were repeated using deleted
interpolation as a modeling tool and the training/testing setup
described above. The results for different equivalence classifications
of the word-parse k-prefix $(W_k, T_k)$ are presented in
Table~\ref{table:_w}.
 \begin{table}[ht]
   \begin{center}
     \begin{tabular}{|ll|c|c|}\hline
       \multicolumn{2}{|c|}{Equivalence Classification} & Cond. PPL & Voc. Size\\\hline
       HH & $\underline{z}=h_0.tag, h_0.word, h_{-1}.tag, h_{-1}.word$ &
       115 & 10,000\\
       WW & $\underline{z}=w_{-1}.tag, w_{-1}.word, w_{-2}.tag, w_{-2}.word$ &
       156 & 10,000\\
       hh & $\underline{z}=h_0.word, h_{-1}.word$ &
       154 & 10,000\\
       ww & $\underline{z}=w_{-1}.word, w_{-2}.word$ &
       167 & 10,000 \\  \hline
     \end{tabular}
     \caption{WORD-PREDICTOR conditional perplexities}
     \label{table:_w}
   \end{center}
 \end{table}
 The different equivalence classifications of the word-parse k-prefix
 retain the following predictors:
 \begin{enumerate}
 \item ww: the two previous words --- regular 3-gram model;
 \item hh: the two most recent exposed headwords --- no POS/NT label
   information;
 \item WW: the two previous exposed words along with their POS tags;
 \item HH: the two most recent exposed heads --- headwords along with
   their NT/POS labels;
 \end{enumerate}
 It can be seen that the most informative predictors for the next word
 are the exposed heads --- HH model. Except for the ww
 model\footnote{regular 3-gram model}, none of the others is a valid
 word-level perplexity since it conditions the prediction on hidden
 information (namely the tags present in the treebank parses); the
 entropy of guessing the hidden information would need to be factored
 in.

\subsubsection{Selecting the TAGGER Equivalence Classification}
 
 The results for different equivalence classifications of the
 word-parse k-prefix $(W_k, T_k)$ for the TAGGER model are presented
 in Table~\ref{table:_t}.
 \begin{table}[ht]
   \begin{center}
     \begin{tabular}{|ll|c|c|}\hline
       \multicolumn{2}{|c|}{Equivalence Classification} & Cond. PPL & Voc. Size \\\hline
       HHw & $\underline{z}=w_k, h_0.tag, h_0.word, h_{-1}.tag, h_{-1}.word$ &
       1.23 & 40\\
       WWw & $\underline{z}=w_k, w_{-1}.tag, w_{-1}.word, w_{-2}.tag, w_{-2}.word$ &
       1.24 & 40\\
       ttw & $\underline{z}=w_k, h_0.tag, h_{-1}.tag$ &
       1.24 & 40\\  \hline
     \end{tabular}
     \caption{TAGGER conditional perplexities}
     \label{table:_t}
   \end{center}
 \end{table}
 The different equivalence classifications of the word-parse k-prefix
 retain the following predictors:
 \begin{enumerate}
 \item WWw: the two previous exposed words along with their POS tags
   and the word to be tagged;
 \item HHw: the two most recent exposed heads --- headwords along with
   their NT/POS labels and the word to be tagged;
 \item ttw: the NT/POS labels of the two most recent exposed heads and
   the word to be tagged;
 \end{enumerate}
 It can be seen that among the equivalence classifications considered,
 none performs significantly better than the others, and the
 prediction of the POS tag for a given word is a relatively easy task
 --- the conditional perplexities are very close to one.
 Because of its simplicity, we chose to work with the ttw equivalence classification.

\subsubsection{Selecting the PARSER Equivalence Classification}

 The results for different equivalence classifications of the
 word-parse k-prefix $(W_k, T_k)$ for the PARSER model are presented
 in Table~\ref{table:_p}.
 \begin{table}[ht]
   \begin{center}
     \begin{tabular}{|ll|c|c|}\hline
       \multicolumn{2}{|c|}{Equivalence Classification} & Cond. PPL & Voc. Size\\ \hline
       HH  & $\underline{z}= h_0.tag, h_0.word, h_{-1}.tag, h_{-1}.word$ &
       1.68 & 107\\
       hhtt & $\underline{z}= h_0.tag, h_{-1}.tag, h_0.word, h_{-1}.word$ &
       1.54 & 107\\
       tt  & $\underline{z}= h_0.tag, h_{-1}.tag$ &
       1.71 & 107\\ \hline
     \end{tabular}
     \caption{PARSER conditional perplexities}
     \label{table:_p}
   \end{center}
 \end{table}
 The different equivalence classifications of the word-parse k-prefix
 retain the following predictors:
 \begin{enumerate}
 \item HH: the two most recent exposed heads --- headwords along with
   their NT/POS labels and the word to be tagged;
 \item hhtt: same as HH just that the backing-off order is changed;
 \item ttw: the NT/POS labels of the two most recent exposed heads;
 \end{enumerate}
 It can be seen that the presence of headwords improves the accuracy
 of the PARSER component; also, the backing-off order of the predictors is
 important --- hhtt vs.\ HH. We chose to work with the hhtt
 equivalence classification.

\subsection{Fudged TAGGER and PARSER Scores}
\label{section:fudged}

The probability values for the three model components fall into
different ranges. As pointed out at the beginning of this chapter, the
WORD-PREDICTOR vocabulary is of the order of thousands whereas the
TAGGER and PARSER have vocabulary sizes of the order of tens. This
leads to the undesirable effect that the contribution of the TAGGER
and PARSER to the overall probability of a given partial parse
$P(W,T)$ is very small compared to that of the WORD-PREDICTOR. We
explored the idea of bringing the probability values into the same
range by \emph{fudging} the TAGGER and PARSER probability values,
namely:
\begin{eqnarray}
  \label{eq:fudged}
  \lefteqn{P(W,T) =}\nonumber \\
  & & \prod_{k=1}^{n+1}[P(w_k|W_{k-1}T_{k-1}) \cdot
  {\left\{P(t_k|W_{k-1}T_{k-1},w_k) \cdot P(T_{k-1}^k|W_{k-1}T_{k-1},w_k,t_k)\right\}}^{\gamma}]\\
  \lefteqn{P(T_{k-1}^k|W_{k-1}T_{k-1}) =
  \prod_{i=1}^{N_k}P(p_i^k|W_{k-1}T_{k-1},w_k,t_k,p_1^k\ldots p_{i-1}^k)}
\end{eqnarray}
where $\gamma$ is the fudge factor. For $\gamma \neq 1.0$ we do not
have a valid probability assignment anymore, however the L2R-PPL
calculated using Eq.~(\ref{eq:ppl1}) is still a valid word-level
probability assignment due to the re-normalization of the interpolation
coefficients. Table~\ref{table:fudged} shows the PPL values calculated
using Eq.~(\ref{eq:ppl1}) where $P(W,T)$ is calculated using
Eq.~(\ref{eq:fudged}). As it can be seen the optimal fudge factor
turns out to be 1.0, corresponding to the \emph{correct} calculation
of the probability $P(W,T)$.
\begin{table}[ht]
  \begin{center}
    \begin{tabular}{l|rrrrrrrrrrrrr}
      fudge & 0.01 & 0.02 & 0.05 & 0.1 & 0.2 & 0.5 & 1.0 & 2.0 & 5.0
      & 10.0 & 20.0 & 50.0 & 100.0 \\ \hline
      PPL   & 341 & 328 & 296 & 257 & 210 & 168 & \underline{167}
      & 189 & 241 & 284 & 337 & 384 & 408
    \end{tabular}
    \caption{Perplexity Values: Fudged TAGGER and PARSER}
    \label{table:fudged}
  \end{center}
\end{table}

\subsection{Maximum Depth Factorization of the Model} 
\label{section:max_depth_factorization}

The word level probability assignment used by the SLM ---
Eq.~(\ref{eq:ppl1}) --- can be thought of as a model factored over
different maximum reach depths. Let $D(T_k)$ be the ``depth'' in the
word-prefix $W_k$ at which the headword $h_{-1}.word$ can be found.

Eq.~(\ref{eq:ppl1}) can be rewritten as:
\begin{eqnarray}
  P(w_{k+1}|W_{k}) & = &
  \sum_{d=0}^{d=k}P(d|W_{k})\cdot P(w_{k+1}|W_{k},d),\label{eq:ppl_depth}\\
  \textrm{where:} & & \nonumber \\
  P(d|W_{k}) & = & \sum_{T_{k}\in S_{k}} \rho(W_{k},T_{k}) \cdot 
  \delta(D(T_k),d)\nonumber\\
  P(w_{k+1}|W_{k},d) & = & \sum_{T_{k}\in S_{k}}
  P(T_{k}|W_{k},d)\cdot P(w_{k+1}|W_{k},T_{k})\nonumber\\
  P(T_{k}|W_{k},d) & = & \rho(W_{k},T_{k}) \cdot \delta(D(T_k),d) / P(d|W_{k})\nonumber
\end{eqnarray}

We can interpret Eq.~(\ref{eq:ppl_depth}) as a linear interpolation of
models that reach back to different depths in the word prefix $W_k$. The
expected value of $D(T_k)$ shows how far does the SLM reach in the
word prefix:
\begin{eqnarray}
  \label{eq:expected_depth}
  {E_{SLM}}[D] & = & 1/N \sum_{k=0}^{k=N} \sum_{d=0}^{d=k} d \cdot P(d|W_{k})
\end{eqnarray}

For the 3-gram model we have ${E_{3-gram}}[D] = 2$.
We evaluated the expected depth of the SLM using the formula in
Eq.~(\ref{eq:expected_depth}). The results are presented in
Table~\ref{table:slm_depth}.
\begin{center}
  \begin{table}[h]
     \begin{center}
       \begin{tabular}{||l|c|c||}    \hline
         iteration  &  expected depth\\ 
         number     &  E[D]      \\ \hline \hline
         E0         &  3.35    \\ 
         E1         &  3.46    \\ 
         E2         &  3.45    \\ \hline
       \end{tabular}
     \end{center}
     \caption{Maximum Depth Evolution During Training} \label{table:slm_depth}
   \end{table}
\end{center}
It can be seen that the memory of the SLM is considerably higher than
that of the 3-gram model --- whose depth is 2.

\begin{figure}[h]
  \epsfig{file=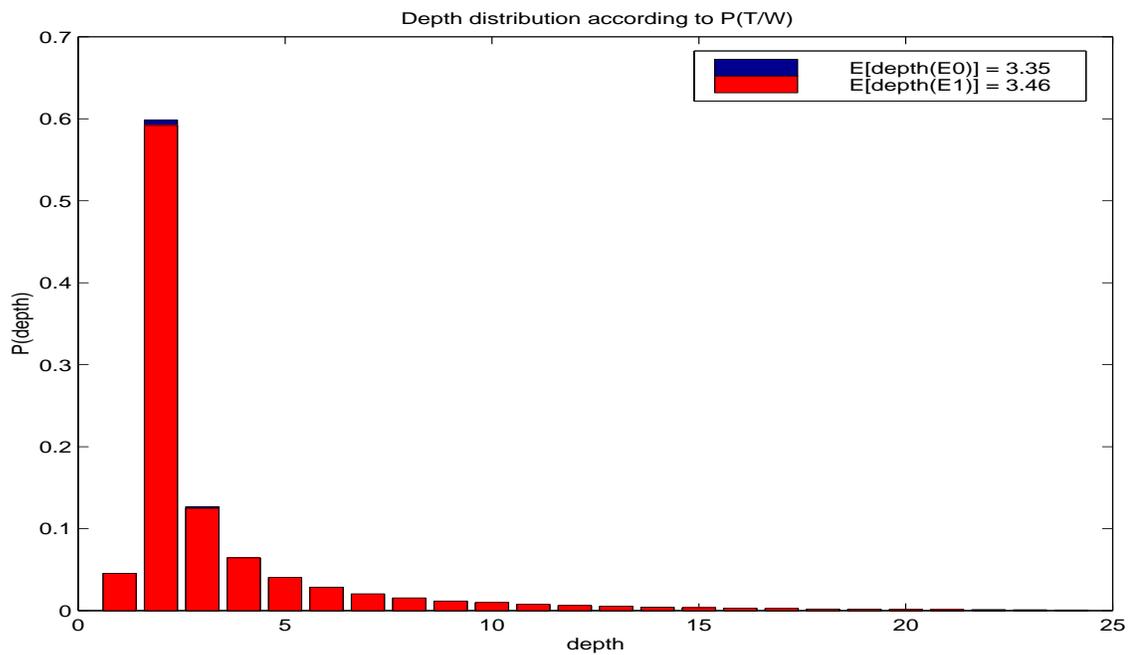,height=9cm,width=15cm}
  \caption{Structured Language Model Maximum Depth Distribution} 
  \label{fig:depth_distrib}
\end{figure}

Figure~\ref{fig:depth_distrib} shows
\footnote{The nonzero value of $P(1|W)$
  is due to the fact that the prediction of the first word in a
  sentence is based on context of length 1 in both
  SLM and 3-gram models} the distribution $P(d|W_k)$, averaged
over all positions $k$ in the test string:
$$
P(d|W) = 1/N \sum_{k=1}^{N} P(d|W_k)
$$

It can be seen that the SLM makes a prediction which reaches farther
than the 3-gram model in about 40\% of cases, on the average.

\chapter{$A^*$ Decoder for Lattices} \label{chapter:a_star_lattice_decoder}

\section{Two Pass Decoding Techniques}

In a two-pass recognizer, a computationally cheap decoding step is run
in the first pass, a set of hypotheses is retained as an intermediate
result and then a more sophisticated recognizer is run over these in a
second pass --- usually referred to as the rescoring pass. The search
space in the second pass is much more restricted compared to the first
pass so we can afford using better --- usually also computationally
more intensive ---  acoustic and/or language models.

The two most popular two-pass strategies differ mainly in the number
of intermediate hypotheses saved after the first pass and the form in
which they are stored. 

In the so-called ``N-best\footnote{The value
of N is typically 100--1000} rescoring'' method, a list of complete
hypotheses along with acoustic/language model scores are retained and
then rescored using more complex acoustic/language models. 

Due to the limited number of hypotheses in the N-best list, the second
pass recognizer might be too constrained by the first pass so a more
comprehensive list of hypotheses is often needed. The alternative
preferred to N-best list rescoring is ``lattice rescoring''. The
intermediate format in which the hypotheses are stored is now a
directed acyclic graph in which the nodes are a subset of the language
model states in the composite hidden Markov model and the arcs are labeled with
words. Typically, the first pass acoustic/language model scores
associated with each arc --- or link --- in the lattice are saved and
the nodes contain time alignment information.

For both cases one can calculate the ``oracle'' word error rate: the
word error rate along the hypothesis with the minimum number of
errors. The oracle-WER decreases with the number of hypotheses saved.

Of course, a set of N-best hypotheses can be assembled as a lattice,
the difference between the two being just in the number of different
hypotheses --- with different time-alignments --- stored in the
lattice. One reason which makes the N-best rescoring framework
attractive is the possibility to use ``whole sentence'' language
models: models that are able to assign a score only to complete
sentences due to the fact that they do not operate in a left-to-right
fashion. The drawbacks are that the number of hypotheses explored is too 
small and their quality reminiscent of the models used in the first
pass. To clarify the latter assertion, assume that the second pass
language model to be applied is dramatically different from the one
used in the first pass and that if we afforded to extract the N-best using
the better language model they would have a different kind of errors,
specific to this language model. In that case simple rescoring of the
N-best list generated using the weaker language model may constrain
too much the stronger language model used in the second pass, not
allowing it to show its merits.

It is thus desirable to have a sample of the possible word hypotheses
which is as complete as possible --- not biased towards a given model
--- and at the same time of manageable size. This is what makes
lattice rescoring the chosen method in our case, hoping that simply
by increasing the number of hypotheses retained one reduces the bias
towards the first pass language model.

\section{$A^*$ Algorithm} \label{astar_theory}

The $A^*$ algorithm~\cite{astar} is a tree search strategy that could be compared
to depth-first tree-traversal: pursue the most promising path as
deeply as possible.

Let a set of hypotheses
$$L=\{h:x_1,\ldots, x_n\},\ x_i \in \mathcal{W}^*\ \forall\ i$$
be organized as a prefix tree. We wish to obtain the maximum scoring
hypothesis under the scoring function $f:\mathcal{W}^* \rightarrow \Re$: 
$$h^*=\arg\max_{h \in L}f(h)$$
without scoring all the hypotheses in $L$, if possible with a minimal
computational effort.

The algorithm operates with prefixes and suffixes of hypotheses in the 
set $L$; we will denote prefixes --- anchored at the root of the tree
--- with $x$ and suffixes --- anchored at a leaf --- with $y$. A
complete hypothesis $h$ can be regarded as the concatenation of a $x$
prefix and a $y$ suffix: $h = x.y$. We assume that the function
$f(\cdot)$ can be evaluated at any prefix $x$, i.e.\ $f(x)$ is a
meaningful quantity.

\begin{figure}[htbp]
  \begin{center}
    \epsfig{file=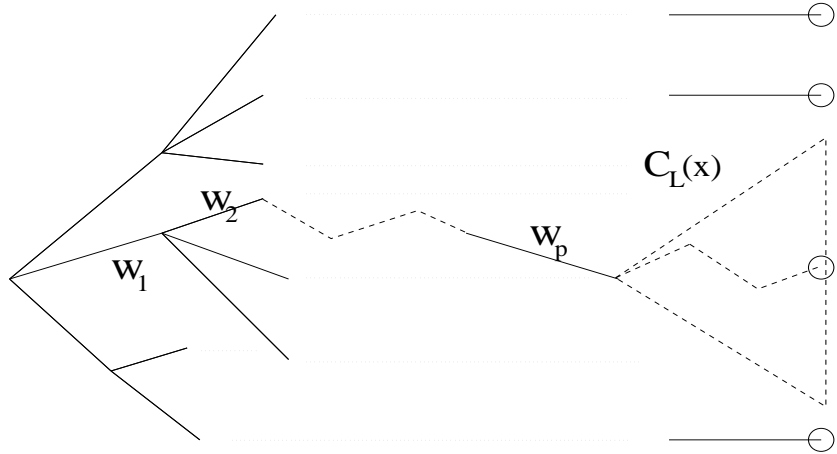, height=6cm,width=11cm}
    \caption{Prefix Tree Organization of a Set of Hypotheses L}
    \label{fig:prefix_tree}
  \end{center}
\end{figure}

To be able to pursue the most promising path, the algorithm needs to
evaluate all the possible suffixes for a given prefix
$x=w_1,\ldots,w_p$ that are allowed in $L$ --- see
figure~\ref{fig:prefix_tree}.  Let $C_L(x)$ be the set of suffixes
allowed by the tree for a prefix $x$ and assume we have an
overestimate for the $f(x.y)$ score of any \emph{complete} hypothesis $x.y$,
$g(x.y)$:
$$g(x.y) \doteq f(x) + h(y|x) \geq f(x.y)$$
Imposing the condition that $h(y|x) = 0$ for empty $y$, we have
$$g(x) = f(x), \forall\ complete\ x \in L$$ that is, the overestimate 
becomes exact for complete hypotheses $h \in L$.
Let the \emph{$A^*$ ranking function} $g_L(x)$ be defined as:
\begin{eqnarray}
  g_L(x) & \doteq & \max_{y \in C_L(x)} g(x.y) = f(x) + h_L(x),\ where
  \label{eq:g_overestimate}\\
  h_L(x) & \doteq & \max_{y \in C_L(x)} h(y|x)
\end{eqnarray}
$g_L(x)$ is an overestimate for the $f(\cdot)$ score of any complete
hypothesis that has the prefix $x$; the overestimate becomes exact for
complete hypotheses:

\begin{eqnarray}
  g_L(x) & \geq & f(x.y), \forall y \in C_L(x)\label{eq:astar1}\\
  g_L(h) & =    & f(h), \forall\ complete\ h \in L\label{eq:astar2}
\end{eqnarray}

The $A^*$ algorithm uses a potentially infinite stack\footnote{The
  stack need not be larger than $|L|=n$} in which prefixes $x$ are
ordered in decreasing order of the $A^*$ ranking function
$g_L(x)$\footnote{In fact any overestimate satisfying
both Eq.\ (\ref{eq:astar1}) and (\ref{eq:astar2}) will ensure
correctness of the algorithm}; 
at each extension step the top-most prefix $x=w_1,\ldots,w_p$ is popped
form the stack, expanded with all possible one-symbol continuations of
$x$ in $L$ and then all the resulting expanded prefixes --- among
which there may be complete hypotheses as well --- are inserted back
into the stack. The stopping condition is: whenever the popped
hypothesis is a complete one, retain it as the overall best
hypothesis $h^*$ --- see Algorithm~\ref{algorithm:a_star}.

\begin{algorithm}[h]
\begin{verbatim}
//empty_hypothesis;
//top_most_hypothesis;
//a_hypothesis;
insert empty_hypothesis in stack;

do
{ // one Astar extension step
    top_most_hypothesis = pop top-most hypothesis from stack;

    for all possible one symbol continuations w of top_most_hypothesis
    {
        a_hypothesis = expand top_most_hypothesis with w;
        insert a_hypothesis in stack;
    }
}while(top_most_hypothesis is incomplete)

//top_most_hypothesis is the highest f(.) scoring one
\end{verbatim}
\caption{$A^*$ search} \label{algorithm:a_star}
\end{algorithm}

The justification for the correctness of the algorithm lies in the 
fact that upon completion, any other prefix $x$ in the stack has a lower
stack-score than $h^*$:
$$g_L(x) < g_L(h^*)=f(h^*)$$
But $g_L(x) \geq f(x.y), \forall y \in C_L(x)$ which means that no
complete hypothesis $x.y$ could possibly result in a higher
$f(\cdot)$ score than $h^*$, formally: 
$$ f(x.y) \leq g_L(x) < g_L(h^*)=f(h^*),\ \forall x \in stack$$
Since the stack is infinite, it is guaranteed to contain prefixes
for all hypotheses $h \in L$ --- see Algorithm~\ref{algorithm:a_star}
---  which means that:
$$ f(x.y) \leq g_L(x) < g_L(h^*)=f(h^*),\ \forall x.y \in L$$

To get a better grasp of the workings of $A^*$ we examine two
limiting cases: perfect estimation of the scoring function $f()$ value
along the most promising suffix for any given prefix, and no clue at all.

In the first case we have $g(x.y) = f(x) + h(y|x) = f(x.y)$; notice that
the $A^*$ ranking function becomes
$g_L(x) = \max_{y \in C_L(x)} f(x.y), \forall y \in C_L(x)$, which means that we are able to
find the best continuation of the current prefix. This makes the
entire $A^*$ algorithm pointless: for $x$ being the empty hypothesis, we
just calculate $g_L(x)$ and retain the complete ``continuation'' $y=h^*$ that
yielded maximal $g_L(x)$. The $A^*$ algorithm simply builds $h^*$ by
traversing $y$ left to right; the topmost entry in the stack will
always have score $f(h^*)$, differently distributed among $x$ and $y$
in $x.y$: 
$f(x) + h(y|x) = f(h^*)$. The number of $A^*$ extension steps (see
Algorithm~\ref{algorithm:a_star}) will be equal to the length of
$h^*$ making the search effort minimal. Notice that in this particular case a 
truncated stack at depth 1 suffices, suggesting that there is a correlation
between the search effort and the goodness of the estimate in the
$A^*$ ranking function.

In the second case we can set $h(y|x)=\infty$ for $y$ non-empty and, 
of course, $h(y|x)=0$ for empty y.
This will make $g_L(x) = f(x)$, if $x$ is complete and $g_L(x) =
\infty$, if $x$ is incomplete; any incomplete hypothesis will thus
have a higher score than any complete hypothesis, causing $A^*$ to
evaluate all the complete hypotheses in $L$ hence degenerating into an
exhaustive search; the search effort is maximal.

In practice the $h(y|x)$ function is chosen heuristically.

\subsection{$A^*$ for Lattice Decoding}

There are a few reasons that make $A^*$ appealing for our problem:
\begin{itemize}
\item the lattice can be conceptually structured as a prefix tree of
  hypotheses --- the time alignment is taken into consideration when
  comparing two word prefixes;
\item the algorithm operates with whole prefixes $x$, making
  it ideal for incorporating language models whose memory is the
  entire utterance prefix;
\item a reasonably good overestimate $h(y|x)$ and an efficient way to
  calculate $h_L(x)$ are readily available using the n-gram model, as
  we will explain later.
\end{itemize}

Before explaining our approach to lattice decoding using the $A^*$
algorithm, let us define a few terms. 

The lattices we work with retain the following information after the first pass:
\begin{itemize}
\item time-alignment of each node;
\item for each link connecting two nodes in the lattice we retain: 
  \begin{itemize}
  \item word identity $w(link)$;
  \item acoustic model score --- log-probability of acoustic segment
    covered by the link given the word, $logP_{AM}(A(link)|w,link)$;
    to make this possible, the ending nodes of the link must contain
    all contextual information necessary for assigning acoustic model
    scores; for example, in a crossword triphone system, all the words 
    labeling the links leaving the end node must have the same first phone;
  \item n-gram language model score --- log-probability of the word,
    $logP_{NG}(w|link)$; again, to make this possible, the start node
    of the link must contain the context $(n-1)$-gram --- it is a
    state in the finite state machine describing the n-gram language
    model used to generate the lattice; we thus refer to lattices as
    bigram or trigram lattices depending on the order of the language
    model that was used for generating it. The size of the lattice
    grows exponentially fast with the language model order.
  \end{itemize}
\end{itemize}
The lattice has a unique starting and ending node, respectively.

A \emph{link} in the lattice is an arc connecting two nodes of the
lattice. Two links are considered identical if and only if their word
identity is the same and their starting and ending nodes are the same,
respectively.

A \emph{path} $p$ through the lattice is an ordered set of links 
$l_0 \ldots l_n$ with the constraint that any two consecutive links cover
adjacent time intervals:
\begin{equation}
  \label{eq:dummy}
  p = \{l_0 \ldots l_n: \forall i=0 \ldots n-1, ending\_node(l_i) = starting\_node(l_{i+1})\}
\end{equation}
We will refer to the starting node of $l_0$ as  the starting node of
path $p$ and to the ending node of $l_n$ as the ending node of path $p$.

A \emph{partial path} is a path whose starting node is the same as the
starting node of the entire lattice and a \emph{complete path} is one
whose starting/ending nodes are the same as those of the entire lattice,
respectively.

With the above definitions, a lattice can be conceptually organized as a
prefix tree of paths. When rescoring the lattice using a different language model than the
one that was used in the first pass, we seek to find the complete path
$p=l_0 \ldots l_n$ maximizing:
\begin{equation}
  \label{eq:scoring_function}
  f(p) = \sum_{i=0}^{n} [logP_{AM}(l_i) + LMweight \cdot logP_{LM}(w(l_i)|w(l_0) \ldots
  w(l_{i-1})) - logP_{IP}]
\end{equation}
where:
\begin{itemize}
\item $logP_{AM}(l_i)$ is the acoustic model log-likelihood assigned to link $l_i$;
\item $logP_{LM}(w(l_i)|w(l_0)\ldots w(l_{i-1}))$ is the language model
log-probability assigned to link $l_i$ given the previous links on the partial path $l_0 \ldots l_i$; 
\item $LMweight>0$ is a constant weight which multiplies the language model score 
  of a link; its theoretical justification is unclear but experiments
  show its usefulness;
\item $logP_{IP}>0$ is the ``insertion penalty''; again, its theoretical
  justification is unclear but experiments show its usefulness. 
\end{itemize}

To be able to apply the $A^*$ algorithm we need to find an appropriate 
stack entry scoring function
$g_L(x)$ where $x$ is a partial path and $L$ is the set of complete paths
in the lattice. Going back to the definition~(\ref{eq:g_overestimate})
of $g_L(\cdot)$ we need an overestimate $g(x.y)=f(x) + h(y|x) \geq
f(x.y)$ for all possible $y=l_k \ldots l_n$ complete continuations of
$x$ allowed by the lattice. We propose to use the heuristic:
\begin{eqnarray}
  h(y|x) = \sum_{i=k}^{n} [logP_{AM}(l_i) + LMweight \cdot
  (logP_{NG}(l_i) + logP_{COMP}) - logP_{IP}] \nonumber\\
  + LMweight \cdot logP_{FINAL} \cdot \delta(k<n)  \label{eq:h_function}
\end{eqnarray}
A simple calculation shows that if $logP_{LM}(l_i)$ satisfies:
$$logP_{NG}(l_i) + logP_{COMP} \geq logP_{LM}(l_i), \forall l_i$$ 
then $g_L(x) = f(x) + max_{y \in C_L(x)} h(y|x)$ is a an appropriate
choice for the $A^*$ stack entry scoring function.

The justification for the $logP_{COMP}$ term is that it is supposed to
compensate for the per word difference in log-probability between the
n-gram model $NG$ and the superior model $LM$ with which we rescore the
lattice --- hence $logP_{COMP} > 0$. Its expected value can be
estimated from the difference in perplexity between the two
models $LM$ and $NG$. Theoretically we should use a higher value than
the maximum pointwise difference between the two models:
$$logP_{COMP} \geq \max_{\forall l_i} [logP_{LM}(l_i|l_0 \ldots l_{i-1})-logP_{NG}(l_i)]$$
but in practice we set it by trial and error starting with the
expected value as an initial guess. 

The $logP_{FINAL} > 0$ term is used for practical considerations as
explained in the next section.

The calculation of $g_L(x)$ (\ref{eq:g_overestimate}) is made very
efficient after realizing that one can use the dynamic programming technique in the Viterbi
algorithm~\cite{viterbi}. Indeed, for a given lattice $L$, the value of $h_L(x)$ is completely
determined by the identity of the ending node of $x$; a Viterbi backward pass
over the lattice can store at each node the corresponding value of
$h_L(x) = h_L(ending\_node(x))$ such that it is readily available in the
$A^*$ search.

\subsection{Some Practical Considerations }\label{section:practical_considerations}

In practice one cannot maintain a potentially infinite stack. We chose 
to control the stack depth using two thresholds:
one on the maximum number of entries in the stack, called
\verb+stack-depth-threshold+ and another one on the maximum log-probability
difference between the top most and the bottom most hypotheses in the
stack, called \verb+stack-logP-threshold+.

As glimpsed from the two limiting cases analyzed in
Section~(\ref{astar_theory}), there is a clear interaction between the
quality of the stack entry scoring function~(\ref{eq:g_overestimate})
and the number of hypotheses explored, which in practice has to be
controlled by the maximum stack size.

A gross overestimate used in connection with a finite stack may lure
the search to a cluster of paths which is suboptimal --- the desired
cluster of paths may fall out of the stack if the overestimate
happens to favor a wrong cluster. 

Also, longer prefixes --- thus having shorter suffixes ---
benefit less from the per word $logP_{COMP}$ compensation which means
that they may fall out of a stack already full with shorter hypotheses 
--- which have high scores due to compensation. 
This is the justification for the $logP_{FINAL}$ term in the compensation
function $h(y|x)$: the variance $var[logP_{LM}(l_i|l_0 \ldots
l_{i-1})-logP_{NG}(l_i)]$ is a finite positive quantity so
the compensation is likely to be closer to the expected value
$E[logP_{LM}(l_i|l_0 \ldots l_{i-1})-logP_{NG}(l_i)]$ for longer
$y$ continuations than for shorter ones; introducing a constant
$logP_{FINAL}$ term is equivalent to an adaptive $logP_{COMP}$ depending on
the length of the $y$ suffix --- smaller equivalent $logP_{COMP}$ for long
suffixes $y$ for which $E[logP_{LM}(l_i|l_0 \ldots l_{i-1})-logP_{NG}(l_i)]$
is a better estimate for $logP_{COMP}$ than it is for shorter ones.

Because the structured language model is computationally expensive, a strong limitation is 
being placed on the width of the search --- controlled by the\\
\verb+stack-depth-threshold+ and the \verb+stack-logP-threshold+. For an acceptable search
width --- runtime --- one seeks to tune the compensation parameters to maximize
performance measured in terms of WER. 
However, the correlation between these parameters and the WER is not
clear and makes the diagnosis of search problems extremely
difficult. Our method for choosing the search parameters was to sample
a few complete paths $p_1,\ldots,p_N$ from each lattice, rescore those
paths according to the $f(\cdot)$ function (\ref{eq:scoring_function})
and then rank the $h^*$ path output by the $A^*$  search among the
sampled paths. A correct $A^*$ search should result in average rank
0. 
In practice this doesn't happen but one can trace the topmost path
$p^*$ in the offending cases --- $p^* \neq h^*$ and $f(p^*) > f(h^*)$:
\begin{itemize}
\item if a prefix of the $p^*$ hypothesis is still present in the
stack when $A^*$ returns then the search failed strictly because of
insufficient compensation; 
\item if no prefix of $p^*$ is present in the stack then the incorrect
  search outcome was caused by an interaction between compensation and
  insufficient search width.
\end{itemize}

The method we chose for sampling paths from the lattice was an N-best
search using the n-gram language model scores; this is appropriate for 
pragmatic reasons --- one prefers lattice rescoring to N-best
list rescoring exactly because of the possibility to extract a path that is
not among the candidates proposed in the N-best list --- as well as
practical reasons --- they are among the ``better'' paths in terms of
WER.

\chapter{Speech Recognition Experiments} \label{chapter:lattice_decoding_experiments}

The set of experiments presented in Section~\ref{section:experiments}
showed improvement in perplexity over the 3-gram language model. The
experimental setup is however fairly restrictive and artificial when
compared to a real world speech recognition task:
\begin{itemize}
\item although the headword percolation and binarization procedure is
  automatic, the treebank used as training data was generated by human
  annotators;
\item albeit statistically significant, the amount of training data
  (approximatively 1 million words) is small compared to that used for
  developing language models used in real world speech recognition
  experiments;
\item the word level tokenization of treebank text is different than
  that used in the speech recognition community, the former being
  tuned to facilitate linguistic analysis.
\end{itemize}

In the remaining part of the chapter we will describe the experimental
setup used for speech recognition experiments involving the structured
language model, results and conclusions. The experiments were run on
three different corpora --- Switchboard (SWB), Wall Street Journal
(WSJ) and Broadcast News (BN) --- sampling different points of the
speech recognition spectrum --- conversational speech over telephone
lines at one end and read grammatical text recorded in ideal acoustic
conditions at the other end.

In order to evaluate our model's potential as part of a speech
recognizer, we had to address as follows the problems outlined above:
\begin{itemize}
\item \underline{manual vs.\ automatic parse trees} There are two
  corpora for which there exist treebanks, although of limited size:
  Wall Street Journal (WSJ) and Switchboard (SWB). The UPenn
  Treebank~\cite{Upenn} contains manually parsed WSJ text. There also
  exists a small part of Switchboard which was manually parsed at
  UPenn ---- approx.\ 20,000 words. This allows the training of an
  automatic parser --- we have used the Collins parser~\cite{mike96}
  for SWB and the Ratnaparkhi parser~\cite{ratnaparkhi:parser} for WSJ
  and BN --- which is going to be used to generate an \emph{automatic
    treebank}, possibly with a slightly different word-tokenization
  than that of the two manual treebanks. We evaluated the sensitivity
  of the structured language model to this aspect and showed that the
  reestimation procedure presented in
  Chapter~\ref{chapter:model_reest} is powerful enough to overcome any
  handicap arising from automatic treebanks.
\item \underline{more training data} The availability of an automatic
  parser to generate parse trees for the SLM training data --- used
  for initializing the SLM --- opens the possibility of training the
  model on much more data than that used in the experiments presented
  in Section~\ref{section:experiments}. The only limitations are of
  computational nature, imposed by the speed of the parser used to
  generate the automatic treebank and the efficiency and speed of the
  reestimation procedure for the structured language model parameters.
  As our experiments show, the reestimation procedure leads to a
  better structured model --- under both measures of perplexity and
  word error rate\footnote{Reestimation is also going to smooth out
    peculiarities in the automatically generated treebank}.  In
  practice the speed of the SLM is the limiting factor on the amount
  of training data. For Switchboard we have only 2 million words of
  language modeling training data so this is not an issue; for WSJ we
  were able to accommodate only 20 million words of training data,
  much less than the 40 million words used by standard language models
  on this task; for BN the discrepancy between the baseline 3-gram and
  the SLM is even bigger, we were able to accommodate only 14 million
  words of training data, much less than the 100 million words used by
  standard language models on this task.
\item \underline{different tokenization} We address this problem in
  the following section.
\end{itemize}

\section{Experimental Setup} \label{section:experimental_setup}

In order to train the structured language model (SLM) as described
in Chapter~\ref{chapter:model_reest} we use parse trees from which to initialize the
parameters of the model\footnote{The use of initial statistics
  gathered in a different way is an interesting direction of research; 
  the convergence properties of the reestimation procedure become 
  essential in such a situation} . Fortunately a part of the SWB/WSJ data has been
manually parsed at UPenn~\cite{Upenn},\cite{ws97}; let us refer to this
corpus as a Treebank. The training data used for speech
recognition --- CSR ---  is different from the Treebank in two
aspects:
\begin{itemize}
\item the Treebank is only a subset of the usual CSR training data;
\item the Treebank tokenization is different from that of the
  CSR corpus; among other spurious small differences, the most
  frequent ones are of the type presented in Table~\ref{tab:trbnk_csr_mismatch}.
\end{itemize}
\begin{table}[htbp]
  \begin{center}
    \begin{tabular}{|c|c|} \hline
      Treebank     & CSR \\ \hline \hline
      do n't       & don't   \\
      it 's        & it's\\
      jones '      & jones'\\
      i 'm         & i'm \\
      i 'll        & i'll \\
      i 'd         & i'd \\
      we 've       & we've\\
      you 're      & you're\\ \hline
    \end{tabular}
    \caption{Treebank --- CSR tokenization mismatch}
    \label{tab:trbnk_csr_mismatch}
  \end{center}
\end{table}
Our goal is to train the SLM on the CSR corpus. 

\subsubsection{Training Setup}\label{training_procedure}
The training of the SLM model proceeds as follows:
\begin{itemize}
%\item train SLM on SWB-Treebank --- using the SWB-Treebank closed vocabulary ---
%  as described in chapter~\ref{chapter:model_reest}; this is possible because for this
%  data we have parse trees from which we can gather initial statistics;
\item Process the CSR training data to bring it closer to the
  Treebank format. We applied the transformations suggested by
  Table~\ref{tab:trbnk_csr_mismatch}; the resulting corpus will be
  called CSR-Treebank, although at this stage we only have words
  and no parse trees for it;
\item Transfer the syntactic knowledge from the Treebank onto the CSR-Treebank training
  corpus; as a result of this stage, CSR-Treebank is truly a
  ``treebank'' containing binarized and headword annotated trees:
  \begin{itemize}
  \item for the SWB experiments we parsed the SWB-CSR-Treebank corpus using the SLM
    trained on the SWB-Treebank --- thus using the SLM as a parser; 
    the vocabulary for this step was the union between the SWB-Treebank
    and the SWB-CSR-Treebank closed vocabularies. The resulting trees
    are already binary and have headword annotation.
  \item for the WSJ and BN experiments we parsed the WSJ-CSR-Treebank corpus using 
    the Ratnaparkhi maximum entropy parser~\cite{ratnaparkhi:parser},
    trained on the UPenn Treebank data\footnote{The parser is
      mismatched, the most important difference being the fact that in 
      the training data of the parser numbers are written as
      ``\$123'' whereas in the data to be parsed they are expanded to %$
      ``one hundred twenty three dollars''; we rely on the SLM parameter reestimation
      procedure to smooth out this mismatch}. 
    The resulting trees were binarized and annotated with headwords
    using the procedure described in
    Section~\ref{section:headword_percolation}.
  \end{itemize}
\item Apply the SLM parameter reestimation procedure on the
  CSR-Treebank training corpus using the parse trees obtained at the
  previous step for gathering initial statistics.
\end{itemize}

Notice that we have avoided ``transferring'' the syntactic knowledge from the
Treebank tokenization directly onto the CSR tokenization; the reason
is that CSR word tokens like ``he's'' or ``you're'' cross boundaries
of syntactic constituents in the Treebank corpus and the transfer of
parse trees from the Treebank to the CSR corpus is far from obvious
and likely to violate syntactic knowledge present in the treebank.

\subsubsection{Lattice Decoding Setup}\label{lattice_decoding}

\begin{figure}[htbp]
  \begin{center}
    \epsfig{file=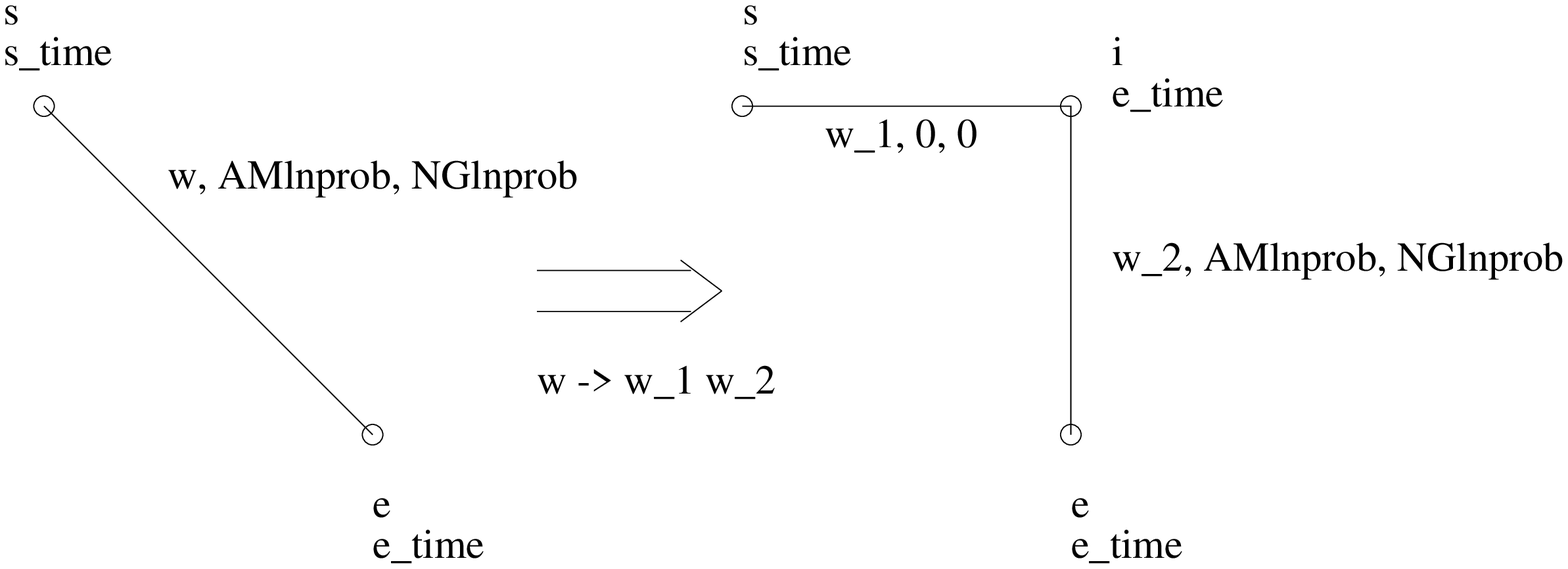, height=4cm,width=11cm}
    \caption{Lattice CSR to CSR-Treebank Processing}
    \label{fig:split_link}
  \end{center}
\end{figure}

To be able to run lattice decoding experiments we need to bring the
lattices --- in CSR tokenization --- to the CSR-Treebank
format. The only operation involved in this transformation is
splitting certain words into two parts, as suggested by
Table~\ref{tab:trbnk_csr_mismatch}. Each link whose word needs to be
split is cut into two parts and an intermediate node is inserted into 
the lattice as shown in figure~\ref{fig:split_link}. The acoustic
and language model scores of the initial link are copied onto the
second new link. 
{\em For all the decoding experiments we have carried out, the WER
  is measured after undoing the transformations highlighted above;
  the reference transcriptions for the test data were not touched and
  the NIST  SCLITE\footnote{SCLITE is a standard program supplied by
    NIST for scoring speech recognizers} package was used for measuring the WER}.

The refinement of the SLM presented in
Section~\ref{section:word_level_ppl},
Eq.~(\ref{eq:ppl1_l2r}---\ref{eq:ppl1_rho})
was not used at all during the following experiments due to its low
ratio of improvement versus computational cost.

\section{Perplexity Results}\label{ppl_results}
%perplexity;
As a first step we evaluated the perplexity performance of the 
SLM relative to that of a deleted interpolation 3-gram model
\emph{trained in the same conditions}. As outlined in the previous
section, we worked on the CSR-Treebank corpus.

\subsection{Wall Street Journal Perplexity Results}\label{WSJ_ppl_results}

We chose to work on the DARPA'93 evaluation HUB1 test setup. The size
of the test set is 213 utterances, 3446 words. The 20kwds open
vocabulary and baseline 3-gram model are the standard ones provided by
NIST and LDC.

As a first step we evaluated the perplexity performance of the SLM
relative to that of a deleted interpolation 3-gram model trained under
the same conditions: training data size 20Mwds (a subset of the
training data used for the baseline 3-gram model), standard HUB1 open
vocabulary of size 20kwds; both the training data and the vocabulary
were re-tokenized such that they conform to the Upenn Treebank
tokenization. We have linearly interpolated the SLM with the above
3-gram model:
$$P(\cdot)=\lambda \cdot P_{3gram}(\cdot) +(1-\lambda) \cdot
P_{SLM}(\cdot)$$
showing a 10\% relative reduction over the perplexity
of the 3-gram model. The results are presented in
Table~\ref{tab:wsj_ppl_results}. The SLM parameter reestimation
procedure\footnote{Due to the fact that the parameter reestimation
  procedure for the SLM is computationally expensive we ran only a single
  iteration} reduces the PPL by 5\% ( 2\% after interpolation with
the 3-gram model ).  The main reduction in PPL comes however from the
interpolation with the 3-gram model showing that although overlapping,
the two models successfully complement each other. The interpolation
weight was determined on a held-out set to be $\lambda = 0.4$. In this
experiment both language models operate in the UPenn Treebank text tokenization.
\begin{table}[htbp]
  \begin{center}
    \begin{tabular}{|l|c|c|c|} \hline
      \multicolumn{1}{|c|}{Language Model}& \multicolumn{3}{c|}{L2R Perplexity}\\\cline{2-4}
      & DEV set & \multicolumn{2}{c|}{TEST set}\\\cline{2-4}
      & \multicolumn{2}{c|}{no int} & 3-gram int\\ \hline\hline
      Trigram                         & 33.0 & 147.8 & 147.8\\ \hline \hline
      SLM; Initial stats(iteration 0) & 39.1 & 151.9 & 135.9 \\ \hline
      SLM; Reestimated(iteration 1)   & 34.6 & 144.1 & 132.8 \\ \hline
    \end{tabular}
    \caption{WSJ-CSR-Treebank perplexity results}
    \label{tab:wsj_ppl_results}
  \end{center}
\end{table}

\subsection{Switchboard Perplexity Results}\label{SWB_ppl_results}

For the Switchboard experiments the size of the training data was 2.29
Mwds; the size of the test data set aside for perplexity measurements
was 28 Kwds --- WS97 DevTest~\cite{ws97}.  We used a closed vocabulary
of size 22Kwds. Again, we have also linearly interpolated the SLM with
the deleted interpolation 3-gram baseline showing a modest reduction
in perplexity:
$$P(w_i|W_{i-1})=\lambda \cdot P_{3-gram}(w_i|w_{i-1},w_{i-2}) +
(1-\lambda) \cdot P_{SLM}(w_i|W_{i-1})$$
The interpolation weight was
determined on a held-out set to be $\lambda= 0.4$.  The results are
presented in Table~\ref{tab:swb_ppl_results}.
\begin{table}[htbp]
  \begin{center}
    \begin{tabular}{|l|c|c|c|} \hline
      \multicolumn{1}{|c|}{Language Model}& \multicolumn{3}{c|}{L2R Perplexity}\\\cline{2-4}
      & DEV set & \multicolumn{2}{c|}{TEST set}\\\cline{2-4}
      & \multicolumn{2}{c|}{no int} & 3-gram int\\ \hline\hline
      Trigram                        & 22.53 & 68.56 & 68.56\\ \hline \hline
      SLM; Seeded with Auto-Treebank & 23.94 & 72.09 & 65.80 \\ \hline
      SLM; Reestimated(iteration 4)  & 22.70 & 71.04 & 65.35 \\ \hline
    \end{tabular}
    \caption{SWB-CSR-Treebank perplexity results}
    \label{tab:swb_ppl_results}
  \end{center}
\end{table}

\subsection{Broadcast News Perplexity Results}\label{BN_ppl_results}

For the Broadcast News experiments the size of the training data was 14
Mwds; the size of the test data set aside for perplexity measurements
was 23150 wds --- DARPA'96 HUB4 dev-test.  We used an open vocabulary
of size 61Kwds. Again, we have also linearly interpolated the SLM with
the deleted interpolation 3-gram baseline built on exactly the same
training data showing an overall 7\% relative reduction in perplexity:
$$P(w_i|W_{i-1})=\lambda \cdot P_{3-gram}(w_i|w_{i-1},w_{i-2}) +
(1-\lambda) \cdot P_{SLM}(w_i|W_{i-1})$$
The interpolation weight was
determined on a held-out set to be $\lambda= 0.4$.  The results are
presented in Table~\ref{tab:bn_ppl_results}.
\begin{table}[htbp]
  \begin{center}
    \begin{tabular}{|l|c|c|c|} \hline
      \multicolumn{1}{|c|}{Language Model}& \multicolumn{3}{c|}{L2R Perplexity}\\\cline{2-4}
      & DEV set & \multicolumn{2}{c|}{TEST set}\\\cline{2-4}
      & \multicolumn{2}{c|}{no int} & 3-gram int\\ \hline\hline
      Trigram                        & 35.4 & 217.8 & 217.8\\ \hline \hline
      SLM; Seeded with Auto-Treebank & 57.7 & 231.6 & 205.5 \\ \hline
      SLM; Reestimated(iteration 2)  & 40.1 & 221.7 & 202.4 \\ \hline
    \end{tabular}
    \caption{SWB-CSR-Treebank perplexity results}
    \label{tab:bn_ppl_results}
  \end{center}
\end{table}

\section{Lattice Decoding Results}\label{wer_results}

We proceeded to evaluate the WER performance of the SLM using the
$A^*$ lattice decoder described in
Chapter~\ref{chapter:a_star_lattice_decoder}. 
Before describing the experiments we need to make clear one point;
there are two language model scores associated with each link in
the lattice: 
\begin{itemize}
\item the language model score assigned by the model that
  generated the lattice, referred to as the LAT3-gram; this model
  operates on text in the CSR tokenization; 
\item the language model score assigned by rescoring each
  link in the lattice with the deleted interpolation 3-gram built on
  the data in the CSR-Treebank tokenization, referred to as the
  TRBNK3-gram;
\end{itemize}

\subsection{Wall Street Journal Lattice Decoding Results}\label{WSJ_wer_results}

The lattices on which we ran rescoring experiments were obtained using
the standard 20k (open) vocabulary language model (LAT3-gram)
trained on more training data than the SLM --- about 40Mwds. The
deleted interpolation 3-gram model (TRBNK3-gram) built on much less
training data --- 20Mwds, same as SLM --- and using the same standard
open vocabulary --- after re-tokenizing it such that it matches the
UPenn Treebank text tokenization --- is weaker than the one used for generating the
lattices, as confirmed by our experiments. Consequently, we ran
lattice rescoring experiments in two setups:
\begin{itemize}
\item using the language model that generated the lattice ---
  LAT3-gram --- as the baseline model; language model scores are
  available in the lattice.
\item using the TRBNK3-gram language model --- same training
  conditions as the SLM; we had to assign new language model scores to
  each link in the lattice.
\end{itemize}

The 3-gram lattices we used have an ``oracle'' WER\footnote{The
  ``oracle'' WER is calculated by finding the path with the least
  number of errors in each lattice} of 3.4\%; the baseline
WER is 13.7\%, obtained using the standard 3-gram model provided by DARPA
(dubbed LAT3-gram) --- trained on 40Mwds and using a 20k open vocabulary. 

\subsubsection{Comparison between LAT3-gram and TRBNK3-gram}\label{wsj_3-gram_comparison}

A first batch of experiments evaluated the power of the two 3-gram
models at our disposal. The LAT3-gram scores are available in the
lattice from the first pass and we can rescore each link in the
lattice using the TRBNK3-gram model. The Viterbi algorithm can be used
to find the best path through the lattice according to the scoring
function~(\ref{eq:scoring_function}) where $logP_{LM}(\cdot)$ can be
either of the above or a linear combination of the two. Notice that
the linear interpolation of link language model scores:
$$
P(l) = \lambda \cdot P_{LAT3-gram}(l) + (1-\lambda) \cdot
P_{TRBNK3-gram}(l) $$
doesn't lead to a proper probabilistic model due
to the tokenization mismatch.  In order to correct this problem we
adjust the workings of the TRBNK3-gram to take two steps whenever a
split link is encountered and interpolate with the correct LAT3-gram
probability for the two links.  For example:
\begin{eqnarray}
P(don't|x,y) & = & \lambda \cdot P_{LAT3-gram}(don't|x,y) + \nonumber \\
& & (1-\lambda) \cdot P_{TRBNK3-gram}(do|x,y) \cdot
P_{TRBNK3-gram}(n't|y, do) \label{eqn:split_link}
\end{eqnarray}
The
results are shown in Table~\ref{tab:wsj_viterbi_results}.  The
parameters in~(\ref{eq:scoring_function}) were set to: 
\verb+LMweight = 16+, \verb+logP_{IP} = 0+, usual values for WSJ.
%%
%% table viterbi results
%%
\begin{table}[htbp]
  \begin{center}
    \begin{tabular}{|l|c|c|c|c|c|c|}\hline
      $\lambda$ &  0.0        &   0.2  &   0.4  &   0.6  &   0.8  & 1.0 \\ \hline \hline
      WER(\%)   & {\bf 14.7}  &  14.2  &  13.8  &  13.7  &  13.5  & {\bf 13.7} \\ \hline
    \end{tabular}
    \caption{3-gram Language Model; Viterbi Decoding Results}
    \label{tab:wsj_viterbi_results}
  \end{center}
\end{table}

\subsubsection{LAT3-gram driven search using the SLM}\label{wsj_astar_lat}

A second batch of experiments evaluated the performance of the SLM. 
The perplexity results show that interpolation with the 3-gram model
is beneficial for our model. The previous experiments show that the
LAT3-gram model is more powerful than the TRBNK3-gram model.
The interpolated language model score:
$$ P(l) = \lambda \cdot P_{LAT3-gram}(l) + (1-\lambda) \cdot P_{SLM}(l) $$
is calculated as explained in the previous section --- see Eq.~\ref{eqn:split_link}.

The results for different interpolation coefficient values are shown
in Table~\ref{tab:wsj_slm_int_results}.
The parameters controlling the SLM were the same as in
Chapter~\ref{chapter:model_reest}.  

As explained previously, due to the fact that the SLM's memory extends over the entire prefix
we need to apply the $A^*$ algorithm to find the overall best path in
the lattice. The parameters controlling the $A^*$ search were set to:
$logP_{COMP}$ = 0.5, $logP_{FINAL}$ = 0, $LMweight$ = 16, $logP_{IP}$ = 0,
\verb+stack-depth-threshold=30+, \verb+stack-depth-logP-threshold=100+
(see~\ref{eq:scoring_function} and ~\ref{eq:h_function}).

The $logP_{COMP}$, $logP_{FINAL}$ and \verb+stack-depth-threshold+,\\
\verb+stack-depth-logP-threshold+ were optimized directly on test data
for the best interpolation value found  in the perplexity
experiments. The $LMweight$, $logP_{IP}$ parameters are the ones typically
used with the 3-gram model for the WSJ task; we did not adjust them to
try to fit the SLM better.
%%
%% table SLM results
%%
\begin{table}[htbp]
  \begin{center}
    \begin{tabular}{|l|c|c|c|}\hline
      $\lambda$ &  0.0  &   0.4  &   1.0 \\ \hline \hline
      WER(\%) (iteration 0 SLM )   & 14.4 & \underline{13.0} & 13.7 \\ \hline
      WER(\%) (iteration 1 SLM )   & 14.3 & 13.2 & 13.7 \\ \hline
    \end{tabular}
    \caption{LAT-3gram $+$ Structured Language Model; $A^*$ Decoding Results}
    \label{tab:wsj_slm_int_results}
  \end{center}
\end{table}

The structured language model achieved an absolute improvement in WER of 0.7\%
(5\% relative) over the baseline.

\subsubsection{TRBNK3-gram driven search using the SLM}\label{wsj_astar_trbnk}

We rescored each link in the lattice using the TRBNK3-gram language
model and used this as a baseline for further experiments. As showed
in Table~\ref{tab:wsj_viterbi_results}, the baseline WER becomes
14.7\%. The relevance of the experiments using the TRBNK3-gram
rescored lattices is somewhat questionable since the
lattice was generated using a much stronger language model ---
the LAT3-gram. Our point of view is the following: assume that we have a
set of hypotheses which were produced in some way; we then rescore
them using two language models, M1 and M2; if model M2 is truly
superior to M1\footnote{From a speech recognition perspective}, then
the WER obtained by rescoring the set of hypotheses using model M2
should be lower than that obtained using model M1.

We repeated the experiment in which we linearly interpolate 
the SLM with the 3-gram language model:
$$ P(l) = \lambda \cdot P_{TRBNK3-gram}(l) + (1-\lambda) \cdot P_{SLM}(l)$$ 
for different interpolation coefficients. The $A^*$ search parameters were the same as before.
The results are presented in Table~\ref{tab:wsj_slm_int_trbnk_results}. 
%%
%% table TRBNK-3gram + SLM results
%%
\begin{table}[htbp]
  \begin{center}
    \begin{tabular}{|l|c|c|c|}\hline
      $\lambda$ &  0.0  &   0.4  &   1.0 \\ \hline
      WER(\%) (iteration 0 SLM )   & 14.6 & 14.3 & 14.7 \\ \hline
      WER(\%) (iteration 3 SLM )   & 13.8 & 14.3 & 14.7 \\ \hline
    \end{tabular}
    \caption{TRBNK-3gram $+$ Structured Language Model; $A^*$ Decoding Results}
    \label{tab:wsj_slm_int_trbnk_results}
  \end{center}
\end{table}
The structured language model interpolated with the trigram model
achieves 0.9\% absolute (6\% relative) reduction over the trigram baseline; the
parameters controlling the $A^*$ search have not been tuned for this
set of experiments. 

\subsection{Switchboard Lattice Decoding Results}\label{SWB_wer_results}

On the Switchboard corpus, the lattices for which we ran decoding
experiments were obtained using a language model (LAT3-gram) trained in very
similar conditions --- roughly same training data size and vocabulary, 
closed over test data --- to the ones under which the SLM and the
baseline deleted interpolation 3-gram model (TRBNK3-gram) were
trained. The only difference is the tokenization --- 
CSR vs.\ CSR-Treebank, see Section~\ref{training_procedure} ---
which makes the LAT3-gram act as phrase based language model when compared 
to TRBNK3-gram. The experiments confirmed that LAT3-gram is stronger
than TRBNK-3gram.

Again, we ran lattice rescoring experiments in two setups:
\begin{itemize}
\item using the language model that generated the lattice --- LAT3-gram 
  --- as the baseline model; language model scores are available in
  the lattice.
\item using the TRBNK3-gram language model --- same training
  conditions as the SLM; we had to assign new language model scores to
  each link in the lattice.
\end{itemize}

\subsubsection{Comparison between LAT3-gram and TRBNK3-gram}\label{swb_3-gram_comparison}

The results are shown in Table~\ref{tab:viterbi_results}, for
different interpolation values:
$$
P(l) = \lambda \cdot P_{LAT3-gram}(l) + (1-\lambda) \cdot P_{TRBNK3-gram}(l) $$
The parameters in~(\ref{eq:scoring_function}) were set to:
\verb+LMweight = 12+, \verb+logP_{IP} = 10+.
%%
%% table viterbi results
%%
\begin{table}[htbp]
  \begin{center}
    \begin{tabular}{|l|c|c|c|c|c|c|}\hline
      $\lambda$ &  0.0        &   0.2  &   0.4  &   0.6  &   0.8  & 1.0 \\ \hline \hline
      WER(\%)   & {\bf 42.3}  &  41.8  &  41.2  &  41.0  &  41.0  & {\bf 41.2} \\ \hline
    \end{tabular}
    \caption{3-gram Language Model; Viterbi Decoding Results}
    \label{tab:viterbi_results}
  \end{center}
\end{table}

\subsubsection{LAT3-gram driven search using the SLM}\label{swb_astar_lat}

The previous experiments show that the LAT3-gram model is more
powerful than the TRBNK3-gram model.  We thus wish to interpolate the
SLM with the LAT3-gram model:
$$
P(l) = \lambda \cdot P_{LAT3-gram}(l) + (1-\lambda) \cdot
P_{SLM}(l) $$
We correct the interpolation the same way as described
in the WSJ experiments --- see Section~\ref{wsj_3-gram_comparison},
Eq.~\ref{eqn:split_link}.

The parameters controlling the SLM were the same as in
chapter~\ref{chapter:model_reest}.  The parameters controlling the
$A^*$ search were set to: $logP_{COMP}$ = 0.5, $logP_{FINAL}$ = 0,
$LMweight$ = 12, $logP_{IP}$ = 10, \verb+stack-depth-threshold=40+,
\verb+stack-depth-logP-threshold=100+ (see~\ref{eq:scoring_function}
and ~\ref{eq:h_function}). The $logP_{COMP}$, $logP_{FINAL}$ and
\verb+stack-depth-threshold+,\\ \verb+stack-depth-logP-threshold+ were
optimized directly on test data for the best interpolation value found
in the perplexity experiments. In all other experiments they were kept
fixed to these values. The $LMweight$, $logP_{IP}$ parameters are the
ones typically used with the 3-gram model for the Switchboard task; we
did not adjust them to try to fit the SLM better.

The results for different interpolation coefficient values are shown
in Table~\ref{tab:slm_int_results}.
%%
%% table SLM results
%%
\begin{table}[htbp]
  \begin{center}
    \begin{tabular}{|l|c|c|c|}\hline
      $\lambda$                 &  0.0        &   0.4  &   1.0 \\ \hline \hline
      WER(\%) (SLM iteration 0) &  41.8       &  40.7  &  41.2 \\ \hline
      WER(\%) (SLM iteration 3) &  41.6       &  \underline{40.5}  &  41.2 \\ \hline
    \end{tabular}
    \caption{LAT-3gram $+$ Structured Language Model; $A^*$ Decoding Results}
    \label{tab:slm_int_results}
  \end{center}
\end{table}

The structured language model achieved an absolute improvement of
0.7\% WER over the baseline; the improvement is statistically
significant at the 0.001 level according to a sign test at the
sentence level.

For tuning the search parameters we have applied the N-best lattice
sampling technique described in
section~\ref{section:practical_considerations}. As a by-product, the
WER performance of the structured language model on N-best list
rescoring --- N = 25 --- was 40.4\%. The average rank of the
hypothesis found by the $A^*$ search among the N-best ones --- after
rescoring them using the structured language model interpolated with
the trigram --- was 0.3. There were 329 offending sentences --- out of
a total of 2427 sentences --- in which the $A^*$ search lead to a
hypothesis whose score was lower than that of the top hypothesis among
the N-best(0-best). In 296 cases the prefix of the rescored 0-best was
still in the stack when $A^*$ returned --- inadequate compensation ---
and in the other 33 cases, the 0-best hypothesis was lost during the
search due to the finite stack size.

\subsubsection{TRBNK3-gram driven search using the SLM}\label{swb_astar_trbnk}

We rescored each link in the lattice using the TRBNK3-gram language
model and used this as a baseline for further experiments. As showed
in Table~\ref{tab:viterbi_results}, the baseline WER is 42.3\%.

We then repeated the experiment in which we linearly interpolate the
SLM with the 3-gram language model:
$$
P(l) = \lambda \cdot P_{TRBNK3-gram}(l) + (1-\lambda) \cdot
P_{SLM}(l)$$
for different interpolation coefficients.  The parameters
controlling the $A^*$ search were set to: $logP_{COMP}$ = 0.5,
$logP_{FINAL}$ = 0, $LMweight$ = 12, $logP_{IP}$ = 10,
\verb+stack-depth-threshold=40+, \verb+stack-depth-logP-threshold=100+
(see~\ref{eq:scoring_function} and ~\ref{eq:h_function}). The results
are presented in Table~\ref{tab:slm_int_trbnk_results}.
%%
%% table TRBNK-3gram + SLM results
%%
\begin{table}[htbp]
  \begin{center}
    \begin{tabular}{|l|c|c|c|}\hline
      $\lambda$ &  0.0  &   0.4  &   1.0 \\ \hline
      WER(\%) (iteration 0 SLM ) & 42.0  &  41.6  &  42.3 \\ \hline
      WER(\%) (iteration 3 SLM ) & 42.0  &  41.6  &  42.3 \\ \hline
    \end{tabular}
    \caption{TRBNK-3gram $+$ Structured Language Model; $A^*$ Decoding Results}
    \label{tab:slm_int_trbnk_results}
  \end{center}
\end{table}
The structured language model interpolated with the trigram model
achieves 0.7\% absolute reduction over the trigram baseline.

\subsection{Broadcast News Lattice Decoding Results}\label{SWB_wer_results}

The Broadcast News (BN) lattices for which we ran
decoding experiments were obtained using a language model (LAT3-gram)
trained on much more training data than the SLM; a typical figure for
BN is 100Mwds. We could accommodate 14Mwds of training data for the SLM
and the baseline deleted interpolation 3-gram model (TRBNK3-gram).
The experiments confirmed that LAT3-gram is stronger than TRBNK-3gram.

The set set on which we ran the experiments was the DARPA'96 HUB4
dev-test. We used an open vocabulary of 61kwds.  Again, we ran lattice
rescoring experiments in two setups:
\begin{itemize}
\item using the language model that generated the lattice ---
  LAT3-gram --- as the baseline model; language model scores are
  available in the lattice.
\item using the TRBNK3-gram language model --- same training
  conditions as the SLM; we had to assign new language model scores to
  each link in the lattice.
\end{itemize}

The test set is segmented in different focus conditions summarized in
Table~\ref{tab:bn_foci}.
%%
%% table bn foci
%%
\begin{table}[htbp]
  \begin{center}
    \begin{tabular}{|l|l|}\hline
      Focus & Description \\\hline
      F0    & baseline broadcast speech (clean, planned)\\\hline
      F1    & spontaneous broadcast speech (clean)\\\hline
      F2    & low fidelity speech (typically narrowband)\\\hline
      F3    & speech in the presence of background music\\\hline
      F4    & speech under degraded acoustical conditions\\\hline
      F5    & non-native speakers (clean, planned)\\\hline
      FX    & all other speech (e.g. spontanous non-native)\\\hline
    \end{tabular} 
    \caption{Broadcast News Focus conditions}
    \label{tab:bn_foci}
  \end{center}
\end{table}

\subsubsection{Comparison between LAT3-gram and TRBNK3-gram}\label{swb_3-gram_comparison}

The results are shown in Table~\ref{tab:bn_viterbi_results}, for
different interpolation values:
$$
P(l) = \lambda \cdot P_{LAT3-gram}(l) + (1-\lambda) \cdot P_{TRBNK3-gram}(l) $$

The parameters in~(\ref{eq:scoring_function}) were set to:
\verb+LMweight = 13+, \verb+logP_{IP} = 10+.
%%
%% table viterbi results
%%
\begin{table}[htbp]
  \begin{center}
    \begin{tabular}{|l|c|c|c|c|c|c|}\hline
      $\lambda$ &  0.0        &   0.2  &   0.4  &   0.6  &   0.8  & 1.0 \\ \hline \hline
      WER(\%)   & {\bf 35.2}  &  34.0  &  33.2  &  33.0  &  32.9  & {\bf 33.1} \\ \hline
    \end{tabular} 
    \caption{3-gram Language Model; Viterbi Decoding Results}
    \label{tab:bn_viterbi_results}
  \end{center}
\end{table}

\subsubsection{LAT3-gram driven search using the SLM}\label{swb_astar_lat}

The previous experiments show that the LAT3-gram model is more
powerful than the TRBNK3-gram model.  We thus wish to interpolate the
SLM with the LAT3-gram model:
$$
P(l) = \lambda \cdot P_{LAT3-gram}(l) + (1-\lambda) \cdot
P_{SLM}(l) $$
We correct the interpolation the same way as described
in the WSJ experiments --- see Section~\ref{wsj_3-gram_comparison},
Eq.~\ref{eqn:split_link}.

The parameters controlling the SLM were the same as in
chapter~\ref{chapter:model_reest}.  The parameters controlling the
$A^*$ search were set to: $logP_{COMP}$ = 0.5, $logP_{FINAL}$ = 0,
$LMweight$ = 13, $logP_{IP}$ = 10, \verb+stack-depth-threshold=25+,
\verb+stack-depth-logP-threshold=100+ (see~\ref{eq:scoring_function}
and ~\ref{eq:h_function}).

The results for different interpolation coefficient values are shown
in Table~\ref{tab:bn_slm_int_results}.
%%
%% table SLM results
%%
\begin{table}[htbp]
  \begin{center}
    \begin{tabular}{|l|c|c|c|}\hline
      $\lambda$                      &   0.0 &   0.4  &   1.0 \\ \hline \hline
      WER(\%) (SLM iteration 0)      &  34.4 &  33.0  &  33.1 \\ \hline
      WER(\%) (SLM iteration 2)      &  35.1 &  33.0  &  33.1\\ \hline
    \end{tabular}
    \caption{LAT-3gram $+$ Structured Language Model; $A^*$ Decoding Results}
    \label{tab:bn_slm_int_results}
  \end{center}
\end{table}
The breakdown on different focus conditions is shown in
Table~\ref{tab:bn_slm_int_results_foci}.
\begin{table}[htbp]
  \begin{center}
    \begin{tabular}{|lcr|c|c|c|c|c|c|c|c|}\hline
      $\lambda$ & Decoder & SLM iteration 
      &  F0  &  F1  &  F2  &  F3  &  F4  &  F5  &  FX  & overall\\ \hline
%---------------------------------------------------------------------------------------------------
      1.0       & Viterbi & 
      & {\bf 13.0} & 30.8 & 42.1 & 31.0 & 22.8 & 52.3 & 53.9 & 33.1\\ \hline\hline
%---------------------------------------------------------------------------------------------------
      0.0       & $A^*$   & 0
      & 13.3 & 31.7 & 44.5 & 32.0 & 25.1 & 54.4 & 54.8 & 34.4\\ \hline

      0.4       & $A^*$   & 0
      & {\bf 12.5} & 30.5 & 42.2 & 31.0 & 23.0 & 52.9 & 53.9 & 33.0\\ \hline

      1.0       & $A^*$   & 0
      & 12.9 & 30.7 & 42.1 & 31.0 & 22.8 & 52.3 & 53.9 & 33.1\\ \hline
%---------------------------------------------------------------------------------------------------
      0.0       & $A^*$   & 2
      & 14.8 & 31.7 & 46.3 & 31.6 & 27.5 & 54.3 & 54.8 & 35.1\\ \hline

      0.4       & $A^*$   & 2
      & {\bf 12.2} & 30.7 & 42.0 & 31.1 & 22.5 & 53.1 & 54.4 & 33.0\\ \hline

      1.0       & $A^*$   & 2
      & 12.9 & 30.7 & 42.1 & 31.0 & 22.8 & 52.3 & 53.9 & 33.1\\ \hline
%---------------------------------------------------------------------------------------------------
    \end{tabular}
    \caption{LAT-3gram $+$ Structured Language Model; $A^*$ Decoding
      Results; breakdown on different focus conditions}
    \label{tab:bn_slm_int_results_foci}
  \end{center}
\end{table}
The SLM achieves 0.8\% absolute (6\% relative) reduction in WER on the
F0 focus condition despite the fact that the overall WER reduction is
negligible. We also note the beneficial effect training has on the SLM
performance on the F0 focus condition.

\subsubsection{TRBNK3-gram driven search using the SLM}\label{swb_astar_trbnk}

We rescored each link in the lattice using the TRBNK3-gram language
model and used this as a baseline for further experiments. As showed
in Table~\ref{tab:bn_viterbi_results}, the baseline WER is 35.2\%.

We then repeated the experiment in which we linearly interpolate the
SLM with the 3-gram language model:
$$
P(l) = \lambda \cdot P_{TRBNK3-gram}(l) + (1-\lambda) \cdot
P_{SLM}(l)$$
for different interpolation coefficients.  The parameters
controlling the $A^*$ search were set to: $logP_{COMP}$ = 0.5,
$logP_{FINAL}$ = 0, $LMweight$ = 13, $logP_{IP}$ = 10,
\verb+stack-depth-threshold=25+, \verb+stack-depth-logP-threshold=100+
(see~\ref{eq:scoring_function} and ~\ref{eq:h_function}). The results
are presented in Table~\ref{tab:bn_int_trbnk_results}.
%%
%% table TRBNK-3gram + SLM results
%%
\begin{table}[htbp]
  \begin{center}
    \begin{tabular}{|l|c|c|c|}\hline
      $\lambda$                      &   0.0 &   0.4  &   1.0 \\ \hline \hline
      WER(\%) (SLM iteration 0)      &  35.4 &  34.9  &  35.2 \\ \hline
      WER(\%) (SLM iteration 2)      &  35.0 &  34.7  &  35.2\\ \hline
    \end{tabular}
    \caption{TRBNK-3gram $+$ Structured Language Model; $A^*$ Decoding Results}
    \label{tab:bn_int_trbnk_results}
  \end{center}
\end{table}
The breakdown on different focus conditions is shown in
Table~\ref{tab:bn_int_trbnk_results_foci}.
\begin{table}[htbp]
  \begin{center}
    \begin{tabular}{|lcr|c|c|c|c|c|c|c|c|}\hline
      $\lambda$ & Decoder & SLM iteration 
      &  F0  &  F1  &  F2  &  F3  &  F4  &  F5  &  FX  & overall\\ \hline
%---------------------------------------------------------------------------------------------------
      1.0       & Viterbi & 
      & {\bf 14.5} & 32.5 & 44.9 & 33.3 & 25.7 & 54.9 & 56.1 & 35.2\\ \hline\hline
%---------------------------------------------------------------------------------------------------
      0.0       & $A^*$   & 0
      & 14.6 & 32.9 & 44.6 & 33.1 & 26.3 & 54.4 & 56.9 & 35.4\\ \hline

      0.4       & $A^*$   & 0
      & {\bf 14.1} & 32.2 & 44.4 & 33.0 & 25.0 & 54.2 & 56.1 & 34.9\\ \hline

      1.0       & $A^*$   & 0
      & 14.5 & 32.4 & 44.9 & 33.3 & 25.7 & 54.9 & 56.1 & 35.2\\ \hline
%---------------------------------------------------------------------------------------------------
      0.0       & $A^*$   & 2
      & 13.7 & 32.4 & 44.7 & 32.9 & 26.1 & 54.3 & 56.3 & 35.0\\ \hline

      0.4       & $A^*$   & 2
      & {\bf 13.4} & 32.2 & 44.1 & 31.9 & 25.3 & 54.2 & 56.2 & 34.7\\ \hline

      1.0       & $A^*$   & 2
      & 14.5 & 32.4 & 44.9 & 33.3 & 25.7 & 54.9 & 56.1 & 35.2\\ \hline
%---------------------------------------------------------------------------------------------------
    \end{tabular}
    \caption{TRBNK-3gram $+$ Structured Language Model; $A^*$ Decoding
      Results; breakdown on different focus conditions}
    \label{tab:bn_int_trbnk_results_foci}
  \end{center}
\end{table}
The SLM achieves 1.1\% absolute (8\% relative) reduction in WER on the
F0 focus condition and an overall WER reduction of 0.5\% absolute. We
also note the beneficial effect training has on the SLM performance.

\subsubsection{Conclusions to Lattice Decoding Experiments}

We note that the parameter reestimation doesn't improve the WER
performance of the model in all cases. The SLM achieves an improvement over the 3-gram
baseline on all three corpora: Wall Street Journal, Switchboard and
Broadcast News.

\subsection{Taking Advantage of Lattice Structure}

As we shall see, in order to carry out experiments in which we try to
take further advantage of the lattice, we need to have proper language model scores on each
lattice link. For all the experiments in this section we used the TRBNK3-gram rescored lattices.

\subsubsection{Peeking Interpolation}

As described in Section~\ref{section:word_level_ppl}, the probability assignment for the
word at position $k+1$ in the input sentence is made using:
\begin{eqnarray}
P(w_{k+1}/W_{k}) & = & \sum_{T_{k}\in
  S_{k}}P(w_{k+1}/W_{k}T_{k})\cdot\rho(W_{k},T_{k})\\
  \mathrm{where} & & \nonumber \\ 
\rho(W_{k},T_{k}) & = & P(W_{k}T_{k})/\sum_{T_{k} \in S_{k}}P(W_{k}T_{k})
\end{eqnarray}
which ensures a proper probability over strings $W^*$, where $S_{k}$ is
the set of all parses present in the SLM stacks at the current stage $k$.

One way to take advantage of the lattice is to determine the set of
parses $S_{k}$ over which we are going to interpolate by knowing what the 
possible future words are --- the links leaving the end node of a
given path in the lattice bear only a small set of words --- for our
lattices, less than 10 on the average. The idea is that by knowing the
future word it is much easier to determine the most favorable parse
for predicting it. 
Let $\mathcal{W}_L(p)$ denote {\em the set of words that label the
  links leaving the end node of path $p$ in lattice $L$}. 
We can then restrict the set of parses $S_{k}$ used for interpolation to:
$$S_k^{pruned} = \{T_k^i: T_k^i = \arg \max_{T_k\in S_{k}}
P(w^i/W_{k}T_{k})\cdot\rho(W_{k},T_{k}),\ \forall\ w^i \in
\mathcal{W}_L(p) \}$$
We obviously have $S_k^{pruned}\subseteq S_{k}$.
Notice that this does not lead to a correct probability assignment
anymore since it violates the causality implied by the left-to-right
operation of the language model. In the extreme case of
$|\mathcal{W}_L(p)| = 1$ we have a model which, at each next word
prediction step, picks from among the parses in $S_k$ only the most
favorable one for predicting the next word. This leads to the
undesirable effect that at a subsequent prediction during the same
sentence the parse picked may change, always trying to make the best
possible current prediction. In order to compensate for this unwanted
effect we decided to run a second experiment in which only the parses
in $S_k^{pruned}$ are kept in the stacks of the structured language model
at position $k$ in the input sentence --- the other ones are discarded
and thus unavailable for later predictions in the sentence. This
speeds up considerably the decoder --- approximately 4 times faster
than the previous experiment --- and slightly improves on the results
in the previous experiment but still does not increase the
performance over the standard structured language model, as shown in
%Tables~\ref{tab:wsj_peek_slm_int_trbnk_results}---\ref{tab:peek_slm_int_trbnk_results}.
Table~\ref{tab:peek_slm_int_trbnk_results}. The results for the
standard SLM do not match those in
Table~\ref{tab:slm_int_trbnk_results} due to the fact that in this
case we have not applied the tokenization correction specified in 
Eq.~(\ref{eqn:split_link}), Section~\ref{wsj_3-gram_comparison}.
%% Switchboard
%% table TRBNK-3gram + peeking SLM results
%%
\begin{table}[htbp]
  \begin{center}
    \begin{tabular}{|l|c|c|c|c|c|c|}\hline
      $\lambda$ &  0.0  &   0.2  &   0.4  &   0.6  &   0.8  &   1.0 \\ \hline\hline
      WER(\%) (standard SLM)  & 42.0  &  41.8  &  41.9  &  41.5  &  42.1  &  42.5 \\ \hline\hline
      WER(\%) (peeking SLM)  & 42.3  &        &        &  42.0  &        &       \\ \hline
      WER(\%) (pruned peeking SLM)  & 42.1  &        &        &  41.9  &        &       \\ \hline
    \end{tabular}
    \caption{Switchboard;TRBNK-3gram $+$ Peeking SLM;}
    \label{tab:peek_slm_int_trbnk_results}
  \end{center}
\end{table}

\subsubsection{Normalized Peeking}

Another proper probability assignment for the next word $w_{k+1}$ could
be made according to:
\begin{eqnarray}
P(w_{k+1}/W_{k}) & = & norm(\alpha(w, W_k)),\label{eq:ppl_n}\\
\mathrm{where}\nonumber \\
\alpha(w, W_k)   & \doteq & \max_{T_k\in S_{k}}
P(w/W_{k}T_{k})\cdot\rho(W_{k},T_{k})\\
\mathrm{and} \nonumber\\
norm(\alpha(w, W_k)) & \doteq & \alpha(w_{k+1}, W_k) / \sum_{w \in \mathcal{V}} \alpha(w, W_k)
\end{eqnarray}

The sum over all words in the vocabulary $\mathcal{V}$ --- $|\mathcal{V}| \approx
20,000 $ --- prohibits the use of the above equation in perplexity
evaluations for computational reasons. In the lattice however we have a much smaller list of
future words so the summation needs to be carried only over
$\mathcal{W}_L(p)$ (see previous section) for a given path $p$. 
To take care of the fact that due to the truncation of $\mathcal{V}$ to
$\mathcal{W}_L(p)$ the probability assignment now violates the
left-to-right operation of the language model we can 
{\em redistribute the 3-gram mass} assigned to $\mathcal{W}_L(p)$
according to the formula proposed in Eq.~(\ref{eq:ppl_n}):
\begin{eqnarray}
  P_{SLMnorm}(w_{k+1}/W_{k}(p)) & = & norm(\alpha(w, W_k)) \cdot P_{TRBNK3-gram}(\mathcal{W}_L(p)) \label{eq:ppl_l} \\
  \alpha(w, W_k)   & \doteq & \max_{T_k\in S_{k}}
  P(w/W_{k}T_{k})\cdot\rho(W_{k},T_{k})\\
  norm(\alpha(w, W_k)) & \doteq & \alpha(w_{k+1}, W_k) / \sum_{w \in  \mathcal{W}_L(p)} \alpha(w, W_k)\\
  P_{TRBNK3-gram}(\mathcal{W}_L(p)) & \doteq & \sum_{w \in \mathcal{W}_L(p)} P_{TRBNK3-gram}(w/W_{k}(p))
\end{eqnarray}

Notice that if we let $\mathcal{W}_L(p) = \mathcal{V}$ we get back
Eq.~(\ref{eq:ppl_n}). 
\begin{table}[h]
  \begin{center}
    \begin{tabular}{|l|c|c|c|c|c|c|}\hline
      $\lambda$ &  0.0  &   0.2  &   0.4  &   0.6  &   0.8  &   1.0 \\ \hline\hline
      WER(\%) (standard SLM)  & 42.0  &  41.8  &  41.9  &  41.5  &  42.1  &  42.5 \\ \hline\hline
      WER(\%) (normalized SLM)  & 42.7  &        &  42.1  &  42.0  &  42.1  &   \\ \hline
      WER(\%)(pruned normalized SLM)  &        &        &        &  42.2  &        &  \\ \hline
    \end{tabular}
    \caption{Switchboard; TRBNK-3gram $+$ Normalized Peeking SLM;}
    \label{tab:peek_normalize_slm_int_trbnk_results}
  \end{center}
\end{table}
Again, one could discard from the SLM stacks the parses which do not
belong to $S_{k}^{pruned}$, as explained in the previous section.
Table~\ref{tab:peek_normalize_slm_int_trbnk_results} 
presents the results
obtained when linearly interpolating the above models with the 3-gram model:
$$ P(l/W_{k}(p)) = \lambda \cdot P_{TRBNK3-gram}(l/W_{k}(p)) +
(1-\lambda) \cdot P_{SLMnorm}(l/W_{k}(p))$$

The results for the
standard SLM do not match those in
Table~\ref{tab:slm_int_trbnk_results} due to the fact that in this
case we have not applied the tokenization correction specified in 
Eq.~(\ref{eqn:split_link}), Section~\ref{wsj_3-gram_comparison}.
Although some of the experiments showed improvement over the
WER baseline achieved by the 3-gram language model, none of
them performed better than the standard structured language model
linearly interpolated with the trigram model. 

\chapter{Conclusions and Future Directions}
\label{chapter:conclusions}

\section{Comments on Using the SLM as a Parser}

The structured language model could be used as a parser, namely select
the most likely parse according to our pruning strategy: $T^* =
argmax_{T}P(W,T)$. Due to the fact that the SLM allows parses in which
the words in a sentence are not joined under a single root node ---
see the definition of a complete parse and Figure~\ref{fig:c_parse}
--- a direct evaluation of the parse quality against the UPenn
Treebank parses is unfair. However, a
simple modification will constrain the parses generated by the SLM to
join all words in a sentence under a single root node.
 
Imposing the additional constraint that:
 \begin{itemize}
 \item $P(w_k$=\verb+</s>+$|W_{k-1}T_{k-1}) = 0\ if\ h_{-1}.tag \neq
   SB$ ensures that the end of sentence symbol \verb+</s>+ is
   generated only from a parse in which all the words have been joined
   in a single constituent.
 \end{itemize}
 
 One important observation is that in this case one has to eliminate
 the second pruning step in the model and the hard pruning in the
 cache-ing of the CONSTRUCTOR model actions; it is sufficient if this
 is done only when operating on the last stack vector before
 predicting the end of sentence \verb+</s>+. Otherwise, the parses that
 have all the words joined under a single root node may not be present
 in stacks before the prediction of the \verb+</s>+ symbol, resulting
 in a failure to parse a given sentence.

\section{Comparison with other Approaches} \label{section:comparison}

\subsection{Underlying $P(W,T)$ Probability Model}

The actions taken by the model are very similar to a LR parser.
However the encoding of the word sequence along with a parse tree
$(W,T)$ is different, proceeding bottom-up and interleaving the word
predictions. This leads to a different probability assignment than
that in a PCFG grammar --- which is based on a different encoding of
$(W,T)$.

A thorough comparison between the two classes of probabilistic languages ---
PCFGs and shift-reduce probabilistic push-down automata, to which the
SLM pertains --- has been presented in~\cite{abney}.

\begin{figure}[ht]
  \epsfig{file=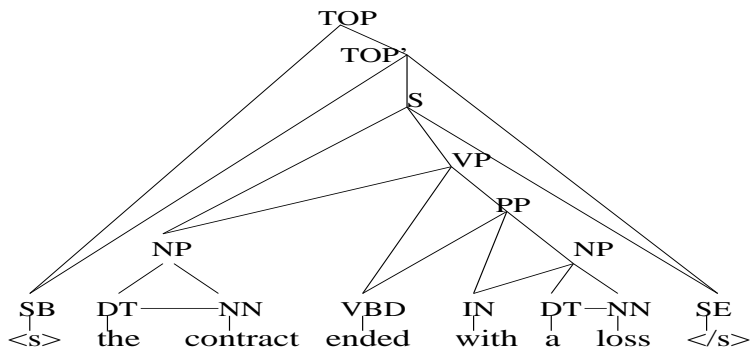,height=4.5cm,width=10cm}
  \caption{CFG dependencies} \label{fig:cfg_parse_w_dependencies}
\end{figure}

\begin{figure}[ht]
  \epsfig{file=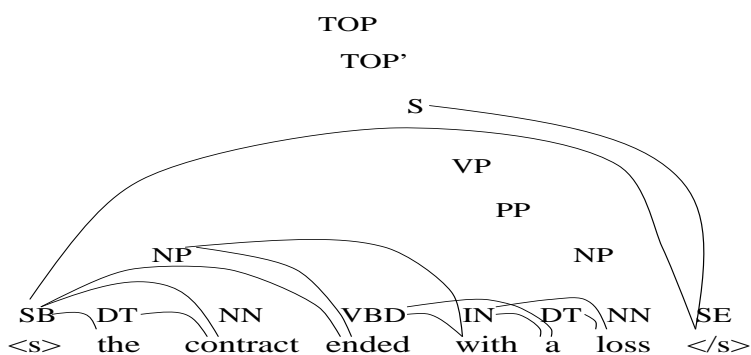,height=4.5cm,width=10cm}
  \caption{Tag reduced WORD-PREDICTOR dependencies} \label{fig:parse_w_dependencies_w}
\end{figure}

\begin{figure}[ht]
  \epsfig{file=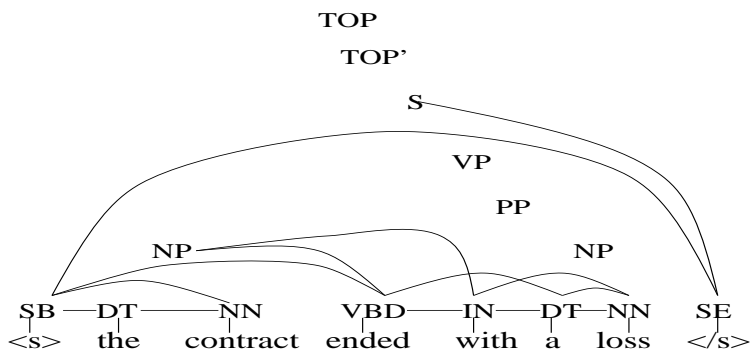,height=4.5cm,width=10cm}
  \caption{TAGGER dependencies} \label{fig:parse_w_dependencies_t}
\end{figure}

\begin{figure}[ht]
  \epsfig{file=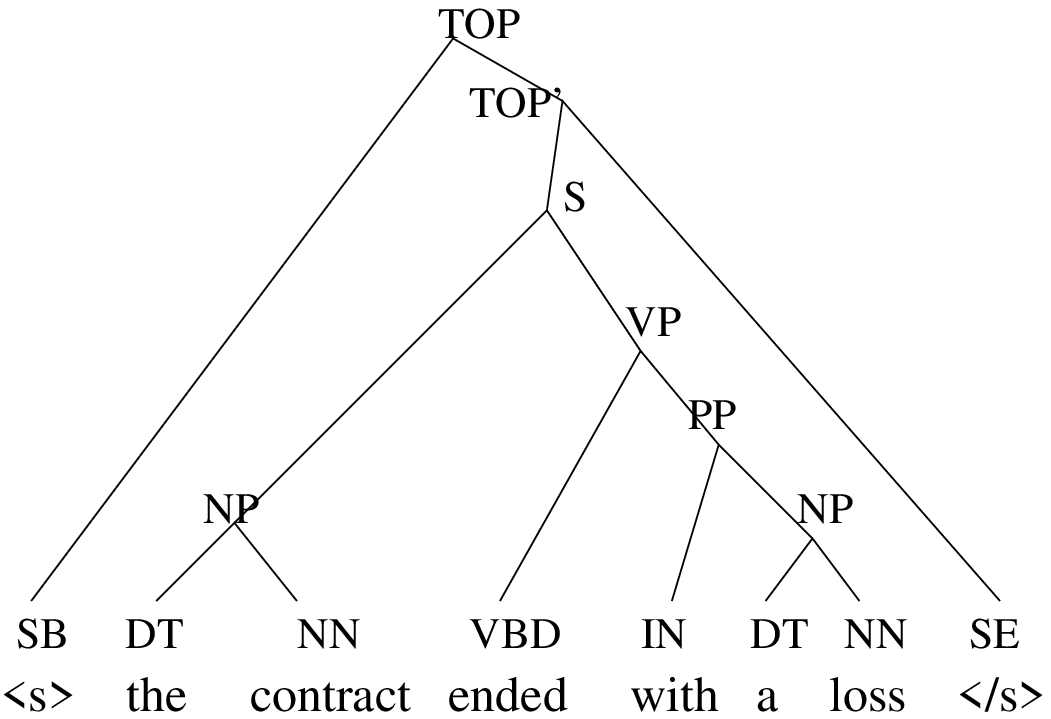,height=4.5cm,width=10cm}
  \caption{Tag reduced CONSTRUCTOR dependencies} \label{fig:parse_w_dependencies_p}
\end{figure}

Regarding $(W,T)$ as a graph,
Figure~\ref{fig:cfg_parse_w_dependencies} shows the dependencies in a
regular CFG; in contrast,
Figures~(\ref{fig:parse_w_dependencies_w}--\ref{fig:parse_w_dependencies_p})
show the probabilistic dependencies for each model component in the
SLM; a complete dependency structure is obtained by super-imposing the
three figures. To make the SLM directly comparable with a CFG we
discard the lexical information at intermediate nodes in the tree ---
headword annotation --- thus assuming the following equivalence
classifications in the model components --- see
Eq.(\ref{eq:word_predictor_prob}--\ref{eq:parser_prob}):
\begin{eqnarray}
  P(w_k|W_{k-1} T_{k-1}) =& P(w_k|[W_{k-1} T_{k-1}]) & = P(w_k|h_0.tag, h_{-1}.tag)\\
  P(t_k|w_k,W_{k-1} T_{k-1}) =& P(t_k|w_k,[W_{k-1} T_{k-1}]) & = P(t_k|w_k, h_0.tag, h_{-1}.tag)\\
  P(p_i^k|W_{k}T_{k}) =& P(p_i^k|[W_{k}T_{k}]) & = P(p_i^k|h_0.tag, h_{-1}.tag)
\end{eqnarray}

It can be seen that the probabilistic dependency structure is more
complex than that in a CFG even in this simplified SLM.

Along the same lines, the approach in~\cite{miller} regards the word
sequence $W$ with the parse structure $T$ as a Markov graph $(W,T)$
modeled using the CFG dependencies superimposed on the regular
word-level 2-gram dependencies, showing improvement in perplexity over
both 2-gram and 3-gram modeling techniques.

\subsection{Language Model}

A structured approach to language modeling has been taken
in~\cite{printz94}: the underlying probability model $P(W,T)$ is a
simple lexical link grammar, which is \emph{automatically} induced and
reestimated using EM from a training corpus containing word sequences
(sentences). The model doesn't make use of POS/NT labels --- which we
found extremely useful for word prediction and parsing.  Another
constraint is placed on the context used by the word predictor: the
two words in the context used for word prediction are always adjacent;
our models' hierarchical scheme allows the exposed headwords to
originate at any two different positions in the word prefix. Both
approaches share the desirable property that the 3-gram model belongs
to the parameter space of the model.

The language model we present is closely related to the one
investigated in~\cite{ws96}\footnote{The SLM might not have happened
  at all, weren't it for the work and creative environment in the WS96
  Dependency Modeling Group and the authors' desire to write a PhD
  thesis on structured language modeling}, however different in a few
important aspects:
\begin{itemize}
\item our model operates in a left-to-right manner, thus allowing its
  use directly in the hypothesis search for $\hat{W}$ in
  (\ref{intro:bayes});
\item our model is a factored version of the one in~\cite{ws96}, thus
  enabling the calculation of the joint probability of words and parse
  structure; this was not possible in the previous case due to the
  huge computational complexity of that model;
\item our model assigns probability at the word level, being a proper
  language model.
\end{itemize}

The SLM shares many features with both class based language models
\cite{brown92} and skip n-gram language models \cite{rosenfeld94}; an
interesting approach combining class based language models and
different order skip-bigram models is presented in \cite{saul97}. It
seems worthwhile to make two comments relating the SLM to these
approaches:
\begin{itemize}
\item the smoothing involving NT/POS tags in the WORD-PREDICTOR is
  similar to a class based language model using NT/POS labels for
  classes. We depart however from the usual approach by \emph{not
    making the conditional independence assumption}
  $P(w_{k+1}|w_k, \textrm{class}(w_k)) =
  P(w_{k+1}|\textrm{class}(w_k))$. Also, in our model the ``class''
  assignment --- through the heads exposed by a given parse $T_k$ for
  the word prefix $W_k$ and its ``weight'' $\rho(W_{k},T_{k})$,
  see~Eq.~(\ref{eq:ppl1}) --- is highly context-sensitive --- it
  depends on the entire word-prefix $W_k$ --- and is syntactically
  motivated through the operations of the CONSTRUCTOR. A comparison
  between the hh and HH equivalence classifications in the
  WORD-PREDICTOR --- see Table~\ref{table:_w} --- shows the usefulness
  of POS/NT labels for word prediction.
\item recalling the depth factorization of the model in
  Eq.~(\ref{eq:ppl_depth}), our model can be viewed as a skip n-gram
  where the probability of a skip $P(d_0,d_1|W_k)$ --- $d_0, d_1$ are
  the depths at which the two most recent exposed headwords $h_0, h_1$
  can be found, similar to $P(d|W_k)$ --- is highly context sensitive.
  Notice that the hierarchical scheme for organizing the word prefix
  allows for contexts that do not necessarily consist of adjacent
  words, as in regular skip n-gram models.
\end{itemize}

\section{Future Directions}

We have presented an original approach to language modeling
that makes use of syntactic structure.  The experiments we have
carried out show improvement in both perplexity and word error rate
over current state-of-the-art techniques.  Preliminary experiments
reported in~\cite{wu99} show complementarity between the SLM and a
topic language model yielding almost additive results --- word error
rate improvement --- on the Switchboard task. Among the directions
which we consider worth exploring in the future, are:
\begin{itemize}
\item automatic induction of the SLM initial parameter values;
\item better integration of the 3-gram model and the SLM;
\item better parameterization of the model components;
\item study interaction between SLM and other language modeling
  techniques such as cache and trigger or topic language models.
\end{itemize}

%%% Local Variables: 
%%% mode: latex
%%% TeX-master: "submission"
%%% TeX-master: "submission"
%%% TeX-master: "main"
%%% TeX-master: "main"
%%% End: 

% appendices, if any;
\appendix
\chapter{Minimizing KL Distance is Equivalent to Maximum Likelihood} \label{app:min_D_max_L}

Let $f_{\mathcal{T}}(Y)$ be the relative frequency
probability distribution induced on $\mathcal{Y}$ by the collection of 
training samples $\mathcal{T}$; this determines the set of desired
distributions $P_{\mathcal{T}} \doteq \{p(X,Y): p(Y) =
f_{\mathcal{T}}(Y)\}$.  Let $Q(\Theta) \doteq \{q_{\theta}(X,Y):\theta
\in \Theta \}$ be the model space. 

\begin{prop}
  Finding the maximum likelihood estimate $g \in Q(\Theta)$ is
  equivalent to finding the pair $(p, q) \in P_{\mathcal{T}} \times
  Q(\Theta)$ which minimizes the KL-distance $D(p \parallel q)$.
\end{prop}

For a given pair $(p, q) \in P_{\mathcal{T}} \times Q(\Theta)$ we
have:
\begin{eqnarray*}
  D(p \parallel q) & = & \sumxy p(x,y) \log \frac{p(x,y)}{q(x,y)}\\
  & = & \sumxy f(y) \cdot r(x|y) \log \frac{f(y) \cdot r(x|y)}{q(y) \cdot q(x|y)}\\
  & = & \sumy f(y) \log f(y) - \likelihood{q} + \sumy f(y)
  \cdot D(r(x|y) \parallel q(x|y))\\
  & \geq & \sumy f(y) \log f(y) - \max_{q \in Q(\Theta)} \likelihood{q} + 0
\end{eqnarray*}

The minimum value of $D(p \parallel q)$ is independent of $p$ and $q$
and is achieved \emph{if and only if} both:
\begin{eqnarray*}
  q(x,y) & = & arg \max_{g_\theta \in Q(\Theta)} \likelihood{g_\theta}\\
  r(x|y) & = & q(x|y)
\end{eqnarray*}
are satisfied. The second condition is equivalent to $p$ being the
\emph{I-projection} of a given $q$ onto $P_\mathcal{T}$:
\begin{eqnarray*}
  p & = & arg \min_{t \in P_\mathcal{T}} D(t \parallel q)\\
    & = & arg \min_{r(x|y)} D(f(y) \cdot r(x|y) \parallel q)
\end{eqnarray*}

So knowing the pair $(p,q) \in P_{\mathcal{T}} \times Q(\Theta)$ that
minimizes $D(p \parallel q)$ implies that the maximum likelihood distribution
$q \in Q(\Theta)$ has been found and reciprocally, once the  maximum likelihood distribution
$q \in Q(\Theta)$ is given we can find the $p$ distribution in
$P_{\mathcal{T}}$ that will minimize $D(p \parallel q), p \in
P_{\mathcal{T}}, q \in Q(\Theta)$. {\flushright $\Box$}

\chapter{Expectation Maximization as Alternating Minimization} \label{app:EM_eq_minD}

Let $f_{\mathcal{T}}(Y)$ be the relative frequency
probability distribution induced on $\mathcal{Y}$ by the collection of 
training samples $\mathcal{T}$; this determines the set of desired
distributions $P_{\mathcal{T}} \doteq \{p(X,Y): p(Y) =
f_{\mathcal{T}}(Y)\}$.  Let $Q(\Theta) \doteq \{q_{\theta}(X,Y):\theta
\in \Theta \}$ be the model space. 

\begin{prop}
  One alternating minimization step between $P_{\mathcal{T}}$ and
  $Q(\Theta)$ is equivalent to an EM update step:
  \begin{eqnarray}
    EM_{\mathcal{T},\theta_i}(\theta) & \doteq & \sum_{y \in \mathcal{Y}}
    f_{\mathcal{T}}(y) E_{q_{\theta_i}(X/Y)}[\log(q_{\theta}(X,Y)|y)], 
    \theta \in \Theta \\
    \theta_{i+1} & = & arg \max_{\theta \in \Theta}
    EM_{\mathcal{T},\theta_i}(\theta) \label{eq:EM_update_app}
  \end{eqnarray}
\end{prop}

\begin{figure}[htbp]
  \begin{center}
    \epsfig{file=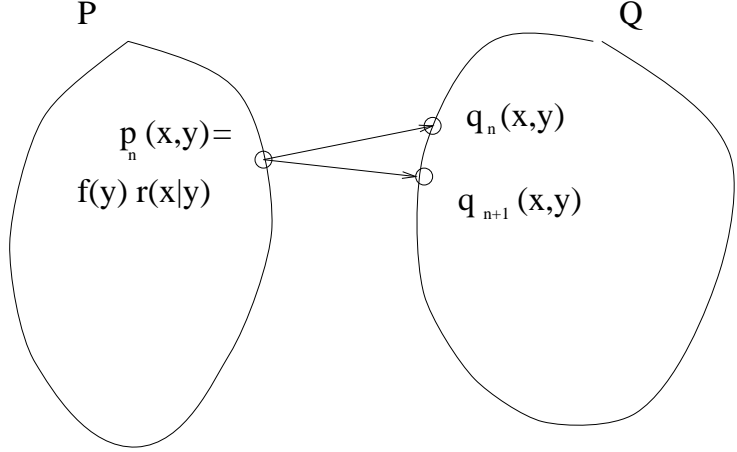, height=6cm}
    \caption{Alternating minimization between
      $P_{\mathcal{T}}$ and $Q(\Theta)$}
    \label{fig:AM_D}
  \end{center}
\end{figure}

One alternating minimization step starts from a given distribution
$q_n \in Q(\Theta)$, finds the I-projection $p_n$ of $q_n$ onto
$P_{\mathcal{T}}$; fixing $p_n$ we then find the I-projection
$q_{n+1}$ of $p_n$ onto $Q(\Theta)$. We will show that this leads to
the EM update equations~\ref{eq:EM_update_app}.

Given $q_n \in Q(\Theta), \forall p \in P_{\mathcal{T}}$, we have:
\begin{eqnarray*}
  D(p \parallel q_n) & = & \sumxy p(x,y) \log \frac{p(x,y)}{q_n(x,y)}\\
  & = & \sumxy f(y) \cdot r(x|y) \log \frac{f(y) \cdot r(x|y)}{q_n(x,y)}\\
  & = & \sumy f(y) \log \frac{f(y)}{q_n(y)} + \sumy f(y) (\sumx r(x/y) 
  \log \frac{r(x/y)}{q_n(x/y)})\\
  & = & \underbrace{\sumy f(y) \log \frac{f(y)}{q_n(y)}}_{independent\
    of\ r(x|y)} + \sumy f(y) \cdot \underbrace{D(r(x/y),q_n(x/y))}_{\geq 0}\\
\end{eqnarray*}
which implies that:
\begin{eqnarray*}
  \min_{p \in P_{\mathcal{T}}} D(p \parallel q_n) & = & \sumy f(y) \log \frac{f(y)}{q_n(y)}
\end{eqnarray*}
is achieved by $p_n = f(y) \cdot q_n(x|y)$. 

Now fixing $p_n$ we seek the $q \in Q(\Theta)$ which minimizes $D(p_n
\parallel q)$:
\begin{eqnarray*}
  D(p_n \parallel q) & = & \sumxy p_n(x,y) \log \frac{p_n(x,y)}{q(x,y)}\\
  & = & \sumxy f(y) \cdot q_n(x|y) \log \frac{f(y) \cdot q_n(x|y)}{q(x,y)}\\
  & = & \underbrace{\sumy f(y) \log \frac{f(y)}{q_n(y)} + 
    \sumy f(y) \cdot [\sumx q_n(x|y) \log q_n(x|y)]}_{independent\
    of\ q(x,y)} \\
  & & - \sumxy f(y) q_n(x|y) \log q(x,y)
\end{eqnarray*}
But the last term can rewritten as:
\begin{eqnarray*}
  \sumxy f(y) q_n(x|y) \log q(x,y) & = & \sumy f(y) \sumx q_n(x|y) \log q(x,y)\\
  & = & \underbrace{\sumy f(y) E_{q_n(X|Y)}[\log q(x,y)|y]}_{EM_{\mathcal{T},\theta_i}(\theta)}
\end{eqnarray*}

Thus finding $$\min_{q \in Q(\Theta)} D(p_n \parallel q)$$ 
is equivalent to finding 
$$
  \max_{q \in Q(\Theta)} EM_{\mathcal{T},\theta_i}(\theta)
$$
which is exactly the EM-update step~(\ref{eq:EM_update_app}).
{\flushright $\Box$}

\chapter{N-best EM convergence} \label{app:Nbest_convergence}

In the ``N-best'' training paradigm we use only a subset of the
conditional hidden event space $\mathcal{X}|y$, for any given seen
$y$. Associated with the model space $Q(\Theta)$ we now have a family
of strategies to sample from $\mathcal{X}|y$ a set of ``N-best''
hidden events $x$, for any $y \in \mathcal{Y}$. Each sampling strategy
is a function that associates a set of hidden sequences to a given
observed sequence: $s:\mathcal{Y}\rightarrow 2^{\mathcal{X}}$. The
family is parameterized by $\theta \in \Theta$:
\begin{eqnarray}
  & & \mathcal{S}({\Theta}) \doteq \{ s_{\theta}:\mathcal{Y}
  \rightarrow 2^{\mathcal{X}}, \forall \theta \in \Theta \}\label{eq:N-best-samplingapp}
\end{eqnarray}
Each $\theta$ value identifies a particular sampling function.

Let:
\begin{eqnarray}
  q_{\theta}^s(X,Y) & \doteq & q_{\theta}(X,Y) \cdot 1_{s_{\theta}(Y)}(X)\label{eq:N-best-model-space1app}\\
  q_{\theta}^s(X|Y) & \doteq & \frac{q_{\theta}(X,Y)}{\sum_{X \in
      \mathcal{s}_{\theta}(Y)} q_{\theta}(X,Y)} \cdot 1_{s_{\theta}(Y)}(X)\label{eq:N-best-model-space2app}\\
  Q(\mathcal{S},\Theta) & \doteq & \{q_{\theta}^s(X,Y):\theta \in \Theta \}
\end{eqnarray}

\begin{prop}
  \label{prop:AM_Nbest_convergenceapp}
  Assuming that $\forall \theta \in \Theta, \supp{q_{\theta}} =
  \mathcal{X} \times \mathcal{Y}$ (``smooth'' $q_{\theta}(x,y)$)
  holds, one alternating minimization step between $P_{\mathcal{T}}$
  and $Q(\mathcal{S},\Theta)$ ---$\theta_{i} \rightarrow \theta_{i+1}$
  --- is equivalent to:
  \begin{eqnarray}
    \theta_{i+1} & = & \arg \max_{\theta \in \Theta} \sum_{y \in \mathcal{Y}}
    f_{\mathcal{T}}(y)
    E_{q_{\theta_i}^s(X|Y)}[\log(q_{\theta}(X,Y)|y)] \label{eq:EM_update_Nbestapp}
  \end{eqnarray}
  if $\theta_{i+1}$ satisfies:
  \begin{eqnarray}
    s_{\theta_{i}}(y) \subseteq  s_{\theta_{i+1}}(y), \forall y \in \mathcal{T}
  \end{eqnarray}
  Only $\theta \in \Theta\ s.t.\ s_{\theta_{i}}(y) \subseteq
  s_{\theta}(y), \forall y \in \mathcal{T}$ are candidates in the
  M-step.
\end{prop}

Proof:

\emph{E-step}:\\
Given $q_{\theta_i}^s(x,y) \in Q(\mathcal{S},\Theta)$, find 
$p_n(x,y) = f(y) \cdot r_n(x|y) \in P(\mathcal{T})\ s.t.\ \kldist{f(y) \cdot
  r_n(x|y)}{q_{\theta_i}^s(x,y)}$ is minimized.  As shown in
appendix~\ref{app:EM_eq_minD}:
\begin{eqnarray}
  r_n(x|y)=q_{\theta_i}^s(x|y),\ \forall y \in \mathcal(T) \label{eq:rn}
\end{eqnarray}
Notice that for smooth $q_{\theta_i}(x|y)$ we have:
\begin{eqnarray}
  \supp{r_n(x|y)} = \supp{q_{\theta_i}^s(x|y)} = s_{\theta_{i}}(y),\
  \forall y \in \mathcal{T} \label{eq:support_identity}
\end{eqnarray}

\emph{M-step}:\\
given $p_n(x,y)=f(y) \cdot q_{\theta_i}^s(x|y),\ find\ \theta_{i+1}
\in \Theta\ s.t.\ \kldist{p_n}{q_{\theta_{i+1}}^s}\ is\ minimized$.

\begin{lemma}
  \label{lemma1}
  For the M-step we only need to consider candidates $\theta \in
  \Theta$ for which we have
  \begin{eqnarray}
    \label{eq:lemma1}
    s_{\theta_{i}}(y) \subseteq s_{\theta}(y), \forall y \in \mathcal{T}
  \end{eqnarray}
\end{lemma}

Indeed, assuming that $\exists\ (x_0,y_0)\ s.t.\ y_0 \in \mathcal{T}\
\mathrm{and}\ x_0 \in s_{\theta_{i}}(y)\ \mathrm{but}\ x_0 \notin s_{\theta}(y)$, we
have: $(x_0,y_0) \in \supp{f(y) \cdot r_n(x|y)}$
(see~(\ref{eq:support_identity})) and $(x_0,y_0) \notin
\supp{q_{\theta}^s(x,y)}$ (see~(\ref{eq:N-best-model-space1app}))
which means that $f(y_0) \cdot r_n(x_0|y_0) > 0$ and
$q_{\theta}^s(x_0,y_0) =0$, rendering\\ $\kldist{f(y) \cdot
  r_n(x|y)}{q_{\theta}^s(x,y)} = \infty$.  {\flushright $\Box$}

Following the proof in appendix~\ref{app:EM_eq_minD}, it is easy to
show that:
\begin{eqnarray}
  \label{eq:theta_star}
  {\theta}^* = \arg \max_{\theta \in \Theta} \sum_{y \in \mathcal{Y}}
  f_{\mathcal{T}}(y) E_{q_{\theta_i}^s(X|Y)}[\log(q_{\theta}^s(X,Y)|y)]
\end{eqnarray}
minimizes $\kldist{p_n}{q_{\theta}^s}, \forall \theta \in \Theta$.

Using the result in Lemma~\ref{lemma1}, only $\theta \in \Theta$
satisfying~(\ref{eq:lemma1}) are candidates for the M-step, so:
\begin{eqnarray}
  \label{eq:theta_star2}
  {\theta}^* = \arg \max_{{\theta \in {\Theta} | s_{\theta_{i}}(y) \subseteq s_{\theta}(y), \forall y \in \mathcal{T}}} \sum_{y \in \mathcal{Y}}
  f_{\mathcal{T}}(y) E_{q_{\theta_i}^s(X|Y)}[\log(q_{\theta}(X,Y)
  \cdot 1_{s_{\theta}(Y)}(X)|y)]
\end{eqnarray}

But notice that $\supp{q_{\theta_i}^s(x|y)} = s_{\theta_{i}}(y),\ 
\forall y \in \mathcal{T}$ (see~(\ref{eq:support_identity})) and these
are the only $x$ values contributing to the conditional expectation on
a given $y$ ; for these however we have $1_{s_{\theta}(y)}(x) = 1$
because of~(\ref{eq:lemma1}). This implies that~(\ref{eq:theta_star2})
can be rewritten as:
\begin{eqnarray}
  \label{eq:theta_star3}
  {\theta}^* = \arg \max_{\theta \in {\Theta} | s_{\theta_{i}}(y) \subseteq s_{\theta}(y), \forall y \in \mathcal{T}} \sum_{y \in \mathcal{Y}}
  f_{\mathcal{T}}(y) E_{q_{\theta_i}^s(X|Y)}[\log(q_{\theta}(X,Y)|y)]
\end{eqnarray}

Because the set over which the maximization is carried over depends on
${\theta}_i$ the M-step is not simple. However we notice that if the
maximum on the entire space $\Theta$:
\begin{eqnarray}
  \label{eq:theta_next}
  {\theta}_{i+1} = \arg \max_{\theta \in \Theta} \sum_{y \in \mathcal{Y}}
  f_{\mathcal{T}}(y) E_{q_{\theta_i}^s(X|Y)}[\log(q_{\theta}(X,Y)|y)]
\end{eqnarray}
satisfies: $s_{\theta_{i}}(y) \subseteq s_{\theta_{i+1}}(y), \forall y
\in \mathcal{T}$, then $\theta_{i+1}$ is the correct update
$\theta^*$. {\flushright $\Box$}

\chapter{Structured Language Model Parameter Reestimation} \label{app:slm_reest}

The probability of a $(W,T)$ sequence is obtained by chaining the
probabilities of the elementary events in its derivation, as described in
section~\ref{section:prob_model}:
\begin{eqnarray}
  P(W,T) = P(d(W,T)) = \prod_{i=1}^{length(d(W,T))} p(e_i)
\end{eqnarray}

The E-step is carried by sampling the space of hidden events for a
given seen sequence $W$ according to the pruning strategy outlined in
section~\ref{section:pruning}:
\begin{eqnarray*}
  P_{\theta}^s(W,T) & \doteq & P_{\theta}(W,T) \cdot 1_{s_{\theta}(W)}(T)\\
  P_{\theta}^s(T|W) & \doteq & \frac{P_{\theta}(T,W)}{\sum_{T \in
      \mathcal{s}_{\theta}(W)} P_{\theta}(W,T)} \cdot 1_{s_{\theta}(W)}(T)\\
\end{eqnarray*}

The logarithm of the probability of a given derivation can be
calculated as follows:
\begin{eqnarray*}
  \lefteqn{\log P_{\theta}(W,T)}\\ & = & \sum_{i=1}^{length(d(W,T))} \log P_{\theta}(e_i)\\
  & = & \sum_{m} \sum_{(u^{(m)}, {\underline{z}}^{(m)})}
  \sum_{i=1}^{length(d(W,T))} \log P_{\theta}(u^{(m)},
  {\underline{z}}^{(m)}) \cdot \delta(e_i, (u^{(m)},
  {\underline{z}}^{(m)}))\\
  & = & \sum_{m} \sum_{(u^{(m)}, {\underline{z}}^{(m)})}
  [ \sum_{i=1}^{length(d(W,T))} \delta(e_i, (u^{(m)},
  {\underline{z}}^{(m)})) ] \cdot \log
  P_{\theta}(u^{(m)},{\underline{z}}^{(m)})\\
  & = & \sum_{m} \sum_{(u^{(m)}, {\underline{z}}^{(m)})}
  \#[(u^{(m)}, {\underline{z}}^{(m)}) \in d(W,T)] \cdot \log
  P_{\theta}(u^{(m)},{\underline{z}}^{(m)})
\end{eqnarray*}
where the random variable 
$$\#[(u^{(m)}, {\underline{z}}^{(m)}) \in d(W,T)]$$
denotes the number of occurrences of the 
$(u^{(m)},{\underline{z}}^{(m)})$ event in the derivation of $W,T$. 

Let  
\begin{eqnarray*}
  E_{ P_{{\theta}_i}^s(T|W)}[\#[(u^{(m)}, {\underline{z}}^{(m)}) \in
  d(W,T)]] & \doteq & a_{{\theta}_i}((u^{(m)}, {\underline{z}}^{(m)}),
  W) \\
  \sum_{W \in \mathcal{T}} f(W) \cdot a_{{\theta}_i}((u^{(m)},
  {\underline{z}}^{(m)}), W) & \doteq &  a_{{\theta}_i}(u^{(m)}, {\underline{z}}^{(m)})
\end{eqnarray*}
We then have:
\begin{eqnarray*} 
  \lefteqn{E_{ P_{{\theta}_i}^s(T|W)}[\log P_{\theta}(W,T)]}\\ & = & \sum_{m} \sum_{(u^{(m)}, {\underline{z}}^{(m)})}
  a_{{\theta}_i}((u^{(m)}, {\underline{z}}^{(m)}), W) \cdot \log
  P_{\theta}(u^{(m)},{\underline{z}}^{(m)})
\end{eqnarray*}
and 
\begin{eqnarray}
  \lefteqn{\sum_{W \in \mathcal{T}} f(W) \cdot E_{ P_{{\theta}_i}^s(T|W)}[\log
  P_{\theta}(W,T)]} \label{eq:aux_func.app} \\ & = & \sum_{m} \sum_{(u^{(m)}, {\underline{z}}^{(m)})}
  a_{{\theta}_i}(u^{(m)}, {\underline{z}}^{(m)}) \cdot \log
  P_{\theta}(u^{(m)},{\underline{z}}^{(m)}) 
\end{eqnarray}

The E-step thus consists of the calculation of the expected values
$a_{{\theta}_i}((u^{(m)}, {\underline{z}}^{(m)}))$, for every model
component and every event $(u^{(m)}, {\underline{z}}^{(m)})$ in the 
derivations that survived the pruning process. 

In the M-step we need to find a new parameter value $\theta_{i+1}$
such that me maximize the EM auxiliary
function~(\ref{eq:aux_func.app}):
\begin{eqnarray}
  \theta_{i+1}  & = & 
  \arg \max_{\theta \in \mathcal{\theta}}\sum_{W \in \mathcal{T}} f(W)
  \cdot E_{ P_{{\theta}_i}^s(T|W)}[\log P_{\theta}(W,T)]\\
  & = & \arg \max_{\theta \in \mathcal{\theta}} \sum_{m} \sum_{(u^{(m)}, {\underline{z}}^{(m)})}
  a_{{\theta}_i}((u^{(m)}, {\underline{z}}^{(m)})) \cdot \log
  P_{\theta}(u^{(m)},{\underline{z}}^{(m)})
\end{eqnarray}

The parameters $\theta$ are the maximal order joint counts
$C^{(m)}(u^{(m)}, {\underline{z}}^{(m)})$ for each model component 
$m \in \{$WORD-PREDICTOR, TAGGER, PARSER $\}$.

One can easily notice that the M-step is in fact a problem of maximum
likelihood estimation for each model component $m$ from joint counts 
$a_{{\theta}_i}((u^{(m)},{\underline{z}}^{(m)}))$. 
Taking into account the parameterization of $P_{\theta}(u^{(m)},{\underline{z}}^{(m)})$
(see Section~\ref{subsection:modeling_tools}) the problem can be seen
as an HMM reestimation problem. The EM algorithm can be employed to solve it. 
Convergence takes place in exactly one EM iteration to:
$$C_{i+1}^{(m)}(u^{(m)}, {\underline{z}}^{(m)}) =
a_{{\theta}_i}((u^{(m)},{\underline{z}}^{(m)}))$$.
%%% Local Variables: 
%%% mode: latex
%%% TeX-master: "main"
%%% TeX-master: "main"
%%% TeX-master: "main"
%%% TeX-master: "main"
%%% TeX-master: "main"
%%% End: 

\bibliography{thesis}
\newpage
\begin{vita}
  Ciprian Chelba received a Diploma Engineer title from "Politehnica"
  University, Bucharest, Romania, the Faculty of Electronics and
  Telecommunications, in 1993.  The Diploma Thesis ``Neural Network
  Controller for Buck Circuit'' has been developed
  at Politecnico di Torino, Italy, under the joint advising of
  Prof.~Vasile Buzuloiu and Prof.~Franco Maddaleno, on a Tempus grant
  awarded by the EU. He received an MS degree from The Johns Hopkins
  University in 1996.
  
  He is member of the IEEE and the Association for Computational
  Linguistics.
\end{vita}
\newpage
\end{document}